\documentclass{article}

\usepackage[preprint]{neurips_2025}

\usepackage[utf8]{inputenc} % allow utf-8 input
\usepackage[T1]{fontenc}    % use 8-bit T1 fonts
\usepackage{hyperref}       % hyperlinks
\usepackage{url}            % simple URL typesetting
\usepackage{booktabs}       % professional-quality tables
\usepackage{amsfonts}       % blackboard math symbols
\usepackage{nicefrac}       % compact symbols for 1/2, etc.
\usepackage{microtype}      % microtypography
\usepackage{xcolor}         % colors

\usepackage{microtype}
\usepackage{graphicx}
\usepackage{subfigure}
\usepackage{booktabs} % for professional tables
\usepackage[utf8]{inputenc}
   \DeclareUnicodeCharacter{2212}{\textminus}

\usepackage{hyperref}

\usepackage{amsmath}
\usepackage{amssymb}
\usepackage{mathtools}
\usepackage{amsthm,bm,bbm}
\usepackage{stackengine}
\usepackage{siunitx}  % For better alignment of numbers
\usepackage{subfigure} % For side-by-side figures
\usepackage{booktabs}  % For better table formatting
\usepackage{adjustbox} % To adjust the table size

\usepackage{soul,color}
\usepackage{url}
\usepackage{graphicx}
\usepackage{xspace}
\usepackage{xcolor}
\usepackage{wrapfig}
\usepackage{lipsum}
\usepackage{longtable}
\usepackage{tikz}
\usetikzlibrary{tikzmark}
\usepackage{enumitem}
\setlist{leftmargin=5.5mm}

\usepackage{booktabs}
\usepackage{multirow}
\usepackage{threeparttable}
\usepackage{caption}
\usepackage{rotating}
\usepackage{algorithm}
\usepackage{algorithmic}
\usepackage{subcaption}
\usepackage{colortbl}

\newcommand{\boldres}[1]{{\textbf{\textcolor{red}{#1}}}}
\newcommand{\secondres}[1]{{\underline{\textcolor{blue}{#1}}}}

\definecolor{fbApp}{HTML}{ffe4e3}
\definecolor{tabhighlight}{HTML}{e5e5e5}

\newcommand{\method}{\textsc{YingLong}\xspace}

\definecolor{pink}{rgb}{1, 0, 0.5}
\definecolor{darkgrey}{rgb}{0.53,0.53,0.53}
\definecolor{mygrey}{rgb}{0.9,0.9,0.9}
\definecolor{purple}{RGB}{230, 227, 254}
\definecolor{lightgreen}{RGB}{238, 252, 241}
\definecolor{lightred}{RGB}{231, 187, 187}
\definecolor{darkred}{RGB}{198, 129, 129}

\definecolor{tabhighlight}{HTML}{e5e5e5}
\definecolor{someorange}{rgb}{0.773,0.353,0.067}
\definecolor{someblue}{rgb}{0.27, 0.35, 0.760}
\definecolor{codegreen}{rgb}{0,0.5,0}
\definecolor{codeblue}{rgb}{0.25,0.5,0.5}
\definecolor{codegray}{rgb}{0.6,0.6,0.6}

\usepackage[most]{tcolorbox}
\newcommand{\rowc}{\rowcolor{fbApp}}

\usepackage[capitalize,noabbrev]{cleveref}

\theoremstyle{plain}
\newtheorem{theorem}{Theorem}[section]

\newtheorem{lemma}[theorem]{Lemma}

\theoremstyle{definition}

\theoremstyle{remark}

\usepackage[textsize=tiny]{todonotes}

\title{Output Scaling: \textsc{YingLong} \\
Delayed Chain of Thought in a Large Pretrained Time Series Forecasting Model}

\author{%
Xue Wang$^*$ \quad Tian Zhou$^*$ \quad Jinyang Gao \quad \textbf{Bolin Ding} \quad Jingren Zhou\\
\texttt{\{xue.w,tian.zt,jinyang.gjy,bolin.ding,jingren.zhou\}@alibaba-inc.com }\\
}

\begin{document}

\maketitle

\let\thefootnote\relax\footnote{$*$ Equal contribution}

\begin{abstract}
% \begin{figure}
% % \footnotesize
% \centering
% %\stackunder{\includegraphics[width=0.38\textwidth]{figures/inference.png}}{Delayed Chain of Thought (DCoT)}
% \stackunder{\includegraphics[width=0.6\textwidth]{figures/Picture1.png}}{Rank Results in GIFT-Eval Benchmark}%
% % \hspace{1cm}%
% % \caption{November to April}
% \end{figure}

 % We introduce an encoder-only transformer model, \method, for time series forecasting, challenging the prevailing focus on autoregressive LLM-style foundation models. \method is a non-causal, bidirectional attention encoder model trained through mask token recovery, aligning better with language understanding tasks than generation tasks. Our experiments demonstrate a novel scaling effect: Longer outputs significantly improve model accuracy due to the delayed chain-of-thought reasoning in our non-causal framework. Additionally, we boost performance by tackling output variance with a multi-input ensemble. We release four foundation models ranging from 6M to 300M parameters, demonstrating superior results in zero-shot tasks on the ETT and Weather datasets. \textsc{YingLong} achieves more than 60\% best performance.  To ensure generalizability, we assessed the models using the GIFT-Eval benchmark, which comprises 23 time series datasets across 7 domains. \textsc{Yinglong} significantly outperformed the time-series foundation models, end-to-end trained models by 15\% and 30\% in rank respectively. Code and pretrained-models will be publicly available at https://anonymous.4open.science/r/Yinglong-7C65
We present a joint forecasting framework for time series prediction that contrasts with traditional direct or recursive methods. This framework achieves state-of-the-art performance for our designed foundation model, \method, and reveals a novel scaling effect: longer outputs significantly enhance model accuracy due to delayed chain-of-thought reasoning in our non-causal approach. \method is a non-causal, bidirectional attention encoder-only transformer trained through masked token recovery, aligning more effectively with language understanding tasks than with generation tasks. Additionally, we boost performance by tackling output variance with a multi-input ensemble. We release four foundation models ranging from 6M to 300M parameters, demonstrating superior results in zero-shot tasks on the ETT and Weather datasets. \textsc{YingLong} achieves more than 60\% best performance.  To ensure generalizability, we assessed the models using the GIFT-Eval benchmark, which comprises 23 time series datasets across 7 domains. \textsc{Yinglong} significantly outperformed the best time-series foundation models, end-to-end trained models by 14\% and 44\% in rank respectively.The pretrained 300M model is available at \url{https://huggingface.co/qcw1314/YingLong_300m}
\\

\begin{minipage}{0.8\textwidth}
\centering
\includegraphics[width=\textwidth]{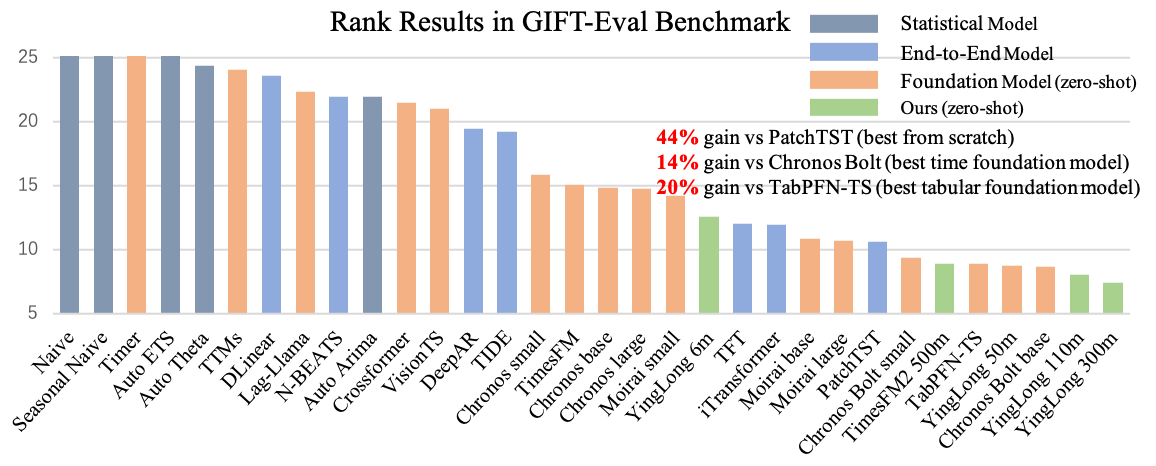}
% \footnotesize{Rank Results in GIFT-Eval Benchmark}
\end{minipage}
\end{abstract} 
% \begin{minipage}{0.8\textwidth}
% \centering
%\includegraphics[width=\textwidth]{figures/Picture1.png}
% \includegraphics[width=\textwidth]{figures/maingraph_GIFT.png}

% \footnotesize{Rank Results in GIFT-Eval Benchmark}
% \end{minipage}
% \begin{figure*}[b]
% \includegraphics[width=\textwidth]{figures/Picture111.png}
% \end{figure*}
\begin{figure*}[t]
\centering
\stackunder{\includegraphics[width=0.9\textwidth]
{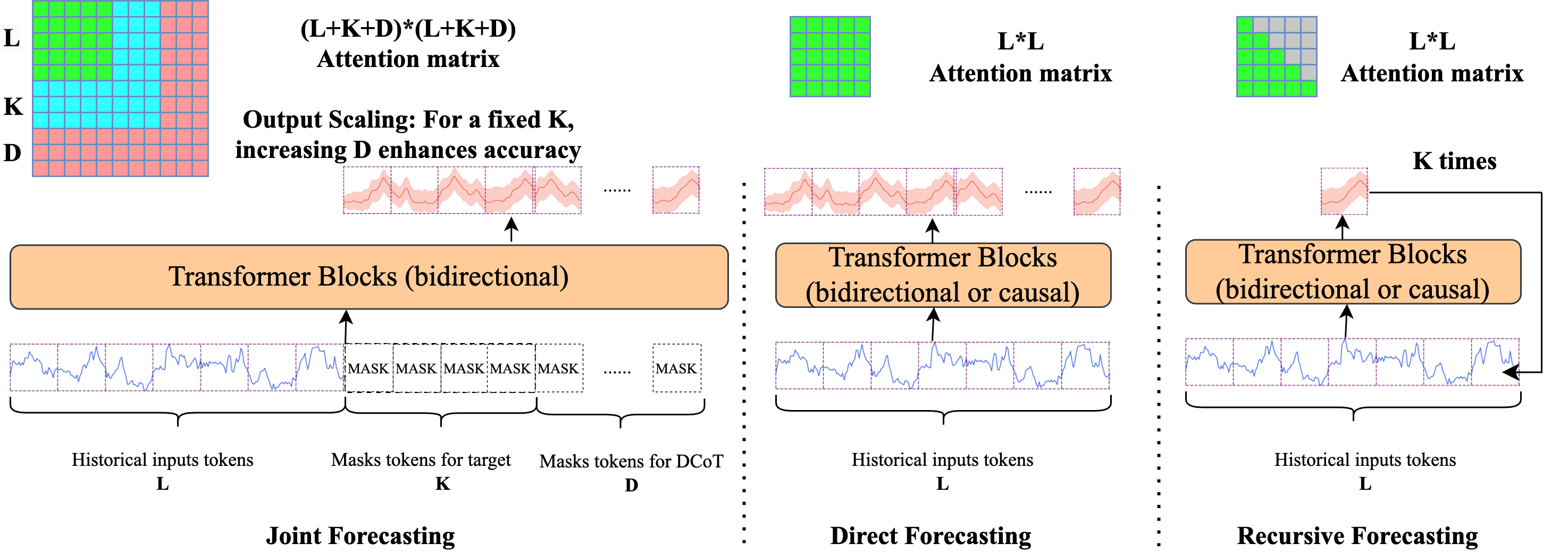}}{}
\caption{ Joint Forecasting with Output Scaling via Delayed Chain of Thought (DCoT). In joint forecasting, each transformer block employs a fully dense attention of size $(L + K + D) \times (L + K + D)$. In the direct forecasting paradigm, a dense attention map of size $L \times L$ is used. Conversely, recursive forecasting utilizes either a half-dense or fully dense $L \times L$ attention map.}
\vskip -0.2in
\label{fig:joint_forecasting}
\end{figure*}
\vspace{-0.5cm}
\section{Introduction}
\vspace{-0.35cm}
Time series data play a crucial role in dynamic real-world systems and applications across various domains~\citep{box2015time,zhang2024self,liang2024foundation}. Analyzing such data is inherently challenging due to their complexity and distribution shifts, yet gaining insights from them is essential for enhancing predictive analytics and decision-making. 
% {\color{cyan}Two major paradigms,  
% rolling forecasting and direct forecasting, have been widely explored in various settings. Classic forecasting methods, like Holt-Winters model, autoregressive integrated moving average (ARIMA), and Prophet~\cite{taylor2018forecasting} apply the rolling forecasting. Those methods enjoy good efficiency in modeling relatively simple problems. When switching to more complex cases, the forecasting competitions in Kaggle and KDD (e.g., \citealt{m5-forecasting-accuracy,zhou2022sdwpf,makridakis2022m5,makridakis2024m6}), multiple winning solutions have primarily employed direct forecasting as their core strategy}. Traditionally, forecasting has been approached in a task-specific, end-to-end manner using statistical or deep learning models. 

Both recursive and direct forecasting paradigms possess inherent limitations that necessitate novel approaches. The recursive forecasting method often operates under the assumption that time series signals exhibit a causal and auto-regressive generative nature. However, this assumption frequently fails to account for the complexities inherent in many time series datasets, which are affected by latent driving factors that are not readily observable and do not exhibit self-autoregressive behavior. Moreover, even in recursive forecasting without a strict causal mask, error accumulation remains a significant challenge when using predicted values as inputs. So forecasting shares similarities with natural language understanding (NLU), wherein the integration of comprehensive input signals through bidirectional information flow is essential for accurate outcome prediction. In the well-established NLU METE Benchmark (Massive Text Embedding Benchmark)~\citep{muennighoff2022mteb}, large language models (LLMs) have surpassed smaller BERT-based models, establishing themselves as the leading solutions within the benchmark. Notably, the top-performing LLM approaches have adopted non-causal bidirectional schemes, utilizing edited masks to enhance their performance~\citep{lee2024nv,li2023towards,gao2023retrieval}.

Conversely, the direct forecasting method, while often exhibiting superior numerical performance in various time series prediction tasks compared to recursive forecasting, faces its own challenges. In direct forecasting, each output is predicted independently, employing a large transformer backbone as a feature extraction mechanism with a single output layer for each prediction point. This approach assumes a complete non-causal and non-auto-regressive nature between outputs, which, similar to recursive forecasting, may not be representative of the intrinsic dynamics of numerous time series datasets. 

Our objective is to integrate the strengths of both recursive and direct forecasting methods by harnessing correlated output modeling while alleviating the constraints of strict causal relationships. We introduce a novel approach termed the joint forecasting paradigm, which promises to be a more robust framework for time series prediction tasks. Within this paradigm, we have developed a large pre-trained time series forecasting model based on a non-causal bidirectional encoder-only transformer architecture. By employing mask token recovery during training, we aim to fully exploit the bidirectional flow of information, offering an innovative method that enhances forecasting accuracy.

With this joint forecasting paradigm, we not only achieve state-of-the-art performance but also uncover an intriguing output scaling effect: the longer the overall forecast duration, the greater the predictive accuracy for outputs of a fixed length. We refer to this phenomenon as the delayed chain-of-thought (DCoT). By incorporating a non-causal bidirectional approach, tokens on the right can influence those on the left. Unlike traditional causal chain-of-thought (CoT) methods, our model allows for thoughts and reasoning tokens to be positioned after the final answers or targets. This newly discovered scaling effect significantly enhances the model's predictive power, as demonstrated in Figure~\ref{fig:ablation:DCOT}. Figure~\ref{fig:joint_forecasting} compares the joint forecasting paradigm with direct and recursive forecasting approaches. 

Furthermore, we propose a multi-input ensemble method to address challenges related to input sequence length in time series forecasting. Extended lookbacks may better capture low-frequency patterns, while shorter lookbacks might be more effective for high-frequency patterns. By employing multi-input ensembling during inference, we aim to enhance prediction accuracy and reduce forecasting variance.
% we cannot assess the quality of the generated value by merely examining the output tokens. Research indicates that different COT designs can produce significantly varying results. 
% We therefore adopted the multi-input ensembling approach during the inference stage. 
% This framework effectively reduces variance and significantly improves prediction accuracy.

In a nutshell, our contributions can be summarized as follows:
\begin{enumerate}[leftmargin=*,label=\textbf{\arabic*.}]
%\vspace{-0.5cm}
\item We introduce a novel joint forecasting framework for time series prediction, distinct from the conventional direct and recursive forecasting approaches. Our framework unveils an unexpected and innovative scaling phenomenon: output scaling. We further develop a delayed chain-of-thought (DCoT) method to exploit this effect. To the best of our knowledge, this is the first work to show this scaling effect, connected to the COT process. The DCoT method significantly enhances the performance of our model, achieving an improvement exceeding 10.5\% in MASE on the GIFT-Eval benchmark. as shown in figure~\ref{fig:ablation:DCOT}.

\item We present \method, a flexible encoder-only architecture for time series forecasting utilizing a masking token during training. Our approach stems from the premise that time series forecasting shares greater similarity with natural language understanding (NLU) rather than natural language generation (NLG). \method achieves exceptional performance across a variety of zero-shot forecasting tasks on both traditional ETT and weather datasets. In our evaluations using the GIFT-Eval benchmarks, which include 23 datasets, we outperformed all baseline methods in CRPS and ranking metrics by a large margin.
%This is an instance where a foundation model has significantly exceeded PatchTST in a zero-shot setting across such large and complex datasets from diverse domains.
\item We propose an innovative ensemble method that leverages mirror symmetry with inputs of varying lengths. This approach serves as a "free lunch," boosting performance with n-time inference compared to traditional single inference.
\end{enumerate}

\vspace{-0.3cm}
\section{Related Works}
\vspace{-0.1cm}
\label{sec:related_works}
\noindent\textbf{Time Series Forecasting.} In recent years, deep learning models have significantly advanced time series forecasting, primarily categorized into: (1) \emph{univariate models}, like TFT~\citep{lim2021tft}, DeepAR~\citep{salinas2020deepar}, and N-BEATS~\citep{oreshkinn}, focused on single time series, and (2) \emph{multivariate models} that include transformer-based techniques~\citep{wu2021autoformer,zhou2022fedformer,patchtstnietime,liuitransformer} and others~\citep{sen2019think,zhou2022film}. Although these models excel within their training areas, the quest continues for pretrained models like LLMs with powerful zero-shot generalization capabilities.

\noindent\textbf{Time Series Foundation Models.} Universal forecasting approaches with foundational time series models can be divided into: (1) \emph{Encoder-only models}, such as Moirai \citep{woo2024unified}, Moment \citep{goswamimoment} and VisionTS \cite{chen2024visionts} utilizing masked reconstruction, and TabPFN~\citep{hoo2025tabular} encoding to tabular data; (2) \emph{Encoder-decoder models}, like Chronos \citep{ansari2024chronos} and TTMS \cite{ekambaram2024tiny} with T5 and MLP architectures respectively; and (3) \emph{Decoder-only models}, including TimesFM \citep{dasdecoder} Lag-Llama \citep{rasul2023lagllama} and Timer \citep{liutimer}. Our work follows the encoder-only approach, highlighting the efficiency of the masked reconstruction learning framework and its capability for inference scaling, leading to superior zero-shot forecasting accuracy through self-consistency ensembles. We argue that time series forecasting aligns more with tasks requiring bidirectional consistency, unlike decoder-based generative tasks handled by LLMs \citep{ni2021sentence}.

\noindent\textbf{Chain-of-Thought (CoT) Reasoning.} Chain-of-thought refers to methods generating intermediate reasoning before deriving a final answer, including LLM prompting \citep{yang2024large,wei2022chain} and reasoning chain training \citep{zhou2022least}. CoT enhances model expressivity \citep{feng2024towards} by feeding generated outputs back as inputs, yet its autoregressive nature limits complex reasoning tasks requiring planning \citep{xie2024self}.In our work, we utilize a bidirectional attention encoder-only framework for time series forecasting, effectively leveraging latent COT reasoning to overcome the autoregressive constraints and discrete language space limitations commonly found in large language models (LLMs).

%\noindent\textbf{Ensemble .}

% In contrast to these dense models, \method introduces a scalable, unified architecture with a sparse mixture-of-experts design, optimized for larger time series forecasting models while reducing inference costs. Trained on our \dataset dataset, comprising over 300B time points, \method is scaled to 2.4B parameters for the first time. It outperforms existing models with the same number of activated parameters, significantly enhancing both model efficiency and forecasting precision, while avoiding limitations such as fixed context lengths or hardcoded heuristics.

% \input{tables/related_work.tex}

\vspace{-0.3cm}
\section{Methods}
\vspace{-0.1cm}
{\bf Preliminary}\
% \ We consider the univariate forecasting setting. Let $x_{1},x_2,...,x_{T}$ be the historical observation from time $t = 1$ to $t= T$ and our goal is to predict the next $P$ times points. We denote $\hat{\bm{x}}_{T+i}$ and $x_{T+i}$ as the set of predicted statistics and real observation for time $T+i$ for $i=1,2,..,P$. Without loss of generality, we refer to the predicted statistics as the $\alpha$-quantiles from some $\alpha\in (0,1)$.
We consider the univariate forecasting setting. Let $x_{1}, x_{2}, \ldots, x_{T}$ be the historical observations from time $t = 1$ to $t = T$. Our goal is to predict the next $P$ time points. We denote by $\hat{\bm{x}}_{T+i}$ the predicted statistics and by $x_{T+i}$ the real observation at time $T+i$, for $i = 1, 2, \ldots, P$. Without loss of generality, we refer to the predicted statistics as $\alpha$-quantiles, for some $\alpha \in (0,1)$. % Without loss of generality, we refer to the predicted statistics as $\alpha$-quantiles, for some $\alpha \in (0,1)$.
% Due to the autoregressive nature of the time series data, in order to optimize a probabilistic forecasting model $f_{\theta}$ parameterized with $\theta$, a straightforward loss choice would be the negative log-likelihood function: 
% \begin{align}
%     \min_{\theta} - \mathbbm{E}\left[\sum_{j=1,...,T-1}\log f_{\theta}(x_{j+1}|x_{1:j})\right]\label{eq:loss_n}
% \end{align}
% where $x_{1:j} = \{x_1,...,x_j\}$ and $f_{\theta}$ is a model with learnable parameter $\theta$.
For probabilistic forecasting models, the negative log-likelihood loss functions for direct and recursive paradigms can be compactly expressed as:
\vspace{-0.1cm}
\begin{align}
\text{Direct:} \min_{\theta} -\mathbb{E}\left[\sum_{k=j+1}^{T-1} \log f_{\theta}(x_k \mid x_{1:j})\right], \quad
\text{Recursive:} \min_{\theta} -\mathbb{E}\left[\sum_{j=1}^{T-1} \log f_{\theta}(x_{j+1} \mid x_{1:j})\right]
\label{eq:loss_n}
\end{align}
where $x_{1:j} = \{x_1, \ldots, x_j\}$.
This formulation highlights the structural similarity between paradigms: direct forecasting optimizes multi-step predictions, while recursive forecasting optimizes single-step predictions with autoregressive conditioning. The negative log-likelihood ensures probabilistic calibration by minimizing divergence between predictions and observations.

Although prevalent in time series modeling, both paradigms have limitations. Recursive forecasting focuses on transitions between observations but can accumulate errors over long horizons. Direct forecasting uses independent models for each horizon, enabling parallel predictions but sacrificing scalability and inter-output correlation, potentially compromising temporal coherence.

A toy example for illustration: Consider a random walk starting from the origin: $x_0 = 0$, with $x_{t+1} = x_t + \epsilon$, where $\epsilon \sim (-1,1)$ for $t=1,2,\ldots$. Given observations up to time $m$, we aim to forecast the next $n$ points using mean square error (MSE) as the metric. The optimal forecasts here are $\hat{x}_{m+j} = x_m$ for $j = 1, 2, \ldots, n$.

Assume an ideal model $f_{\theta}$ perfectly learns this one-step process, i.e., $f_{\theta}(x_{t+1} \mid x_{t}) = x_{t} \pm 1$ with equal probability. Generating $N$ sample paths with $f_{\theta}$ and averaging them provides the future point forecast, illustrating the anti-concentration phenomenon as follows:

\begin{lemma}\label{lem:1}
    Let $\{x_{m+1}^i,\ldots,x_{m+n}^i\}$ for $i=1,\ldots, N$ be $N$ sample paths drawn from $f_{\theta}$  starting at $x_{m}$. For $\epsilon \in (0, \sqrt{j/N})$, there exists a positive constant $C$ such that
    \begin{equation*}
        \mathbbm{P}\Bigl(\Bigl|\tfrac{1}{N}\sum_{i=1}^N x_{m+j}^i \;-\; x_{m}\Bigr| \;>\; \epsilon\Bigr)
        \;\ge\; C\Bigl(1-\tfrac{\epsilon^2 N}{j}\Bigr)^2.
    \end{equation*}
    % \vspace{-0.3cm}
\end{lemma}

% \begin{lemma}\label{lem:1}
%     Let $\{x_{m+1}^i,\ldots,x_{m+n}^i\}$ for $i=1,\ldots, N$ be $N$ sample paths drawn from $f_{\theta}$ starting at $x_{m}$. For $\epsilon \in (0, \sqrt{j/N})$, there exists a positive constant $C$ such that
%     \begin{equation*}
%         \mathbbm{P}\left(\left|\frac{1}{N}\sum_{i=1}^N x_{m+j}^i - x_{m}\right| > \epsilon\right) \ge C\left(1-\frac{\epsilon^2 N}{j}\right)^2
%     \end{equation*}
% \end{lemma}
% \vspace{-0.3cm}
% \begin{proof}
%     See Section~\ref{proof:lem:1} in Appendix.
% \end{proof}
% \vspace{-0.2cm}
%Now we suppose we have already perfectly learned the underlying process and the forecasts are got by recursively generating time points. Since the change between the adjacent points can never be zero, the 

% The ultimate goal of time series forecasting is to ensure multiple time points 

Lemma~\ref{lem:1} shows that for any fixed number of sample paths, there is always a non-negligible probability that the average estimator will deviate significantly from its target. Moreover, as the time horizon $j$ grows, the chance of such a large deviation also increases. This suggests that relying on an average of sample paths can be problematic over relatively long forecasting windows. 
% We believe Lemma~\ref{lem:1} also provides an explanation for why patch-level models are popular in the literature, as they effectively reduce the forecasting horizon and thus mitigate these deviations.

% Above Lemma \ref{lem:1} states that for a fixed number of sample paths there always exists a positive chance that the average estimator can't go close enough to its target. Moreover, when the time step goes deep (e.g., larger $j$), the large deviation chance becomes larger. It implies
% that the sample path average approach may suffer when the predicted sequence goes relatively long. And we believe Lemma \ref{lem:1} can also provide an explanation on why the patch-level models are more popular in literature as it could significantly shorten the forecasting length. 

Joint forecasting enhances performance beyond time series forecasting, benefiting NLP and CV domains as well. Diffusion methods generate all image tokens concurrently, and recent autoregressive image models adopt joint token generation approaches \cite{chang2023muse,teng2024accelerating}. Although language models are typically recursive, techniques like beam search \cite{freitag2017beam} capture token correlations and compute joint probabilities, unlike the greedy methods used in time series forecasting. Recent studies show that simultaneous token generation improves performance \cite{gloeckle2024better}. We propose that this paradigm may shape the future of time series forecasting.

\begin{figure*}[t]
\centering

\includegraphics[width=0.80\textwidth]{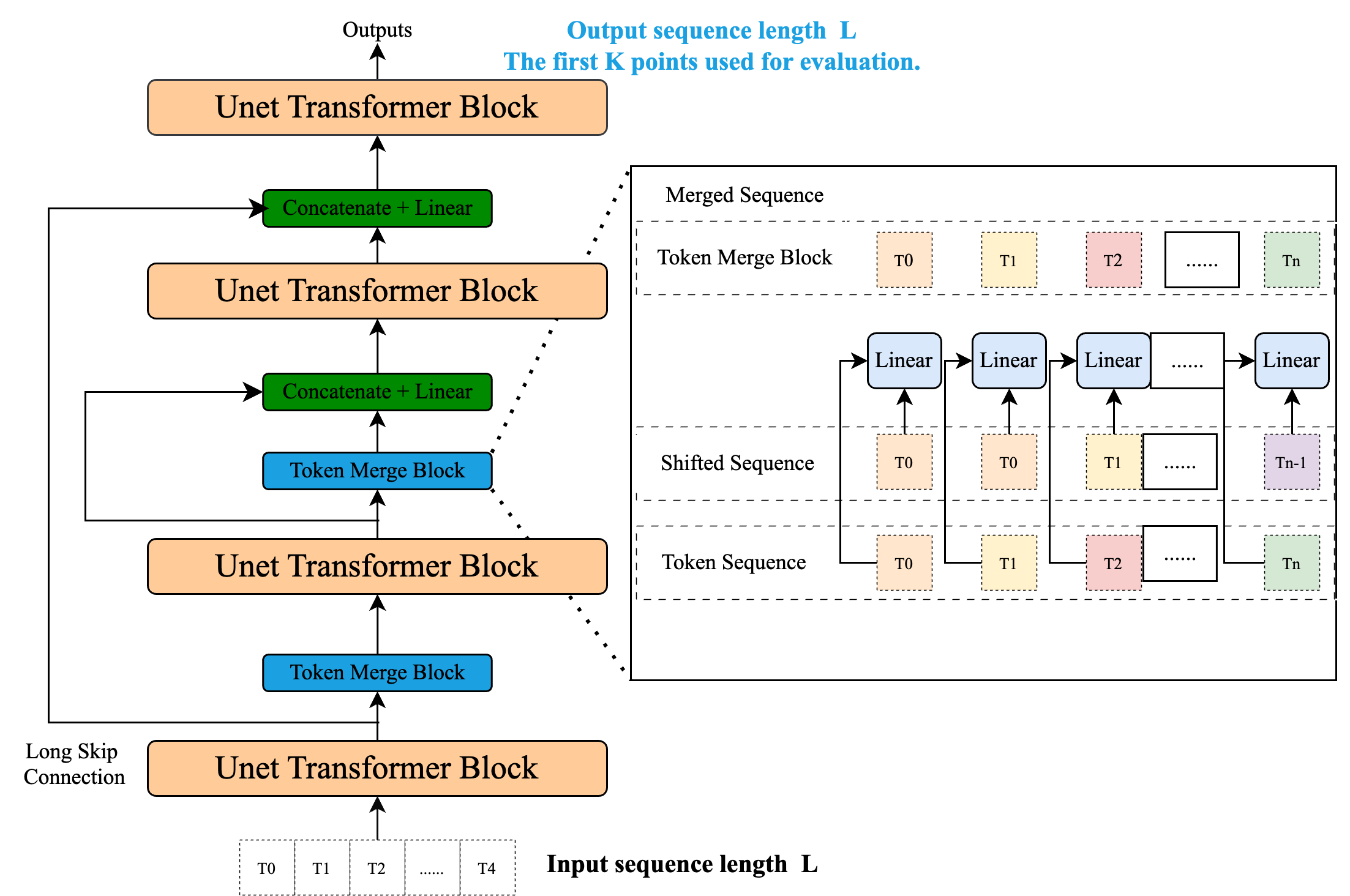}
%\includegraphics[width=0.80\textwidth]{figures/unet2.drawio.png}
%\vspace{2mm}
\caption{Illustration of the architecture of \method.}
\label{fig:architecture}
\vskip -0.2in
\end{figure*}
\section{Model Architecture}
\vspace{-0.3cm}
To overcome these limitations, we propose a joint forecasting framework using a non-autoregressive architecture that generates probabilistic predictions for all temporal horizons in parallel while preserving inter-horizon dependencies. One architectural solution is a bidirectional transformer with masked tokens (see Figure~\ref{fig:architecture}). We describe our model architecture in detail below. Notably, \textbf{a vanilla transformer still achieves top performance on the GIFT-Eval benchmark} (see structural ablation section), demonstrating that our Unet transformer design is not essential. \textbf{The key factor is the joint forecasting framework leveraging DCoT}.

% To overcome the drawbacks aforementioned, we proposed to consider using non-autoregressive style forecasting, i.e., directly predicting the targeted statistics for all forecasting horizons simultaneously. One possible solution is through a bi-direction transformer model with masked tokens. An illustration example is shown in Figure~\ref{fig:architecture}.  We next provide the details description on our model construction.

\subsection{Tokenization}

We adopt the patching technique \citep{nie2023patchtst} to convert the input time series $[x_1, \ldots, x_T]$ into tokens, where the patch length is $P$. The series is transformed into a matrix $\tilde{X} \in \mathbb{R}^{N \times P}$, where $N = \lfloor T/P + 1 \rfloor$. During training, we randomly exclude a fraction $\rho \in (0,1)$ of the patches as forecast targets. In testing, we append these masked patches as placeholders, extending the sequence to $X_{\mathrm{m}}$ with corresponding indices $\mathcal{M}$.

\subsection{Unet-Transformer}

For our transformer block, we adopt the standard architecture \citep{vaswani2017attention} with bidirectional attention. We use RMSNorm \citep{zhang2019root} for pre-normalization, SwiGLU \citep{shazeer2020glu} as the activation function, and rotary positional embeddings \citep{su2024roformer}.

\paragraph{U-shaped design}
Inspired by U-shaped generative models (e.g., \citealt{bao2023all,zhang2023crossformer}), we incorporate a token merging module in the shallow layers and introduce long skip connections from shallow to deep layers. This design allows the network to process varying granularities of information (from coarse to fine), which is beneficial for point-level forecasting. The skip connections help propagate fine-grained information through the network, improving learning and performance.

Let $x^{i,j,\mathrm{in}}, x^{i,j,\mathrm{out}} \in \mathbb{R}^{1\times d}$ represent the input and output of the $i$-th token in the $j$-th transformer block, with $i=1,2,\ldots,N$ and $j=1,2,\ldots,D$. Assuming $D$ is even, after each shallow layer we generate coarse-grid tokens using a fully connected layer $F_c: 2d \to d$:
\begin{align}
    x^{i+1,j+1,\mathrm{in}}
    \;=\; F_{c}\Bigl(\bigl[x^{i,j,\mathrm{out}},\, x^{i+1,j,\mathrm{out}}\bigr]\Bigr).
\end{align}
% \vspace{-0.2cm}
Moreover, before each deep layer, we apply another merging layer $F_m: 2d \to d$ to combine tokens from the corresponding shallow layer:
\begin{align}
    x^{i,j+1,\mathrm{in}}
    \;=\; F_{m}\Bigl(\bigl[x^{i,j,\mathrm{out}},\, x^{i,D-j+1,\mathrm{out}}\bigr]\Bigr).
\end{align}
\vspace{-0.5cm}

\subsection{Output Layer and Loss Function}
In this work, we directly predict $R$ quantiles using a set of fully connected layers $F_{\alpha_k}: d \to P$ for each $\alpha_k \in (0,1)$, $k=1,2,\ldots,R$:
\begin{align}
    q_{\alpha_k}^i \;=\;
    F_{\alpha_k}\Bigl(x^{i,D,\mathrm{out}}\Bigr)
    \;\cdot\;\bigl(\sigma_{X_{\mathrm{m}}} + \epsilon\bigr)
    \;+\; \mu_{X_{\mathrm{m}}},
\end{align}
where $q_{\alpha_k}^i$ is the predicted $\alpha_k$-quantile for token $i$.
\paragraph{Weighted quantiles loss}
We adopt a weighted quantiles loss (WQL), leading to the following optimization objective:
\begin{align}
    \min_{\theta}
    \;\mathbbm{E}\Biggl[\sum_{i\in\mathcal{M}}
    \sum_{k=1}^R
    \sum_{s=1}^P
    w_{\alpha_k,x^i} \,
    \ell_{\alpha_k}\bigl(q_{\alpha_k}^{i,s},\, x^{i,s}\bigr)\Biggr],
    \label{eq:loss_g}
\end{align}
where $q_{\alpha_k}^{i,s}$ and $x^{i,s}$ denote the $s$-th elements of $q_{\alpha_k}^{i}$ and the ground-truth patch $x^{i}$, respectively. Here, $\ell_{\alpha_k}$ is the standard $\alpha_k$-quantile loss, and $w_{\alpha_k,x^i} > 0$ is a weight parameter that balances the contribution of each individual quantile loss term.

\vspace{-0.2cm}
\section{Output Scaling}
\subsection{Delayed Chain of Thoughts}
Beyond the efficient direct prediction scheme described earlier, our approach also facilitates a new form of \emph{chain of thoughts} (CoT) for time series forecasting. Chain of thoughts has been extensively studied in the field of natural language processing (NLP), where generating additional tokens (i.e., “thoughts”) can significantly improve model performance. In a classic NLP setting, CoT is obtained by training on sequences of the form [\emph{prompt, CoT, target}].

In time series forecasting, future data is typically unavailable, rendering conventional Chains of Thought (CoT), which position intermediate tokens before the target, impractical. Historical observations function as the "prompt," directly preceding the forecasting targets. We propose Delayed CoT (DCoT), wherein time points beyond the forecasting horizon form the chain of thoughts. This approach exploits the periodic behavior and low-frequency structures inherent in time series data, allowing certain future points to be more predictable and serve as conditional anchors for target forecasting.

\vspace{-0.3cm}
\paragraph{Probabilistic Interpretation of CoT} Consider the probabilistic viewpoint of a classical CoT sequence:
\[
    \mathrm{prompt},\, \mathrm{CoT}_1,\,\ldots,\mathrm{CoT}_n,\,\mathrm{target}.
\]
With $n$ intermediate CoT steps, improved accuracy is expressed as:
\[
    \mathbbm{P}\bigl(\mathrm{target} = \mathrm{truth} \mid \mathrm{prompt},\, \mathrm{CoT}_1,\ldots,\mathrm{CoT}_n\bigr) \ge \mathbbm{P}\bigl(\mathrm{target} = \mathrm{truth} \mid \mathrm{prompt}\bigr).
\]
This suggests that CoT tokens provide additional information, thereby increasing the likelihood of correctly predicting the target. In unidirectional language models, such auxiliary information appears before the target. However, when using a bidirectional model:
\[
    \mathrm{prompt},\, \mathrm{target},\, \mathrm{CoT}_1,\,\ldots,\mathrm{CoT}_n,
\]
can still maintain the same conditional probability structure. Hence, positioning CoT after the target is a delayed CoT (DCoT).

\begin{figure}[t]
\centering

\hspace*{-0.5cm}\includegraphics[width=0.8\textwidth]{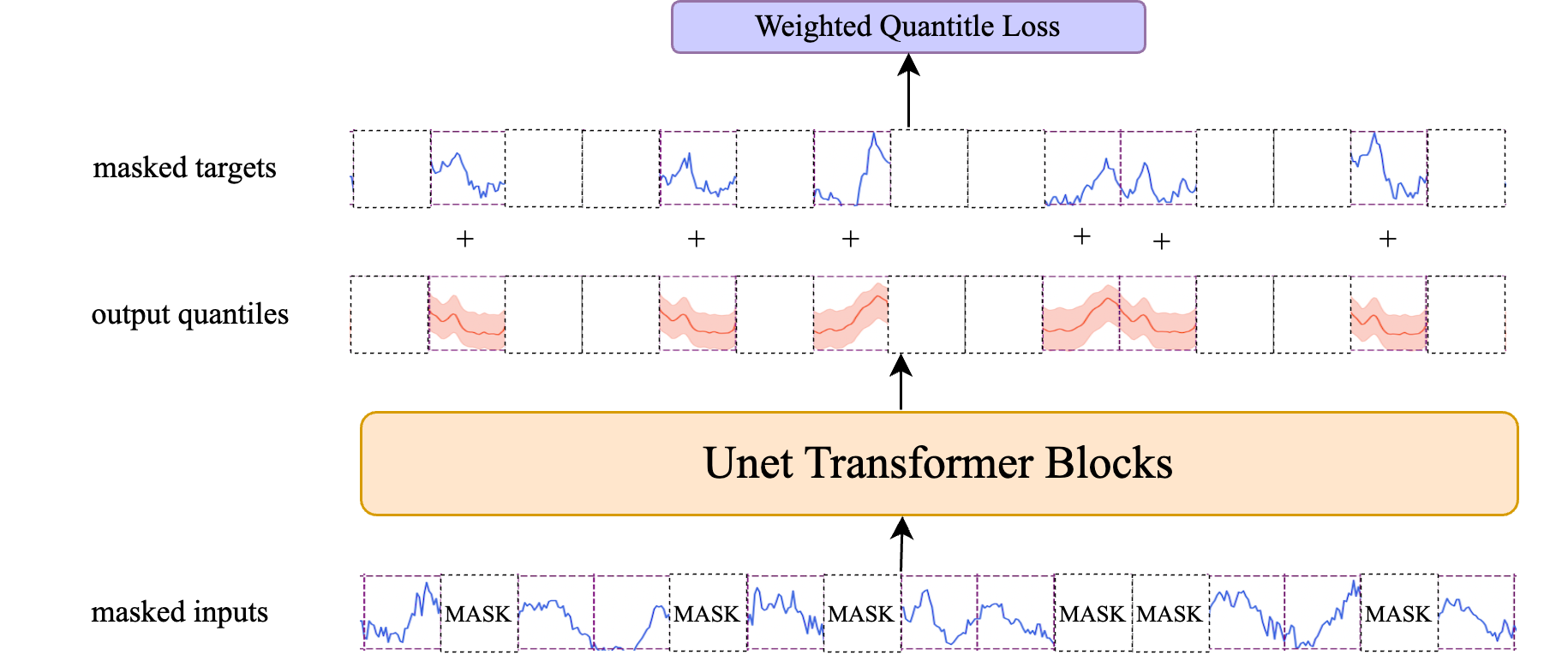}
%\vspace{2mm}
\caption{Training with Masked Token Prediction.}
\label{fig:Train}
\vskip -0.2in
\end{figure}

% \vspace{-0.2cm}

% {\bf \color{cyan}better normalization}
% \vspace{-0.2cm}
\subsection{Multi-Input Ensemble}
% Unlike in natural language processing, where longer contexts often improve model performance, many time series models reach peak forecasting accuracy at a certain input length and then degrade rapidly as the input grows. This phenomenon complicates the design of a unified “foundation” model that handles diverse forecasting tasks.
A challenge in multi-horizon forecasting lies in optimizing the input window length—our preliminary experiments identify an accuracy tradeoff where short inputs enhance immediate predictions but degrade long-horizon performance, while extended windows exhibit the inverse behavior. To mitigate these horizon-specific limitations, we introduce an input-length ensemble that adaptively combines forecasts from varying temporal contexts. Furthermore, we implement temporal mirroring: reversing input sequences while correspondingly flipping prediction targets, then aggregating outputs through ensemble averaging to exploit bidirectional temporal patterns and enhance forecast stability.

Concretely, let $x_{1:n}$ denote the time series over the total lookback window of length $n$. We select $k$ indices $1 = n_1 < n_2 < \dots < n_k < n$ for shorter lookback windows. The final prediction $y$ is then given by:
\vspace{-0.1cm}
\begin{align}
    y \;=\; \frac{1}{2k} \;\sum_{j=1}^k 
    \Bigl[\mathrm{Model}\bigl(x_{n_j : n}\bigr)
    \;-\;
    \mathrm{Model}\bigl(-\,x_{n_j : n}\bigr)\Bigr].
    \label{eq:result_ensemble}
        \vspace{-1cm}
\end{align}
% Specifically, let \( x_{1:n} \) denote the time series over a lookback window of length \( n \). We select \( k \) indices \( 1 = n_1 < n_2 < \dots < n_k < n \) for shorter lookback windows. The final prediction \( y \) is then given by
% \begin{equation}
%     y = \frac{1}{2k} \sum_{j=1}^k \left[ \mathrm{Model}\left(x_{n_j : n}\right) - \mathrm{Model}\left(-x_{n_j : n}\right) \right].
%     \label{eq:result_ensemble}
% \end{equation}
While existing ensemble approaches combine multiple models trained under varied configurations \cite{oreshkin2020nbeats}, our method achieves enhanced forecasting robustness through input-space bootstrapping—applying sign inversion and variable-length sampling to a single pre-trained model. Crucially, our architecture generates all horizon predictions via parallel computation in one forward pass (eliminating sample-path averaging), maintaining computational efficiency even when implementing the ensemble strategy in \eqref{eq:result_ensemble}.

\vspace{-0.2cm}
\section{Experiments}
\subsection{Datasets and Training Details}
% In the training stage, we use a subset of the Monash Forecasting Repository \cite{godahewa2021monash}, 5.625 degree WeatherBench \cite{rasp2020weatherbench} and the Gaussian process synthetic data in \cite{ansari2024chronos}. In total it contains $xx$ B time points. In recent literature (e.g., \citealt{lin2024moe,woo2024unified,liu2024autotimes}), scaling law in the size pretraining data has been observed. However, in this paper we don't focus on this direction and we thus try to minimize the training cost due to the training efficiency consideration.

% We consider both point forecasting as well as probabilistic forecasting. To test the point forecasting ability, we adopt four experiments on Electricity Transform Temperature (ETT) datasets \cite{zhou2021informer} comprising of two hourly and two 15-
% minute datasets, one 10-minute weather forecasting experiment \cite{weatherdataset}. We also used the GIFT-Eval \cite{aksu2024giftevalbenchmarkgeneraltime} benchmarks for probabilistic forecasting which contains over 23 datasets. To avoid data leaking we carefully remove the data from the training set that possible exists in GIFT-Eval's testing result. It's worth mention that several baseline's work might has such data leaking problem in their pretraining data, but since we don't have resource to remove clear their training data and retraining the baseline foundation model, we simipliy used their current result in reported GIFT-Eval benchmark. 

In the training phase, we leverage a subset of datasets from the Monash Forecasting Repository \cite{godahewa2021monash}, the $5.625^\circ$ WeatherBench \cite{rasp2020weatherbench}, and a subset of data in \cite{ansari2024chronos}, which altogether comprise approximately 78 billion time points. Recent studies (e.g., \citealt{lin2024moe,woo2024unified,liu2024autotimes}) note the critical impact of pretraining dataset size on overall model performance. However, in this work we do not specifically explore that aspect; rather, we aim to keep training costs low for efficiency.

We evaluate models on various tasks. For point forecasting, we use the ETT datasets \cite{zhou2021informer} (two hourly series and two 15-minute series) and a 10-minute weather forecasting dataset \cite{weatherdataset}. For broader generalization and probabilistic forecasting, we employ the GIFT-Eval benchmarks \cite{aksu2024giftevalbenchmarkgeneraltime}, which cover 23 different datasets. To avoid data leakage, we carefully exclude any data from our training sets that appears in the GIFT-Eval benchmarks. For baseline results, we utilize published data from the GIFT-Eval benchmark.

% In the training phase, we utilize a subset of datasets from the Monash Forecasting Repository \cite{godahewa2021monash}, the 5.625-degree WeatherBench \cite{rasp2020weatherbench}, and Gaussian process synthetic data as described in \cite{ansari2024chronos}. Altogether, these datasets comprise a total of  78  billion time points. Recent studies (e.g., \citealt{lin2024moe,woo2024unified,liu2024autotimes}) have noted the impact of pretraining data size on model performance. However, this paper does not explore this aspect; instead, we aim to minimize training costs due to efficiency considerations.

% We evaluate both point and probabilistic forecasting capabilities. For point forecasting, we conduct experiments using the Electricity Transformer Temperature (ETT) datasets \cite{zhou2021informer}, which include two hourly and two 15-minute datasets, as well as a 10-minute weather forecasting experiment \cite{weatherdataset}. For broader generalization and probabilistic forecasting, we utilize the GIFT-Eval benchmarks \cite{aksu2024giftevalbenchmarkgeneraltime}, which encompass 23 datasets. To prevent data leakage, we meticulously remove any data from our training set that may appear in the GIFT-Eval test results. It is noteworthy that several baseline studies may have encountered data leakage issues in their pretraining data. However, due to resource constraints, we could not cleanse their training data or retrain the baseline foundation model; therefore, we rely on their existing results as reported in the GIFT-Eval benchmark.
\vspace{-0.2cm}
\subsection{Training Details}
We train four models with parameter sizes ranging from 7M to 300M, each for 100{,}000 steps. The batch size is 512, and the maximum sequence length is 8192. We use a patch size of 32, resulting in 256 tokens per sequence, and set the random masking ratio $\rho$ to 0.2. Our optimizer is AdamW with a learning rate of $1\times10^{-4}$, weight decay $0.1$, $\beta_1 = 0.9$, and $\beta_2 = 0.95$. The learning rate schedule includes a linear warmup over 2{,}000 steps followed by cosine annealing. All training is performed on eight NVIDIA A100 at BF16 precision. Refer to Section~\ref{sec:training} in the Appendix for further details.

% We trained four models with parameters from 7M to 300M for 100,000 steps each, using a batch size of 512 and a max sequence length of 8192. The patch size is 32, allowing 256 tokens per sequence. We set the random mask ratio $\rho$ as 0.2. The AdamW optimizer was deployed with a learning rate of 1e-4, weight decay 0.1, $\beta_1 = 0.9$, and $\beta_2 = 0.95$. A learning rate scheduler was used, employing a 2,000-step linear warmup followed by cosine annealing. Training utilized 8 NVIDIA A100-80G GPUs with BF16 precision. More details refer to Section~\ref{sec:training} in Appendix.

% \subsection{Model and Training Configurations}
% We evaluate four models with parameters ranging from 7 million to 300 million, with training details listed in Table xx. Each model trains for 100,000 steps using a batch size of 512 and a maximum sequence length of 8,000. The patch size is set at 32, allowing for a maximum of 256 tokens per sequence. We use the AdamW optimizer with hyperparameters: learning rate of 1e-4, weight decay of 0.1, $\beta_1 = 0.9$ and  $\beta_2 = 0.95$. A learning rate scheduler is applied, featuring a linear warmup for the initial 2,000 steps, followed by cosine annealing. Training is performed across 8 NVIDIA A100-80G GPUs with BF16 precision.

\vspace{-0.2cm}
\subsection{Zero-shot Forecasting}\label{sec:5.3}
\vspace{-0.1cm}

We compare our models against popular end-to-end approaches (FEDformer, TimesNet, DLinear, PatchTST) and recently proposed foundation models (Moirai, TimesFM, Chronos, TimeMoE, VisionTS, and Moment). For end-to-end baselines, we report the best performance from existing literature, while for foundation models we provide the best zero-shot results across all scales. Our model employs a forecasting window of 4096 and ensembles multiple input lengths (512, 1024, 2048, and 4096) for robust predictions. We assess performance using mean squared error (MSE) and mean absolute error (MAE).

% We evaluate our models against popular end-to-end baselines: FEDformer, TimesNet, Dlinear, PatchTST, and recent foundation models: Moirai, TimesFM, Chronos, TimeMoE, VisionTS, and Moment. For end-to-end models, we report the best results from existing literature, and for foundation models, we present the best zero-shot results across all scales. Our model utilizes a forecasting window of 4096 and averages input lengths of 512, 1024, 2048, and 4096. Evaluation metrics include mean square error (MSE) and mean absolute error (MAE).

Table~\ref{tab:zero_shot_short:001} summarizes our performance across four model sizes. \textsc{YINGLONG}$_{\textrm{110m}}$and \textsc{YINGLONG}$_{\textrm{300m}}$ rank highest, with \textsc{YINGLONG}$_{\textrm{110m}}$ achieving top 2 results in 70\% of MSE and 75\% of MAE cases.\textsc{YINGLONG}$_{\textrm{300m}}$ also performs well with 60\% and 90\% in MSE and MAE respectively. In complex weather tasks, we observe a scaling law in model size, while in simpler tasks, \textsc{YINGLONG}$_{\textrm{110m}}$ and \textsc{YINGLONG}$_{\textrm{300m}}$ perform similarly, likely due to a low signal-to-noise ratio. Our smallest model, \textsc{YINGLONG}$_{\textrm{6m}}$, efficiently achieves average ranks of 5.45 (MSE) and 5.00 (MAE), outperforming foundation models 30 times its size. For a more detailed result, please refer to Appendix Table~\ref{tab:zero_shot_full}.
%\vspace{-3mm}

\begin{table*}[htbp]
  \centering
  \caption{Zero-shot forecasting experiments for another set of benchmarks. A lower MSE or MAE indicates a better prediction. TimesFM, due to its use of Weather datasets in pretraining, is not evaluated on this dataset and is denoted by a dash ($-$). {\boldres{Red}}: the best, \secondres{Blue}: the 2nd best.} %The below table represents these results by a dash ($-$).}
  \resizebox{1.1\columnwidth}{!}{
    % \renewcommand{\tabcolsep}{2pt}
    % \scalebox{0.5}{
    \setlength\tabcolsep{1pt}
    \begin{tabular}{c|cc|cc|cc|cc|cc|cc|cc|cc|cc|cc|cc|cc|cc}
          \toprule
          & \multicolumn{8}{c}{Ours} & \multicolumn{12}{c}{Foundation Models}&\multicolumn{4}{c}{End-to-end Models} \\
          \cmidrule(lr){2-9} \cmidrule(lr){10-21} \cmidrule(lr){22-25}
                
           Models& \multicolumn{2}{c}{\textsc{YingLong}$_{\textrm{6m}}$} 
          & \multicolumn{2}{c}{\textsc{YingLong}$_{\textrm{50m}}$} 
          & \multicolumn{2}{c}{\textsc{YingLong}$_{\textrm{110m}}$} 
          & \multicolumn{2}{c}{\textsc{YingLong}$_{\textrm{300m}}$} 
          & \multicolumn{2}{c}{Moirai}
          & \multicolumn{2}{c}{TimesFM}
          & \multicolumn{2}{c}{Moment}
          & \multicolumn{2}{c}{visionTS}
          & \multicolumn{2}{c}{Chronos}
          & \multicolumn{2}{c}{TimeMoE}
         %& \multicolumn{2}{c}{FEDformer}
       %& \multicolumn{2}{c}{TimesNet}
          & \multicolumn{2}{c}{Dlinear}
             & \multicolumn{2}{c}{PatchTST}
          \\
          % \cmidrule(lr){3-4} \cmidrule(lr){5-6}\cmidrule(lr){7-8} \cmidrule(lr){9-10}\cmidrule(lr){11-12}\cmidrule(lr){13-14}\cmidrule(lr){15-16}
          \multicolumn{1}{c}{Metrics}&  \textbf{MSE} & \textbf{MAE} & \textbf{MSE} & \textbf{MAE} & \textbf{MSE} & \textbf{MAE} & \textbf{MSE} & \textbf{MAE} & \textbf{MSE} & \textbf{MAE} & \textbf{MSE} & \textbf{MAE}&
          \textbf{MSE} & \textbf{MAE} & \textbf{MSE} & \textbf{MAE}& 
          \textbf{MSE} & \textbf{MAE} & \textbf{MSE} & \textbf{MAE}& 
          \textbf{MSE} & \textbf{MAE} & \textbf{MSE} & \textbf{MAE}\\
          \midrule

          \multirow{1}[1]{*}{ETTh1} 
          % & 192   & \secondres{0.388} & \secondres{0.412} & 0.395 & 0.413 & 0.434 & 0.415 & 0.465 & 0.434 & 0.688 & 0.560 & 0.502 & 0.424& 0.395 & 0.413\\
          % & 336   & \boldres{0.411} & \secondres{0.430} & 0.447 & 0.453 & 0.495 & 0.445 & 0.503 & 0.456 & 0.675 & 0.563 & 0.576 & 0.467 & 0.447 & 0.453 \\
          % & 720   & \boldres{0.427} & 0.455 & 0.457 & 0.462 & 0.611 & 0.510 & 0.511 & 0.481 & 0.683 & 0.585 & 0.835 & 0.583 & 0.457 & 0.462\\
          %\rowcolor{tabhighlight}
          % & {\textbf{avg}}
          % yelong 6m
          & 0.408 & 0.412
         % yelong 50m
          &  0.405 & \secondres{0.408} 
          % yelong 110m
          & 0.399 & \secondres{0.408}
          % yelong 300m
          & 0.398&\boldres{0.407}
          % Moirai 
          & 0.415& 0.418
          % TimesFM
          & 0.473&0.444
          % Moment
          & 0.684&0.566
        % visionTS
          &\boldres{0.390} & 0.414
          %Chronos
          & 0.524&0.455
          %TimeMoE
          & \secondres{0.393}&0.417
          %FEDformer
          %& 0.440&0.460
          %TimesNet
          %& 0.458&0.450
          %Dlinear
          & 0.423&0.437
          %PatchTST
          &0.413&0.431
          \\
    \midrule
    % \multirow{4}[0]{*}{ETTh2} & 96    & 0.302 & 0.354 & \boldres{0.292} & 0.352 & 0.296 & \boldres{0.330} & 0.315 & 0.349 & 0.342 & 0.396 & 0.345 & 0.320 & \boldres{0.292} & 0.352 \\
    %       & 192   & 0.364 & 0.385 & \boldres{0.347} & 0.379 & 0.361 & \boldres{0.371} & 0.388 & 0.395 & 0.354 & 0.402 & 0.406 & 0.399 & \boldres{0.347} & 0.379 \\
    %       & 336   & 0.417 & 0.425 & 0.406 & 0.419 & 0.390 & \boldres{0.390} & 0.422 & 0.427 & \boldres{0.356} & \secondres{0.407} & 0.492 & 0.453 & 0.406 & 0.419 \\
    %       & 720   & 0.537 & 0.496 & 0.439 & 0.447 & 0.423 & \boldres{0.418} & 0.443 & 0.454 & \boldres{0.395} & 0.434 & 0.603 & 0.511 & 0.439 & 0.447 \\
    %       \rowcolor{tabhighlight}
    %       & {\textbf{AVG}} & 0.405 & 0.415 & 0.371 & 0.399 & 0.367 & \boldres{0.377} & 0.392 & 0.406 & \boldres{0.361} & 0.409 & 0.455 & 0.427 & 0.371 & 0.399 \\
     \multirow{1}[1]{*}{ETTh2} 
          % & 192   & \secondres{0.388} & \secondres{0.412} & 0.395 & 0.413 & 0.434 & 0.415 & 0.465 & 0.434 & 0.688 & 0.560 & 0.502 & 0.424& 0.395 & 0.413\\
          % & 336   & \boldres{0.411} & \secondres{0.430} & 0.447 & 0.453 & 0.495 & 0.445 & 0.503 & 0.456 & 0.675 & 0.563 & 0.576 & 0.467 & 0.447 & 0.453 \\
          % & 720   & \boldres{0.427} & 0.455 & 0.457 & 0.462 & 0.611 & 0.510 & 0.511 & 0.481 & 0.683 & 0.585 & 0.835 & 0.583 & 0.457 & 0.462\\
          %\rowcolor{tabhighlight}
          % & {\textbf{avg}}
          % yelong 6m
          & 0.344 & 0.382
         % yelong 50m
          & 0.339 & \secondres{0.370} 
          % yelong 110m
          & \boldres{0.330} & \boldres{0.366}
          % yelong 300m
          & 0.337& \secondres{0.370}
          % Moirai 
          & 0.359&0.377
          % TimesFM
          & 0.392&0.406          
          % Moment
          & 0.362&0.410
        % visionTS
          &\secondres{0.333} & 0.375
          %Chronos
          & 0.398&0.411
          %TimeMoE
          & 0.362&0.399
          %FEDformer
          %&0.437 &0.449
          %TimesNet
          %& 0.414&0.427
          %Dlinear
          & 0.456&0.445
          %PatchTST
          &\boldres{0.330}&0.379
          \\
    \midrule
    % \multirow{4}[0]{*}{ETTm1} & 96    & \secondres{0.309} & 0.357 & \boldres{0.281} & \boldres{0.341} & 0.380 & 0.361 & 0.361 & 0.370 & 0.654 & 0.527 & 0.457 & 0.403 & \boldres{0.281} & \boldres{0.341} \\
    %       & 192   & \secondres{0.346} & 0.381 & \boldres{0.305} & \boldres{0.358} & 0.412 & 0.383 & 0.414 & 0.405 & 0.662 & 0.532 & 0.530 & 0.450 & \boldres{0.305} & \boldres{0.358} \\
    %       & 336   & \secondres{0.373} & 0.408 & \boldres{0.369} & \secondres{0.395} & 0.436 & 0.400 & 0.445 & 0.429 & 0.672 & 0.537 & 0.577 & 0.481 & \boldres{0.369} & \secondres{0.395}\\
    %       & 720   & 0.475 & 0.477 & 0.469 & 0.472 & \secondres{0.462} & \boldres{0.420} & 0.512 & 0.471 & 0.692 & 0.551 & 0.660 & 0.526 & 0.469 & 0.472 \\
    %       \rowcolor{tabhighlight}
    %       & {\textbf{AVG}} & \secondres{0.376} & 0.405 & \boldres{0.356} & \secondres{0.391} & 0.422 & 0.391 & 0.433 & 0.418 & 0.670 & 0.536 & 0.555 & 0.465& \boldres{0.356} & \secondres{0.391}  \\
     \multirow{1}[1]{*}{ETTm1} 
          % & 192   & \secondres{0.388} & \secondres{0.412} & 0.395 & 0.413 & 0.434 & 0.415 & 0.465 & 0.434 & 0.688 & 0.560 & 0.502 & 0.424& 0.395 & 0.413\\
          % & 336   & \boldres{0.411} & \secondres{0.430} & 0.447 & 0.453 & 0.495 & 0.445 & 0.503 & 0.456 & 0.675 & 0.563 & 0.576 & 0.467 & 0.447 & 0.453 \\
          % & 720   & \boldres{0.427} & 0.455 & 0.457 & 0.462 & 0.611 & 0.510 & 0.511 & 0.481 & 0.683 & 0.585 & 0.835 & 0.583 & 0.457 & 0.462\\
         % \rowcolor{tabhighlight}
          % & {\textbf{avg}}
          % yelong 6m
          & 0.370 & 0.368
         % yelong 50m
          & 0.367 & 0.366 
          % yelong 110m
          & \boldres{0.347} & \boldres{0.356}
          % yelong 300m
          & \secondres{0.351}&\secondres{0.358}
          % Moirai 
          & 0.407&0.385
          % TimesFM
          & 0.433&0.418
          % Moment
          & 0.670&0.536
        % visionTS
          &0.374 & 0.372
          %Chronos
          & 0.555&0.465
          %TimeMoE
          &0.356 &0.392
          %FEDformer
          %& 0.448&0.452
          %TimesNet
          %& 0.400&0.406
          %Dlinear
          & 0.362&0.379
          %PatchTST
          &\secondres{0.351}&0.381
          \\
    \midrule
    % \multirow{4}[0]{*}{ETTm2} & 96    & \boldres{0.197} & 0.286 & \secondres{0.198} & 0.288 & 0.211 & 0.274 & 0.202 & \boldres{0.270} & 0.260 & 0.335 & \boldres{0.197} & 0.271 & \secondres{0.198} & 0.288 \\
    %       & 192   & \secondres{0.250} & 0.322 & \boldres{0.235} & \boldres{0.312} & 0.281 & 0.318 & 0.289 & 0.321 & 0.289 & 0.350 & \secondres{0.254} & \secondres{0.314} & \boldres{0.235} & \boldres{0.312} \\
    %       & 336   & 0.337 & 0.375 & \boldres{0.293} & \boldres{0.348} & 0.341 & 0.355 & 0.360 & 0.366 & 0.324 & 0.369 & \secondres{0.313} & 0.353 & \boldres{0.293} & \boldres{0.348} \\
    %       & 720  & 0.480 & 0.461 & 0.427 & 0.428 & \secondres{0.485} & 0.428 & 0.462 & 0.430 & \boldres{0.394} & 0.409 & 0.416 & 0.415 & 0.427 & 0.428 \\
    %       \rowcolor{tabhighlight}
    %       & {\textbf{AVG}} & 0.316 & 0.361 & \boldres{0.288} & 0.344 & 0.329 & 0.343 & 0.328 & 0.346 & 0.316 & 0.365 & \secondres{0.295} & \secondres{0.338} & \boldres{0.288} & 0.344 \\
     \multirow{1}[1]{*}{ETTm2} 
          % & 192   & \secondres{0.388} & \secondres{0.412} & 0.395 & 0.413 & 0.434 & 0.415 & 0.465 & 0.434 & 0.688 & 0.560 & 0.502 & 0.424& 0.395 & 0.413\\
          % & 336   & \boldres{0.411} & \secondres{0.430} & 0.447 & 0.453 & 0.495 & 0.445 & 0.503 & 0.456 & 0.675 & 0.563 & 0.576 & 0.467 & 0.447 & 0.453 \\
          % & 720   & \boldres{0.427} & 0.455 & 0.457 & 0.462 & 0.611 & 0.510 & 0.511 & 0.481 & 0.683 & 0.585 & 0.835 & 0.583 & 0.457 & 0.462\\
          %\rowcolor{tabhighlight}
          % & {\textbf{avg}}
          % yelong 6m
          & 0.250 & 0.302
         % yelong 50m
          & 0.250 & \secondres{0.298} 
          % yelong 110m
          & \boldres{0.246} & \boldres{0.296}
          % yelong 300m
          & \secondres{0.249}&\boldres{0.296}
          % Moirai 
          & 0.303&0.337
          % TimesFM
          & 0.328&0.346
          % Moment
          & 0.316&0.365
        % visionTS
          &0.282 & 0.321
          %Chronos
          & 0.295&0.338
          %TimeMoE
          & 0.288&0.344
          %FEDformer
          %&0.305 &0.349
          %TimesNet
          %& 0.391&0.333
          %Dlinear
          & 0.356&0.331
          %PatchTST
          &0.255&0.315
          \\
    \midrule
    % \multirow{4}[0]{*}{Weather} & 96    & \secondres{0.159} & \secondres{0.213} & \boldres{0.157} & \boldres{0.211} & 0.199 & \boldres{0.211} & - & - & 0.243 & 0.255 & 0.194 & 0.235 &\boldres{0.157} & \boldres{0.211}\\
    %       & 192   & 0.215 & 0.266 & \boldres{0.208} & \secondres{0.256} & 0.246 & \boldres{0.251} & - & - & 0.278 & 0.329 & 0.249 & 0.285 & \boldres{0.208} & \secondres{0.256} \\
    %       & 336   & 0.291 & 0.322 & \boldres{0.255} & \boldres{0.290} & 0.286 & \secondres{0.291} & - & - & 0.306 & 0.346 & 0.302 & 0.327 & \boldres{0.255} & \boldres{0.290} \\
    %       & 720   & 0.415 & 0.400 & 0.405 & 0.397 & 0.373 & \secondres{0.354} & - & - & \secondres{0.350} & 0.374 & 0.372 & 0.378 & 0.405 & 0.397 \\
    %       \rowcolor{tabhighlight}
    %       & {\textbf{AVG}} & 0.270 & 0.300 & \boldres{0.256} & 0.288 & \secondres{0.264} & \boldres{0.273} & - & - & 0.294 & 0.326 & 0.279 & 0.306& \boldres{0.256} & 0.288  \\
     \multirow{1}[1]{*}{Weather} 
          % & 192   & \secondres{0.388} & \secondres{0.412} & 0.395 & 0.413 & 0.434 & 0.415 & 0.465 & 0.434 & 0.688 & 0.560 & 0.502 & 0.424& 0.395 & 0.413\\
          % & 336   & \boldres{0.411} & \secondres{0.430} & 0.447 & 0.453 & 0.495 & 0.445 & 0.503 & 0.456 & 0.675 & 0.563 & 0.576 & 0.467 & 0.447 & 0.453 \\
          % & 720   & \boldres{0.427} & 0.455 & 0.457 & 0.462 & 0.611 & 0.510 & 0.511 & 0.481 & 0.683 & 0.585 & 0.835 & 0.583 & 0.457 & 0.462\\
          %\rowcolor{tabhighlight}
          % & {\textbf{avg}}
          % yelong 6m
          & 0.239 & 0.268
         % yelong 50m
          & 0.231 & 0.257 
          % yelong 110m
          & 0.227 & \secondres{0.255}
          % yelong 300m
          & \boldres{0.217}&\boldres{0.245}
          % Moirai 
          &0.264 &0.273
          % TimesFM
          & - & -
          % Moment
          & 0.294&0.326
        % visionTS
          &0.269 & 0.292
          %Chronos
          & 0.279&0.306
          %TimeMoE
          & 0.256&0.289
          %FEDformer
          %&0.309 & 0.360
          %TimesNet
          %& 0.259&0.287
          %Dlinear
          & 0.240&0.300
          %PatchTST
          &\secondres{0.226}&0.264
          \\
    \midrule
    % \rowc\rowcolor{blue!15}
    % \multicolumn{2}{c|}{\scalebox{1.1}
    % {\textbf{Average}}} & \secondres{0.336} & 0.380 & \boldres{0.322} & \secondres{0.372} & 0.359 & 0.373 & 0.396 & 0.413 & 0.461 & 0.454 & 0.416 & 0.405 & \secondres{0.336} & 0.380 \\
    % \midrule
    \rowc
    \multicolumn{1}{c|}{\bf Rank}
    & 5.45 & 5.00
    & 4.75 & 2.90
    & \boldres{2.30} & \boldres{1.60}
    & \secondres{2.40} & \secondres{1.65}
    & 9.25 & 6.90 
    & 11.50 & 10.50 
    & 11.80 & 12.35
    & 5.45 &6.25
    & 11.20 & 10.65
    & 5.85 & 7,9
    %&11.70 & 12.55
    %& 10.15 &9.80
    &7.45 & 9.15
    & 4.20 & 6.25
    \\
    \bottomrule
    \end{tabular}%
    % }
  }
  \label{tab:zero_shot_short:001}%
  \vskip -0.2in
\end{table*}

\vskip -0.05in
\begin{table}[ht]
\centering
\caption{Results on GIFT-Eval (Best: \textcolor{red}{\bf Red}, Second: \textcolor{blue}{Blue})}
\label{tab:gift_eval:main}
\resizebox{1\textwidth}{!}{ % Resize table to fit the width of the text
\setlength{\tabcolsep}{6pt} % Adjust cell padding
\renewcommand{\arraystretch}{1.2} % Increase row height
\begin{tabular}{@{}>{\columncolor[gray]{0.9}}lccc|>{\columncolor[gray]{0.95}}cccc|>{\columncolor[gray]{0.9}}cccc@{}}
\toprule
\cmidrule(lr){1-4}\cmidrule(lr){5-8}\cmidrule(lr){9-12}
Model & MASE & CRPS & Rank & Model & MASE & CRPS & Rank & Model & MASE & CRPS & Rank\\
\midrule
% \multicolumn{4}{c}{\cellcolor[gray]{0.8}\textbf{Foundation Models}}& 
% \multicolumn{4}{|c|}{\cellcolor[gray]{0.8}\textbf{End-to-end Models}} & 
% \multicolumn{4}{c}{\cellcolor[gray]{0.8}\textbf{Statistical Models}} \\
\multicolumn{4}{c}{\cellcolor[gray]{0.8}\textbf{Foundation Models}}& 
\multicolumn{4}{|c|}{\cellcolor[gray]{0.8}\textbf{Foundation Models}} & 
\multicolumn{4}{c}{\cellcolor[gray]{0.8}\textbf{End-to-end Models}} \\
\textsc{YINGLONG}$_{\textrm{300m}}$ & \textcolor{blue}{0.716} & \textcolor{red}{\bf0.463} & \textcolor{red}{\bf7.38} 
&Chronos$_{\textrm{base}}$  & 0.786 & 0.550 & 14.84 &Crossformer & 2.310 & 1.383 &21.47\\
% & DeepAR & 1.206 & 0.721 &19.43\\

\textsc{YINGLONG}$_{\textrm{110m}}$ & 0.726 & 0.471 & \textcolor{blue}{8.04} 
&TimesFM                    & 0.967 & 0.575 & 15.09 & N-BEATS & 0.842 & 0.689 &21.92 \\

% & PatchTST & 0.762 & 0.496& 10.61 
% & Auto Arima & 0.964 & 0.770 &21.91\\
% 
% & Auto Theta & 0.978 & 1.051&24.34 \\

% &Moirai$_{\textrm{small}}$  & 0.849 & 0.549 & 14.17 

%  & Auto ETS & 1.088 & 6.327&25.21 \\

Chronos-bolt$_\textrm{base}$  & 0.725 & 0.485 & 8.61  
&Chronos$_{\textrm{small}}$ & 0.800 & 0.560 & 15.85 &Lag-Llama                  & 1.101 & 0.744 & 22.34 \\

\textsc{YINGLONG}$_{\textrm{50m}}$& 0.738 & 0.479 & 8.74 & VisionTS    & 0.775 & 0.638 & 21.02 &DLinear & 0.952 & 0.714 &23.56\\

TabPFN-TS &0.748 & 0.480 & 8.88 &TTMs                       & 0.969 & 0.753 & 24.04 &\multicolumn{4}{c}{\cellcolor[gray]{0.8}\textbf{Statistical Models}}\\

TimesFM2$_{500m}$ &\textcolor{red}{\bf0.680} & \textcolor{blue}{0.465} & 8.90 
&Timer                      & 1.019 & 0.820 & 25.67 & Auto Arima & 0.964 & 0.770 &21.91\\

% & &  &  &  \\
Chronos-bolt$_\textrm{small}$&0.738 & 0.487 & 9.36 &\multicolumn{4}{c}{\cellcolor[gray]{0.8}\textbf{End-to-end Models}}& Auto Theta & 0.978 & 1.051&24.34 \\
Moirai$_{\textrm{large}}$  & 0.785 & 0.506 & 10.71 & PatchTST & 0.762 & 0.496& 10.61  & Auto ETS & 1.088 & 6.327&25.21\\

% & Lag-Llama & 1.101 & 0.744 &22.34&  &  &  \\
 Moirai$_{\textrm{base}}$   & 0.809 & 0.515 & 10.83& iTransformer & 0.802 & 0.524 & 11.90 & Seasonal Naive & 1.00 & 1.00&26.36 \\
 \textsc{YINGLONG}$_{\textrm{6m}}$   & 0.790 & 0.515 & 12.55 & TFT & 0.822 & 0.511 &12.03& Naive & 1.260 & 1.383 &28.24\\
Moirai$_{\textrm{small}}$  & 0.849 & 0.549 & 14.17 & TIDE & 0.980 & 0.652&19.22 \\
Chronos$_{\textrm{large}}$ & 0.781 & 0.547 & 14.74 & DeepAR & 1.206 & 0.721 &19.43\\

%  &  &  &  &  &  & & & \\
%  &  &  &  &  &  & & & \\
%  &  &  &  &  &  & & & \\
% &  &  &  &  &  & & & \\
%  &  &  &  &  &  & & & \\
%  &  &  &  &  &  & & & \\
\bottomrule
\end{tabular}
} % End resizebox
\vskip -0.3in
\end{table}

\subsection{Generalization Across Diverse Datasets}

We recognize that ETT series and weather datasets may be insufficient for a truly comprehensive evaluation, which is a subject of ongoing debate in the time series community. Consequently, we adopt GIFT-Eval \cite{aksu2024giftevalbenchmarkgeneraltime} as a benchmark. It includes 23 datasets from domains such as economics, energy, healthcare, nature, sales, transportation, and cloud operations, and evaluates all major foundation time series models, end-to-end methods, and statistical techniques, mitigating concerns about generalization.

% We recognize that the ETT series and weather datasets alone are insufficient for a comprehensive evaluation of time series methods, which is a topic of ongoing debate. To address this, we adopt the guidelines from GIFT-Eval, a benchmark recently introduced at a top conference. GIFT-Eval includes 23 datasets spanning various domains such as economy, energy, healthcare, nature, sales, transportation, and cloud operations. It evaluates all foundational time series models, end-to-end methods, and statistical techniques, potentially easing concerns about generalization.

Our testing further incorporates five statistical approaches: Naive and Seasonal Naive \cite{hyndman2018forecasting}, Auto ARIMA, Auto ETS, and Auto Theta \cite{garza2022statsforecast}. For foundation models, we consider Chronos, Moirai, TimesFM, TTMS, VisionTS, Timer, Lag-Llama, and TabPFN-TS. We also benchmark end-to-end methods including PatchTST, TFT, Tide \cite{das2023long}, DeepAR \cite{salinas2020deepar}, Crossformer \cite{zhang2023crossformer}, N-BEATS, and DLinear \cite{zeng2023dlinear}

% In our testing, we include five statistical methods: Naive and Seasonal Naive from \cite{hyndman2018forecasting}, as well as Auto Arima, Auto ETS, and Auto Theta from \cite{garza2022statsforecast}. For foundational models, we consider Chronos, Moirai, TimesFM, TTMS, VisionTS \cite{chen2024visionts}, Timer, Lag-Llama, and TabPFN-TS \cite{hoo2025tabular}. Our end-to-end model benchmarks include PatchTST, TFT, Tide \cite{das2023long}, deepAR \cite{salinas2020deepar}, Crossformer \cite{zhang2023crossformer}, N-BEATS, and Dlinear \cite{zeng2023dlinear}. We maintain the same inference configuration for our model in point forecasting.

GIFT-Eval performance is reported in terms of mean absolute scaled error (MASE) and continuous
ranked probability score (CRPS). Table~\ref{tab:gift_eval:main} presents the geometric averages of these metrics. Notably, two of our models achieved top rankings, with average ranks of 7.38 and 8.04, respectively, highlighting their robustness and generalizability across diverse datasets. Furthermore, with respect to the MASE and CRPS metrics, our 300M model clearly outperforms the recent TabPFN-TS model, showing significant improvements of 4.3\% and 3.5\%, respectively. It's worth mentioning TabPFN-TS ranks at the top among all baselines, excluding our models, while our participation propels Chronos-Bolts ahead. Moreover, we observe scaling-law improvements from 50M to 300M parameters. Our largest 300M model is similar in size to other models like Morirai (311M), Moment (385M), and Chronos (710M), and is much smaller than Time-MoE as shown in table ~\ref{tab:gift_eval:main}. Our performance gains are thus not solely driven by parameter scaling. Detailed results can be found in the appendix, with data aggregated by domain, prediction length, frequency, and multivariate settings. %Our \method demonstrates robust performance across multiple categories, particularly in Energy, Nature, and Transportation. Appendix~\ref{app:GIFT_Eval_details} provides additional analyses, including breakdowns by prediction length, frequency, and multivariate settings. %In CloudOps and Sales, end-to-end methods outperform some foundation models, possibly due to richer within-channel interactions for these business-oriented domains. 

\vspace{-0.3cm}
\begin{figure*}[ht]
    \centering
    \includegraphics[width=0.325\textwidth]{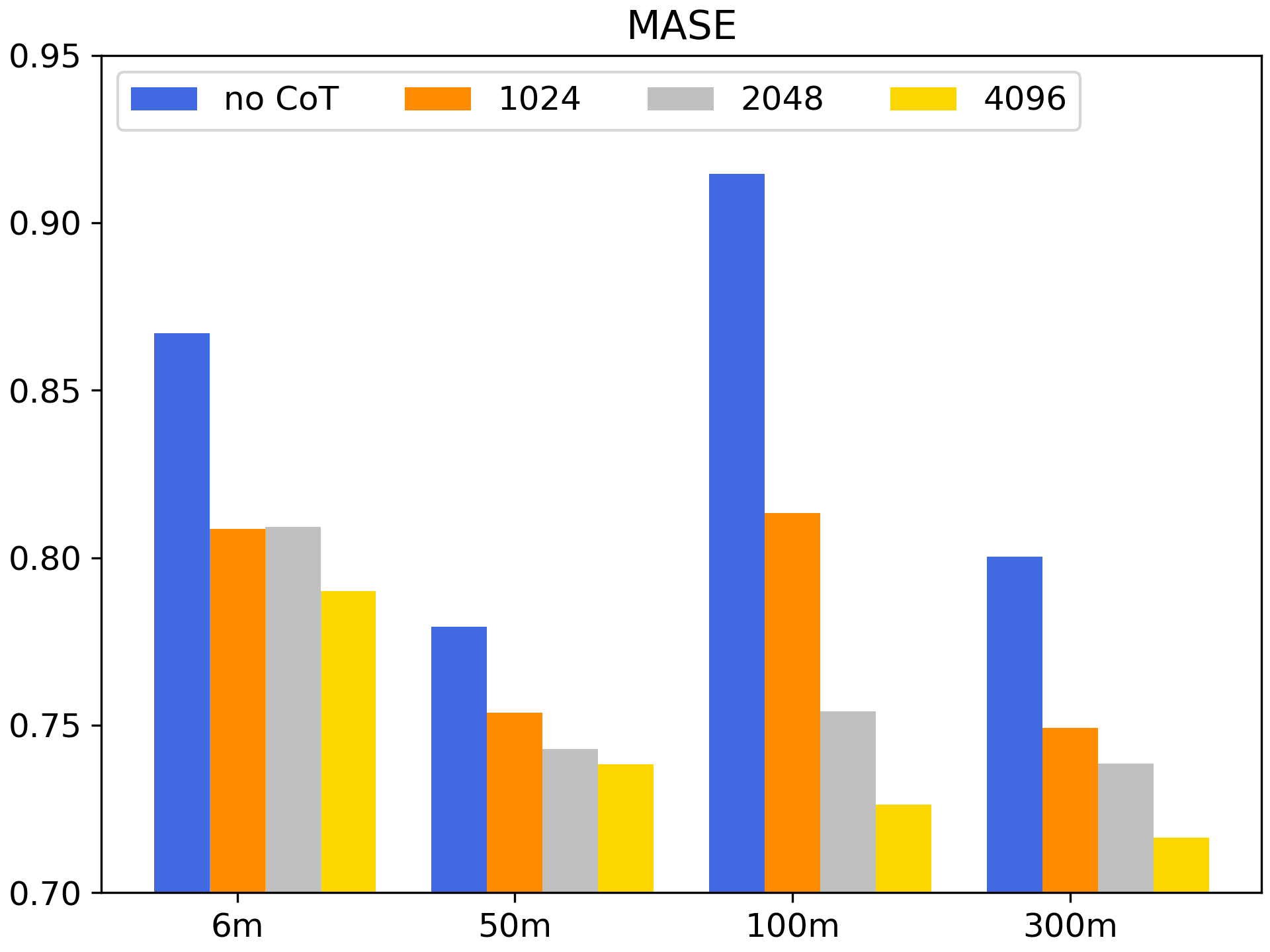}
    \includegraphics[width=0.325\textwidth]{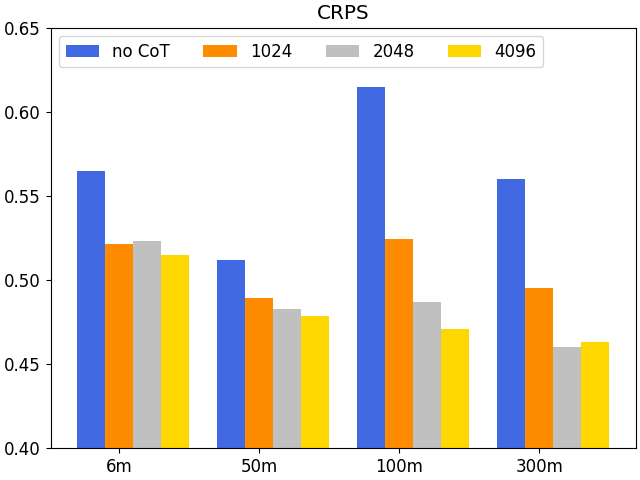}
    \includegraphics[width=0.325\textwidth]{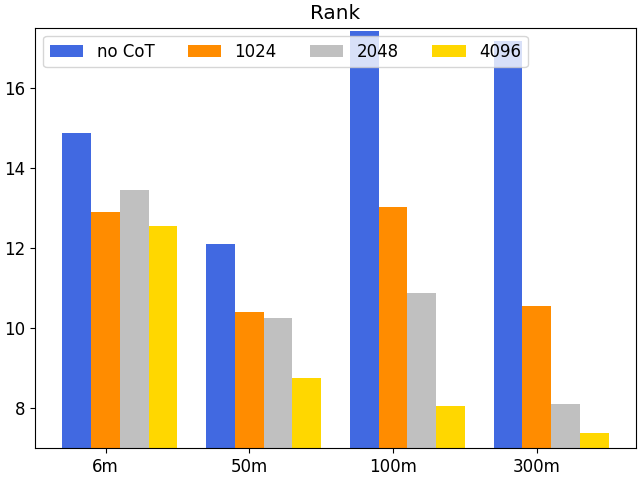}
    \caption{Ablation on DCoT. Four different output lengths up to 4096 are considered. In no CoT setting, we only output the target length. The MASE, CRPS,RANK of Gift-Eval are reported.}
    \label{fig:ablation:DCOT}
    \vspace{-0.6cm}
\end{figure*}

%\subsection{IN-DISTRIBUTION FORECASTING}
%\input{tables/in_distribution_full}

%\subsection{Ablation Study}
\vspace{-0.2cm}
\subsection{Influence of DCoT}
\vspace{-0.2cm}
We perform ablation studies to demonstrate the effectiveness of DCoT by varying its length, as illustrated in Figure~\ref{fig:ablation:DCOT}. DCoT clearly demonstrates the phenomenon of inference scaling, resulting in substantial performance improvements, including reductions of up to 10.1\% in MASE and 11.9\% in CRPS. These advancements are likely the primary contributors to YinLong's success.

% We now perform ablation experiments to demonstrate the effectiveness of DCoT, considering various DCoT lengths as shown in Figure~\ref{fig:ablation:DCoT}.This effect significantly enhances our performance, acting as the primary driver behind YinLong's success.

% \begin{enumerate}[leftmargin=*,label=\textbf{\arabic*.}]
     \textbf{General scaling observations}\ 
    We observe substantial improvements across all pretrained  \textsc{YINGLONG} models (6M--300M). Interestingly, the no-COT baseline model does not strictly follow the usual scaling trends; for example, the 50M model outperforms its 6M, 100M, and 300M counterparts in some cases. This behavior is not uncommon in time series foundation models, where scaling laws are less consistent than in large language models (LLMs). However, by integrating a sufficiently long DCoT and leveraging the output scaling effect, we achieve a more robust scaling relationship. For instance, using \textsc{YINGLONG}$_{300\mathrm{m}}$ with a 4096-length DCoT reduces MASE by 10.5\% (from 0.800 to 0.716),and improves its average rank on GIFT-Eval from 17.2 to 7.38. By contrast, \textsc{YINGLONG}$_{100\mathrm{m}}$ with the same DCoT length yields only a 5.3\% MASE drop (from 0.779 to 0.738). These findings suggest that DCoT has a stronger impact on larger, more robust baseline models. 
    
    \textbf{Output Scaling for Other Model}: We investigated the output scaling effect in vanilla transformer models ranging from 6M to 300M parameters using our joint forecasting paradigm (see Appendix Figure~\ref{fig:ablation:DCOT_transformer}). Applying DCoT with a token length of 4096 significantly improved the 300M model's MASE and CRPS by 24.9\% and 30.0\%, respectively, compared to a non-DCoT setup. This scaling effect persisted across DCOT lengths and various model sizes. validated on the GIFT-Eval benchmark with 23 datasets, these results demonstrate that the scaling effect is robust across different model architectures and datasets.%This effect clearly demonstrates strong output scaling within the large GIFT-Eval benchmark with 23 datasets, indicating that it is neither unique to a particular architectural design nor restricted to a specific dataset. 

     \textbf{Effect of DCoT length.}
    As DCoT length grows from 1k to 4k, the performance gains become more pronounced and consistent. For instance, \textsc{YINGLONG}$_{300\mathrm{m}}$ achieves MASE reductions of 6.4\% and 7.7\% with DCoT lengths of 1024 and 2048, respectively, compared to a 10.5\% reduction at 4096 length. This trend is evident across various model sizes, indicating that for a fixed test horizon, extending the output sequence enhances accuracy. This improvement is attributed to delayed CoT interactions affecting earlier outputs, a phenomenon specific to our joint forecasting paradigm. In contrast, direct/recursive forecasting do not exhibit this effect, as extending the output sequence does not impact prior outputs.
% \end{enumerate}

\textbf{Structural Ablation} We conducted a structure ablation on the uTransformer block and the token merge design, as shown in the Appendix Figure~\ref{fig:structure_ablation}, resulting in moderate performance improvements over the standard transformer block. We compared the  \textsc{YINGLONG}-300M model to vanilla transformer models ranging from 6M to 300M parameters, all trained using our joint forecasting paradigm (see Appendix Figure~\ref{fig:ablation:DCOT_transformer}).  \textsc{YINGLONG}-300M reduced the MASE from 0.726 to 0.716 and the CRPS from 0.473 to 0.463 compared to the 300M transformer baseline. Even vanilla transformer with joint forecasting performs comparably to or better than previous SOTA foundation models, including TabPFN-TS (MASE=0.748, CRPS=0.480) and Chronos-bolt (MASE=0.725, CRPS=0.471).

\textbf{Error Reduction Pattern} Our analysis reveals that the primary error reduction achieved by DCoT is due to a decrease in trend error. As shown in Appendix Table \ref{tab:dcot_performance} and Figure \ref{fig:mse_reduction_side_by_side}, both absolute and relative reductions in MSE are attributable to trend rather than changes in the seasonal component.

\textbf{Influence of Input Ensemble} We compare the average of multiple input-length forecasts with the single-best component in Table~\ref{tab:ablation:input_length} in Appendix. Our input-ensemble strategy improves accuracy by 1\%--4\% without training additional models. This represents a scalable “free lunch” approach. 
%Unlike more complex approaches for LLMs, which require multiple runs and extensive NLP evaluations, our method achieves enhancements with minimal additional effort.
% We compared average results with the best single forecast components (see Table~\ref{tab:ablation:input_length}). Generally, longer inputs enhance testing outcomes, aligning with existing studies, but sometimes impair performance.Our ensemble method for input demonstrates a robust improvement of 1\%-4\% without the need for training additional models, in contrast to NBEATS~\cite{oreshkin2020nbeats}. It is a scalable "free lunch" solution, offering gains via straightforward post-training ensemble. Unlike LLM's complex multiple-run methods needing extensive evaluations for NLP, our strategy enhances performance with simple averaging., providing post-training gains through simple ensemble averaging.

% \subsection
% \subsection{MODEL SCALABILITY}

% \subsection{DATA SCALABILITY}
% \subsubsection{INPUT SCALABILITY}
% \subsubsection{OUTPUT SCALABILITY}
\vspace{-0.3cm}
\section{Conclusion and Future Work}
\vspace{-0.2cm}
% Our study introduces an encoder-only transformer for time series forecasting using non-causal, bidirectional attention, diverging from typical autoregressive models. By extending outputs, we enhance model accuracy through delayed COT reasoning. A post-training ensemble framework was proposed to enhance performance.Scaled from 6M to 300M parameters, our models excel in zero-shot tasks.GIFT-Eval benchmark evaluations demonstrate that YinLong surpasses existing models across various domains.
Our study presents a novel joint forecasting paradigm for time series forecasting, distinguishing itself from traditional direct and recursive approaches by incorporating masked patch placeholders and non-causal, bidirectional attention mechanisms. This approach uncovers a previously unreported scaling phenomenon: longer outputs appear to enhance model accuracy due to a delayed chain-of-thought reasoning process. Comprehensive evaluations using the GIFT-Eval benchmark confirm that YingLong consistently delivers superior results across various metrics, outperforming SOTA methods. Compared to LLMs, we face interpretability challenges: LLMs make reasoning explicit through textual tokens, while our delayed thought process remains hidden in latent space. Future work will focus on better interpreting and understanding COT within time series models.

% In conclusion, we introduce a non-autoregressive, encoder-only transformer architecture for time series forecasting that leverages non-causal, bidirectional attention. This approach—akin to language understanding rather than generation—reveals a novel scaling effect: longer outputs boost accuracy via a delayed chain-of-thought. By applying a post-training ensemble, our models (6M--300M parameters) achieve significant performance gains, excelling in zero-shot tasks. On the GIFT-Eval benchmark, YinLong consistently outperforms existing foundation models, end-to-end methods, and statistical baselines.

% Compared to LLMs, we face interpretability challenges: LLMs make reasoning explicit through textual tokens, while our delayed thought process remains hidden in latent space. Future work will focus on better interpreting and understanding chain-of-thought within time series models.

% \newpage
% \input{sections/7_impact_statement}


\bibliography{main.bib}
\bibliographystyle{icml2025}

%%%%%%%%%%%%%%%%%%%%%%%%%%%%%%%%%%%%%%%%%%%%%%%%%%%%%%%%%%%%%%%%%%%%%%%%%%%%%%%
%%%%%%%%%%%%%%%%%%%%%%%%%%%%%%%%%%%%%%%%%%%%%%%%%%%%%%%%%%%%%%%%%%%%%%%%%%%%%%%
% APPENDIX
%%%%%%%%%%%%%%%%%%%%%%%%%%%%%%%%%%%%%%%%%%%%%%%%%%%%%%%%%%%%%%%%%%%%%%%%%%%%%%%
%%%%%%%%%%%%%%%%%%%%%%%%%%%%%%%%%%%%%%%%%%%%%%%%%%%%%%%%%%%%%%%%%%%%%%%%%%%%%%%
\newpage
\appendix
\onecolumn
% \section{Appendix}
\section{Proof of Lemma~\ref{lem:1}}\label{proof:lem:1}
\begin{proof}
    Since the sample paths of the random walk are martingale and generated independently, we have
    \begin{equation*}
        \mathbbm{E}[x_{m+j}^i] = x_{m}, \quad \mathbbm{E}[(x_{m+j}^i - x_{m})^2] = j.
    \end{equation*}
    Denote
    \begin{equation*}
        Z_j = \left(\frac{1}{N}\sum_{i=1}^N x_{m+j}^i - x_m\right)^2.
    \end{equation*}
    It follows that
    \begin{equation*}
        Z_j \ge 0, \quad \mathbbm{E}[Z_j] = \frac{j}{N}, \quad \mathbbm{E}[Z_j^2] \le \mathcal{O}(j^2 N^{-2}).
    \end{equation*}
    Via the Paley–Zygmund inequality, for $\epsilon \in (0,1)$ we have
    \begin{equation*}
        \mathbbm{P}(Z_j \ge \epsilon \mathbbm{E}[Z_j]) \ge (1-\epsilon)^2 \cdot \frac{\mathbbm{E}[Z_j]^2}{\mathbbm{E}[Z_j^2]}.
    \end{equation*}
    Thus,
    \begin{equation*}
        \mathbbm{P}\left(\left|\frac{1}{N}\sum_{i=1}^N x_{m+j}^i - x_{m}\right| \ge \tilde{\epsilon}\right) \ge C\left(1-\frac{\tilde{\epsilon}^2 N}{j}\right)^2,
    \end{equation*}
    where $C$ is a positive constant and $\tilde{\epsilon} = \sqrt{\epsilon j / N}$.
\end{proof}

\section{Training Details}\label{sec:training}
We summarize the details for pretraining datasets in Table~\ref{tab:train_Dataset}. The weights in \eqref{eq:loss_g} is set as $w_{\alpha_k,x^i} = 1/(\sqrt{\alpha_k(1-\alpha_k)}\sum|x^{i,s}|)$, where we control the influence on the target scale and quantile estimation variance simultaneously. In Table~\ref{tab:model_details} we present the model configurations. In order minimizing the influence of hyperparameter tuning, we keep all training settings the same except model size related ones.

\section{Supplement for  zero-shot forecasting experiment}

Table \ref{tab:dataset_long} summarize the statistics of datasets. We use Mean Absolute Error (MAE) and Mean Squared Error (MSE) as metrics. The full results are provided in Table~\ref{tab:zero_shot_full}. In particular,\textsc{YingLong}$_{\mathrm{100m}}$ and \textsc{YingLong}$_{\mathrm{300m}}$ reach 20 and 16 best results. Those performances beat other foundation models. 

% \section{Detailed GIFT-EVAL Benchmark Result}

\section{Supplement for the GIFT-Eval Benchmark}

For each GIFT-Eval experiment, we measure performance using mean absolute scaled error (MASE) and weighted quantile loss (WQL). Table~\ref{tab:gift_eval:main} presents the geometric average of these metrics. Notably, three of our models achieve top scores in MASE, WQL, and Rank, demonstrating their robustness and generalizability across the diverse datasets.Furthermore, we observe scaling law improvements from 50M to 300M. Our largest 300M model is comparable in size to Morirai 311M, Moment 385M, and Chronos 710M, yet significantly smaller than Time-MoE. Our improvements do not rely on increasing parameter size through scaling laws. A more detailed performance analysis, aggregated by domain, is presented in Table~\ref{Tab:GIFT-Eval domain}. YinLong demonstrates robust performance across various domains, exhibiting leading results in Energy, Nature, and Transportation, with rank improvements from 9.2 to 4.3, 4.6 to 3.8, and 6.9 to 5.2, respectively. Additionally, YinLong achieves top performance in CloudOps and Sales. Interestingly, in business-heavy domains such as CloudOps and Sales, the end-to-end model outperforms the time series foundation model. This may be attributed to their extensive within-channel interactions.

We provide a summary on the datasets of GIFT-Eval in Table~\ref{tab:gift-eval-dataset}. We consider mean absolute scaled error (MASE), continuous ranked probability score (CRPS) and Rank as metrics.
\begin{align}
    \mathrm{MASE}=\frac{m-s}{n}\cdot \frac{\sum_{t=m+1}^{m+n}|\hat{x}_t-x_t|}{\sum_{t}^{m-s}|x_{t}-x_{t+s}|},\notag
\end{align}
where $m$ in the lookback length, $n$ is the forecasting length, and $s$ is the seasonality parameter.

In this work, we follow the setting in \cite{aksu2024giftevalbenchmarkgeneraltime} and use weighted quantile loss (WQL) to approximate CRPS as follows
\begin{align}
    \mathrm{CRPS} = \frac{1}{K}\sum_{i=1}^K\mathrm{WQL}[\alpha_k]\notag\\
    \mathrm{WQL}[\alpha] = 2\frac{\sum_{t=m+1}^{m+n}l_{\alpha}(q_t(\alpha),x_t)}{\sum_{t=m+1}^{m+n}|x_t|}\notag,
\end{align}
where we set $K=9$ and $alpha_k = k/10$ for $k=1,2,...,K$.

Readers may refer \citealt{aksu2024giftevalbenchmarkgeneraltime} for more details. The addition results aggregated in  prediction lengths, frequency and number of variates are given in Table~\ref{tab:gift:length}, \ref{tab:gift:freq} and  \ref{tab:gift:nvariate}.
% \subsection{GIFT-Eval datasets}
% \subsection{GIFT-Eval datasets}
\begin{table*}[h]
\centering
\caption{Pretraining Datasets}
\resizebox{0.7\columnwidth}{!}{
\begin{tabular}{ccccr}
\toprule
Dataset&Source &domain &Frequency & \# Length\\
\midrule
Australian Electricity Demand&\citealt{godahewa2021monash}&Energy &30T&1153584\\
Bitcoin&\citealt{godahewa2021monash}&Econ/Fin &D&68927\\
Cif 2016&\citealt{godahewa2021monash}&Econ/Fin &M&7108\\
Cif 2016&\citealt{godahewa2021monash}&Econ/Fin &M&7108\\
Fred MD&\citealt{godahewa2021monash}&Econ/Fin&M&71624\\
Kaggle Web Traffic Daily&\citealt{godahewa2021monash}&Web/CloudOps&W&15206232\\
Kaggle Web Traffic Weekly&\citealt{godahewa2021monash}&Web/CloudOps&D&332586145\\
London smart meters&\citealt{godahewa2021monash}&Energy&30T&160041727\\
M1 Monthly&\citealt{godahewa2021monash} &Econ/Fin &M &1047\\
M1 Quarterly &\citealt{godahewa2021monash}&Econ/Fin&3M &9628\\
M1 Yearly&\citealt{godahewa2021monash}&Econ/Fin&Y & 3136 \\
M3 Monthly &\citealt{godahewa2021monash}&Econ/Fin&M &538\\
M3 Quarterly&\citealt{godahewa2021monash} &Econ/Fin& 3M & 36960\\
M3 Yearly&\citealt{godahewa2021monash} &Econ/Fin &Y &18319\\
NN5 Daily&\citealt{godahewa2021monash}&Econ/Fin &D &35303\\
Pedestrian Counts&\citealt{godahewa2021monash} &Transport &H &3125914 \\
Sunspot &\citealt{godahewa2021monash}&Nature &D &45312\\
Tourism Monthly&\citealt{godahewa2021monash} &Econ/Fin &M & 98867 \\
Tourism Quarterly&\citealt{godahewa2021monash}& Econ/Fin& Q& 39128 \\
Tourism Yearly &\citealt{godahewa2021monash}&Econ/Fin &Y & 11198\\
Traffic Hourly &\citealt{godahewa2021monash}&Transport &H & 14858016\\
Traffic Weekly &\citealt{godahewa2021monash}&Transport &W & 78816\\
Oikolab Weather&\citealt{godahewa2021monash} &Nature &H &615574\\
Wind Power &\citealt{godahewa2021monash}&Energy &4s&7397147\\
Wind Farm &\citealt{godahewa2021monash}&Energy&T&172165370\\
M5 & \citealt{m5-forecasting-accuracy}&Sales&D&58327370\\
Wiki Daily & \citealt{ansari2024chronos}&Web/CloudOps&D&247099892\\
USHCN& \citealt{ansari2024chronos}&Nature&D&235396770\\
Uber TLC Daily &\citealt{JMLR:v21:19-820}&Transport &D &42533\\
Uber TLC Hourly &\citealt{JMLR:v21:19-820}&Transport &H& 510284\\
Weatherbench Hourly &\citealt{rasp2020weatherbench}&Nature &H & 74630250518 \\
Weatherbench Daily &\citealt{rasp2020weatherbench}&Nature &D & 3223513345 \\
Weatherbench Weekly &\citealt{rasp2020weatherbench}&Nature &W &462956049\\
KernelSynth &\citealt{ansari2024chronos}&Synthetic&-&3221225472\\
% \midrule
% &&&Total&\\
% weatherbench
% exchange_rat
% tsmixup_10m
% weatherbench
% weatherbench
% BizITObs&\citealt{palaskar2023automixer}&Web/CloudOps&10S,5T,H&24&12\\
% Bitbrains&\citealt{shen2015statistical}&Web/CloudOps&5T,H&1750&8\\
% Restaurant&\citealt{recruit-restaurant-visitor-forecasting}&Sales&D&807&1\\
% ETT&\citealt{zhou2021informer}&Energy&15T,H,D,W-THU&2&18\\
% libcity&\citealt{libcity}&Transport&5T,H,D,15T,&518&14\\
% Solar&\citealt{lai2018modeling}&Energy&10T,H,D,W-FRI&137&8\\
% Hierarchical Sales&\citealt{mancuso2021machine}&Salses&D,W-WED&118&2\\
% M4&\citealt{godahewa2021monash}&Econ/Fin& A-DEC,Q-DEC,M,W-SUN,D,H&100741&6\\
% Hospital & \citealt{godahewa2021monash}&Healthcare &M &767&1\\
% COVID Death &\citealt{godahewa2021monash}&Healthcare &D &226&1\\
% US Births& \citealt{godahewa2021monash}&Healthcare &D,W-TUE,M&1&3\\
% Saugeen&\citealt{godahewa2021monash}&Nature&D,W-THU,M&1&3\\
% Temperature Rain&\citealt{godahewa2021monash}&Nature&D&32072&1\\
% KDD CUP 2018&\citealt{godahewa2021monash}&Nature&H,D&270&4\\
% Car Parts&\citealt{godahewa2021monash}&Sales&M&2674&1\\
% Electricity&\citealt{godahewa2021monash}&Energy&15T,H,D,W-FRI&370&8\\
\bottomrule
\end{tabular}\label{tab:train_Dataset}
}
\end{table*}

% \subsection{GIFT-Eval datasets}
\begin{table*}[h]
\centering
\caption{Model Configurations}
\resizebox{0.7\columnwidth}{!}{
\begin{tabular}{c|cccccccccc}
\toprule
Config&\textsc{YingLong}$_{\mathrm{6m}}$&\textsc{YingLong}$_{\mathrm{50m}}$&\textsc{YingLong}$_{\mathrm{100m}}$&\textsc{YingLong}$_{\mathrm{300m}}$\\
\midrule
Max Length& 8192&8192&8192&8192\\
Layers & 6&8&12&18\\
Heads &16&16&12&32\\
Embed Dim&256&512&768&1024\\
Intermediate Dim&1024&2048&3072&4096\\
Query per Groups&4&4&4&4\\
Pos Embed&Rope&Rope&Rope&Rope\\
Norm&RMSNorm&RMSNorm&RMSNorm&RMSNorm\\
MLP&SwiGLU &SwiGLU&SwiGLU&SwiGLU\\
Patch Size&32&32&32&32\\
Batch Size &512&512&512&512\\
Max Step &100000&100000&100000&100000\\
Warm-up Steps &2000 &2000 &2000 &2000\\
Learning Rate &1e-4&1e-4&1e-4&1e-4\\
Scheduler &Cosine&Cosine&Cosine&Cosine\\
Weight Decay &0.1&0.1&0.1&0.1\\
$\beta_1, \beta_2$ & 0.9, 0.95& 0.9, 0.95& 0.9, 0.95& 0.9, 0.95\\
Min LR &1e-5&1e-5&1e-5&1e-5\\
Mix Up Aug&0.2&0.2&0.2&0.2\\
Rescaling Aug&(1,4)&(1,4)&(1,4)&(1,4)\\
mask ratio $\rho$&0.2&0.2&0.2&0.2\\
% \midrule
% &&&Total&\\
% weatherbench
% exchange_rat
% tsmixup_10m
% weatherbench
% weatherbench
% BizITObs&\citealt{palaskar2023automixer}&Web/CloudOps&10S,5T,H&24&12\\
% Bitbrains&\citealt{shen2015statistical}&Web/CloudOps&5T,H&1750&8\\
% Restaurant&\citealt{recruit-restaurant-visitor-forecasting}&Sales&D&807&1\\
% ETT&\citealt{zhou2021informer}&Energy&15T,H,D,W-THU&2&18\\
% libcity&\citealt{libcity}&Transport&5T,H,D,15T,&518&14\\
% Solar&\citealt{lai2018modeling}&Energy&10T,H,D,W-FRI&137&8\\
% Hierarchical Sales&\citealt{mancuso2021machine}&Salses&D,W-WED&118&2\\
% M4&\citealt{godahewa2021monash}&Econ/Fin& A-DEC,Q-DEC,M,W-SUN,D,H&100741&6\\
% Hospital & \citealt{godahewa2021monash}&Healthcare &M &767&1\\
% COVID Death &\citealt{godahewa2021monash}&Healthcare &D &226&1\\
% US Births& \citealt{godahewa2021monash}&Healthcare &D,W-TUE,M&1&3\\
% Saugeen&\citealt{godahewa2021monash}&Nature&D,W-THU,M&1&3\\
% Temperature Rain&\citealt{godahewa2021monash}&Nature&D&32072&1\\
% KDD CUP 2018&\citealt{godahewa2021monash}&Nature&H,D&270&4\\
% Car Parts&\citealt{godahewa2021monash}&Sales&M&2674&1\\
% Electricity&\citealt{godahewa2021monash}&Energy&15T,H,D,W-FRI&370&8\\
\bottomrule
\end{tabular}\label{tab:model_details}
}
\end{table*}

\begin{table}[h]
\caption{Dataset details for zero-shot forecasting experiment}
\label{tab:dataset_long}
% \vskip 0.15in
\begin{center}
\begin{small}
\begin{tabular}{lccccc}
\toprule
% Dataset & num & dim & freq \\
Dataset &Source& Length & Dimension & Frequency \\
\midrule
ETTm1 &\citealt{zhou2021informer}& 69680 & 7 & 15 T\\
ETTm2 &\citealt{zhou2021informer}& 69680 & 7 & 15 T\\
ETTh1 &\citealt{zhou2021informer}& 17420 & 7 & 1H\\
ETTh2 &\citealt{zhou2021informer}& 17420 & 7 & 1H\\
Weather &\citealt{wu2021autoformer}& 52696 & 21 & 10T \\
% Electricity & 26304 & 321 & 1 hour \\
% Traffic & 17544 & 862 & 1 hour \\
\bottomrule
\end{tabular}
\end{small}
\end{center}
\vskip -0.1in
\end{table}
% \vspace{-3mm}

\begin{table*}[htbp]
  \centering
  \caption{Full results of zero-shot forecasting experiments. A lower MSE or MAE indicates a better prediction. TimesFM, due to its use of Weather datasets in pretraining, is not evaluated on this dataset and is denoted by a dash ($-$). {\boldres{Red}}: the best, \secondres{Blue}: the 2nd best.} %The below table represents these results by a dash ($-$).}
  \resizebox{1.1\columnwidth}{!}{
    % \renewcommand{\tabcolsep}{3pt}
    % \scalebox{0.5}{
    \begin{tabular}{cr|cc|cc|cc|cc|cc|cc|cc|cc|cc|cc|cc|cc|cc|cc}
          \toprule
          \multicolumn{2}{c}{\multirow{3}{*}{\textbf{\scalebox{1.2}{Models}}}} & \multicolumn{8}{c}{Ours} & \multicolumn{12}{c}{Foundation Models}&\multicolumn{8}{c}{End-to-end Models} \\
          \cmidrule(lr){3-10} \cmidrule(lr){11-22} \cmidrule(lr){23-30}
          &       
          & \multicolumn{2}{c}{\textsc{YINGLong}$_{6m}$} 
          & \multicolumn{2}{c}{\textsc{YINGLong}$_{50m}$} 
          & \multicolumn{2}{c}{\textsc{YINGLong}$_{110m}$} 
          & \multicolumn{2}{c}{\textsc{YINGLong}$_{300m}$} 
          & \multicolumn{2}{c}{Moirai}
          & \multicolumn{2}{c}{TimesFM}
          & \multicolumn{2}{c}{Moment}
          & \multicolumn{2}{c}{visionTS}
          & \multicolumn{2}{c}{Chronos}
          & \multicolumn{2}{c}{TimeMoE}
         & \multicolumn{2}{c}{FEDformer}
       & \multicolumn{2}{c}{TimesNet}
          & \multicolumn{2}{c}{Dlinear}
             & \multicolumn{2}{c}{PatchTST}
          \\
          % \cmidrule(lr){3-4} \cmidrule(lr){5-6}\cmidrule(lr){7-8} \cmidrule(lr){9-10}\cmidrule(lr){11-12}\cmidrule(lr){13-14}\cmidrule(lr){15-16}
          \multicolumn{2}{c}{\scalebox{1.2}{\textbf{Metrics}}}&  \textbf{MSE} & \textbf{MAE} & \textbf{MSE} & \textbf{MAE} & \textbf{MSE} & \textbf{MAE} & \textbf{MSE} & \textbf{MAE} & \textbf{MSE} & \textbf{MAE} & \textbf{MSE} & \textbf{MAE}&
          \textbf{MSE} & \textbf{MAE} & \textbf{MSE} & \textbf{MAE}& 
          \textbf{MSE} & \textbf{MAE} & \textbf{MSE} & \textbf{MAE}& 
          \textbf{MSE} & \textbf{MAE} & \textbf{MSE} & \textbf{MAE}& 
          \textbf{MSE} & \textbf{MAE} & \textbf{MAE} & \textbf{MAE} \\
          \midrule

          \multirow{4}[1]{*}{ETTh1} & 96    
          % yelong 6m
          & 0.366 & 0.380 
          % yelong 50m
          & 0.359 & 0.375
          % yelong 110m
          & 0.354 & \boldres{0.372}
          % yelong 300m
          & 0.355&\secondres{0.374}
          % Moirai 
          & 0.376&0.388
          % TimesFM
          & 0.414&0.404
          % Moment
          & 0.688&0.557
          % visionTS
          &\secondres{0.353} & 0.383
          %Chronos
          & 0.440&0.390
          %TimeMoE
          &\boldres{0.349}&0.379
          %FEDformer
          &0.376 & 0.419
          %TimesNet
          &0.384 & 0.402
          %Dlinear
          & 0.375&0.399
          %PatchTST
          &0.370&0.399
          \\

          & 192  
          % yelong 6m
          & 0.406 & 0.405 
          % yelong 50m
          & 0.404 & \secondres{0.403}
          % yelong 110m
          & 0.397 & \boldres{0.401}
          % yelong 300m
          & \secondres{0.396}& \boldres{0.401}
          % Moirai 
          & 0.412&0.413
          % TimesFM
          & 0.465&0.434
          % Moment
          & 0.688&0.560
          % visionTS
          &\boldres{0.392} & 0.410
          %Chronos
          & 0.492&0.426
          %TimeMoE
          & 0.384&0.404
          %FEDformer
          & 0.420&0.448
          %TimesNet
          & 0.436&0.429
          %Dlinear
          & 0.405&0.416
          %PatchTST
          &0.413&0.421
          \\

          & 336    
          % yelong 6m
          & 0.433 &0.421
          % yelong 50m
          & 0.429 & \secondres{0.416}
          % yelong 110m
          & 0.419 &\secondres{0.416}
          % yelong 300m
          & 0.418&\boldres{0.415}
          % Moirai 
          & 0.433&0.428
          % TimesFM
          & 0.503&0.456
          % Moment
          & 0.675&0.563
            % visionTS
          &\boldres{0.407} & 0.423
          %Chronos
          & 0.550&0.462
          %TimeMoE
          &\secondres{0.411}&0.430
          %FEDformer
          & 0.459&0.465
          %TimesNet
          & 0.491&0.469
          %Dlinear
          & 0.439&0.443
          %PatchTST
          &0.422&0.466
          \\
          & 720   
          % yelong 6m
          & 0.428 & 0.444
         % yelong 50m
          & 0.427 & 0.438
          % yelong 110m
          & 0.427 & 0.444
          % yelong 300m
          & \secondres{0.422}& \boldres{0.437}
          % Moirai 
          & 0.439&0.444
          % TimesFM
          & 0.511&0.481
          % Moment
          & 0.683&0.585
          % visionTS
          &\boldres{0.406} & \secondres{0.441}
          %Chronos
          & 0.615&0.543
          %TimeMoE
          & 0.427& 0.455
          %FEDformer
          &0.506 & 0.507
          %TimesNet
          & 0.521&0.500
          %Dlinear
          & 0.472&0.490
          %PatchTST
          &0.447&0.466
          \\
          % & 192   & \secondres{0.388} & \secondres{0.412} & 0.395 & 0.413 & 0.434 & 0.415 & 0.465 & 0.434 & 0.688 & 0.560 & 0.502 & 0.424& 0.395 & 0.413\\
          % & 336   & \boldres{0.411} & \secondres{0.430} & 0.447 & 0.453 & 0.495 & 0.445 & 0.503 & 0.456 & 0.675 & 0.563 & 0.576 & 0.467 & 0.447 & 0.453 \\
          % & 720   & \boldres{0.427} & 0.455 & 0.457 & 0.462 & 0.611 & 0.510 & 0.511 & 0.481 & 0.683 & 0.585 & 0.835 & 0.583 & 0.457 & 0.462\\
          \rowcolor{tabhighlight}
          & {\textbf{avg}}
          % yelong 6m
          & 0.408 & 0.412
         % yelong 50m
          &  0.405 & \secondres{0.408} 
          % yelong 110m
          & 0.399 & \secondres{0.408}
          % yelong 300m
          & 0.398&\boldres{0.407}
          % Moirai 
          & 0.415& 0.418
          % TimesFM
          & 0.473&0.444
          % Moment
          & 0.684&0.566
        % visionTS
          &\boldres{0.390} & 0.414
          %Chronos
          & 0.524&0.455
          %TimeMoE
          & \secondres{0.393}&0.417
          %FEDformer
          & 0.440&0.460
          %TimesNet
          & 0.458&0.450
          %Dlinear
          & 0.423&0.437
          %PatchTST
          &0.413&0.431
          \\
    \midrule
    % \multirow{4}[0]{*}{ETTh2} & 96    & 0.302 & 0.354 & \boldres{0.292} & 0.352 & 0.296 & \boldres{0.330} & 0.315 & 0.349 & 0.342 & 0.396 & 0.345 & 0.320 & \boldres{0.292} & 0.352 \\
    %       & 192   & 0.364 & 0.385 & \boldres{0.347} & 0.379 & 0.361 & \boldres{0.371} & 0.388 & 0.395 & 0.354 & 0.402 & 0.406 & 0.399 & \boldres{0.347} & 0.379 \\
    %       & 336   & 0.417 & 0.425 & 0.406 & 0.419 & 0.390 & \boldres{0.390} & 0.422 & 0.427 & \boldres{0.356} & \secondres{0.407} & 0.492 & 0.453 & 0.406 & 0.419 \\
    %       & 720   & 0.537 & 0.496 & 0.439 & 0.447 & 0.423 & \boldres{0.418} & 0.443 & 0.454 & \boldres{0.395} & 0.434 & 0.603 & 0.511 & 0.439 & 0.447 \\
    %       \rowcolor{tabhighlight}
    %       & {\textbf{AVG}} & 0.405 & 0.415 & 0.371 & 0.399 & 0.367 & \boldres{0.377} & 0.392 & 0.406 & \boldres{0.361} & 0.409 & 0.455 & 0.427 & 0.371 & 0.399 \\
     \multirow{4}[1]{*}{ETTh2} & 96    
          % yelong 6m
          & 0.273 & 0.326 
          % yelong 50m
          & 0.275 & \secondres{0.319}
          % yelong 110m
          & \boldres{0.267} & \boldres{0.316}
          % yelong 300m
          & \secondres{0.271}&\secondres{0.319}
          % Moirai 
          & 0.294&0.330
          % TimesFM
          & 0.315&0.349
          % Moment
          & 0.342&0.396
          % visionTS
          & \secondres{0.271} & 0.328
          %Chronos
          & 0.307&0.343
          %TimeMoE
          & 0.292&0.352
          %FEDformer
          & 0.358&0.397
          %TimesNet
          & 0.340&0.374
          %Dlinear
          & 0.289&0.353
          %PatchTST
          &0.274&0.336
          \\

          & 192  
          % yelong 6m
          & 0.339 & 0.372
          % yelong 50m
          & 0.340 & \secondres{0.364} 
          % yelong 110m
          & \secondres{0.330} & \boldres{0.360} 
          % yelong 300m
          & 0.336& \secondres{0.364}
          % Moirai 
          & 0.361&0.371
          % TimesFM
          & 0.388&0.395
          % Moment
          & 0.354&0.402
        % visionTS
          & \boldres{0.328} & 0.367
          %Chronos
          & 0.376&0.392
          %TimeMoE
          & 0.347&0.379
          %FEDformer
          &0.429 &0.439
          %TimesNet
          & 0.402&0.414
          %Dlinear
          & 0.383&0.418
          %PatchTST
          &0.339&0.379
          \\

          & 336    
          % yelong 6m
          & 0.370 & 0.399
          % yelong 50m
          &  0.366& 0.388 
          % yelong 110m
          & 0.354 &0.382
          % yelong 300m
          & 0.364& 0.388
          % Moirai 
          & 0.370&0.390
          % TimesFM
          & 0.422&0.427
          % Moment
          & 0.356&0.407
        % visionTS
          &\secondres{0.345} & \secondres{0.381}
          %Chronos
          & 0.408&0.430
          %TimeMoE
          & 0.391&0.418
          %FEDformer
          & 0.497&0.487
          %TimesNet
          & 0.452&0.452
          %Dlinear
          & 0.448&0.465
          %PatchTST
          &\boldres{0.329}&\boldres{0.380}
          \\
          & 720   
          % yelong 6m
          & 0.394 & 0.429
         % yelong 50m
          & 0.376 & \secondres{0.409} 
          % yelong 110m
          & \boldres{0.369} & \boldres{0.407}
          % yelong 300m
          & 0.378&0.411
          % Moirai 
          & 0.411&0.418
          % TimesFM
          & 0.443&0.454
          % Moment
          & 0.395&0.434
        % visionTS
          &0.388 & 0.422
          %Chronos
          & 0.501&0.477
          %TimeMoE
          & 0.419&0.447
          %FEDformer
          &0.463 &0.474
          %TimesNet
          & 0.462&0.468
          %Dlinear
          & 0.605&0.551
          %PatchTST
          &\secondres{0.379}&0.422
          \\
          % & 192   & \secondres{0.388} & \secondres{0.412} & 0.395 & 0.413 & 0.434 & 0.415 & 0.465 & 0.434 & 0.688 & 0.560 & 0.502 & 0.424& 0.395 & 0.413\\
          % & 336   & \boldres{0.411} & \secondres{0.430} & 0.447 & 0.453 & 0.495 & 0.445 & 0.503 & 0.456 & 0.675 & 0.563 & 0.576 & 0.467 & 0.447 & 0.453 \\
          % & 720   & \boldres{0.427} & 0.455 & 0.457 & 0.462 & 0.611 & 0.510 & 0.511 & 0.481 & 0.683 & 0.585 & 0.835 & 0.583 & 0.457 & 0.462\\
          \rowcolor{tabhighlight}
          & {\textbf{avg}}
          % yelong 6m
          & 0.344 & 0.382
         % yelong 50m
          & 0.339 & \secondres{0.370} 
          % yelong 110m
          & \boldres{0.330} & \boldres{0.366}
          % yelong 300m
          & 0.337& \secondres{0.370}
          % Moirai 
          & 0.359&0.377
          % TimesFM
          & 0.392&0.406          
          % Moment
          & 0.362&0.410
        % visionTS
          &\secondres{0.333} & 0.375
          %Chronos
          & 0.398&0.411
          %TimeMoE
          & 0.362&0.399
          %FEDformer
          &0.437 &0.449
          %TimesNet
          & 0.414&0.427
          %Dlinear
          & 0.456&0.445
          %PatchTST
          &\boldres{0.330}&0.379
          \\
    \midrule
    % \multirow{4}[0]{*}{ETTm1} & 96    & \secondres{0.309} & 0.357 & \boldres{0.281} & \boldres{0.341} & 0.380 & 0.361 & 0.361 & 0.370 & 0.654 & 0.527 & 0.457 & 0.403 & \boldres{0.281} & \boldres{0.341} \\
    %       & 192   & \secondres{0.346} & 0.381 & \boldres{0.305} & \boldres{0.358} & 0.412 & 0.383 & 0.414 & 0.405 & 0.662 & 0.532 & 0.530 & 0.450 & \boldres{0.305} & \boldres{0.358} \\
    %       & 336   & \secondres{0.373} & 0.408 & \boldres{0.369} & \secondres{0.395} & 0.436 & 0.400 & 0.445 & 0.429 & 0.672 & 0.537 & 0.577 & 0.481 & \boldres{0.369} & \secondres{0.395}\\
    %       & 720   & 0.475 & 0.477 & 0.469 & 0.472 & \secondres{0.462} & \boldres{0.420} & 0.512 & 0.471 & 0.692 & 0.551 & 0.660 & 0.526 & 0.469 & 0.472 \\
    %       \rowcolor{tabhighlight}
    %       & {\textbf{AVG}} & \secondres{0.376} & 0.405 & \boldres{0.356} & \secondres{0.391} & 0.422 & 0.391 & 0.433 & 0.418 & 0.670 & 0.536 & 0.555 & 0.465& \boldres{0.356} & \secondres{0.391}  \\
     \multirow{4}[1]{*}{ETTm1} & 96    
          % yelong 6m
          & 0.297 & 0.324 
          % yelong 50m
          & 0.294 & 0.320
          % yelong 110m
          & \boldres{0.281} & \boldres{0.313} 
          % yelong 300m
          & 0.284&\secondres{0.315}
          % Moirai 
          & 0.363&0.356
          % TimesFM
          & 0.361&0.370
          % Moment
          & 0.654&0.527
          % visionTS
          &0.341 & 0.347
          %Chronos
          & 0.454&0.403
          %TimeMoE
          & \boldres{0.281}&0.341
          %FEDformer
          & 0.464&0.416
          %TimesNet
          & 0.338&0.375
          %Dlinear
          & 0.299&0.343
          %PatchTST
          & \secondres{0.290}&0.342
          \\

          & 192  
          % yelong 6m
          & 0.346 &0.355  
          % yelong 50m
          & 0.344 & 0.352 
          % yelong 110m
          & \boldres{0.328} &\boldres{0.343}  
          % yelong 300m
          & \secondres{0.332}&\secondres{0.346}
          % Moirai 
          & 0.388&0.375
          % TimesFM
          & 0.414&0.405
          % Moment
          & 0.662&0.532
        % visionTS
          &0.360 & 0.360
          %Chronos
          & 0.530&0.450
          %TimeMoE
          & 0.305&0.358
          %FEDformer
          &0.426 &0.441
          %TimesNet
          & 0.371&0.387
          %Dlinear
          & 0.355&0.365
          %PatchTST
          & \secondres{0.332}&0.369
          \\

          & 336    
          % yelong 6m
          & 0.384 & 0.378
          % yelong 50m
          & 0.383 & 0.376
          % yelong 110m
          & \boldres{0.362} & \boldres{0.366}
          % yelong 300m
          & 0.365&\secondres{0.369}
          % Moirai 
          & 0.416&0.392
          % TimesFM
          & 0.445&0.429
          % Moment
          & 0.672&0.537
        % visionTS
          &0.377 & 0.374
          %Chronos
          & 0.577&0.481
          %TimeMoE
          & 0.369&0.395
          %FEDformer
          & 0.445&0.459
          %TimesNet
          & 0.410&0.411
          %Dlinear
          & 0.369&0.386
          %PatchTST
          &\secondres{0.366}&0.392
          \\
          & 720   
          % yelong 6m
          & 0.441 & 0.414
         % yelong 50m
          & 0.449 & 0.416 
          % yelong 110m
          & \secondres{0.418} & \boldres{0.401}
          % yelong 300m
          & 0.423&\secondres{0.404}
          % Moirai 
          & 0.460&0.418
          % TimesFM
          & 0.512&0.471
          % Moment
          & 0.692&0.551
        % visionTS
          &0.416 & 0.405
          %Chronos
          & 0.660&0.526
          %TimeMoE
          &0.469 &0.472
          %FEDformer
          & 0.543&0.490
          %TimesNet
          & 0.478&0.450
          %Dlinear
          & 0.425&0.421
          %PatchTST
          &\boldres{0.416}&0.420
          \\
          % & 192   & \secondres{0.388} & \secondres{0.412} & 0.395 & 0.413 & 0.434 & 0.415 & 0.465 & 0.434 & 0.688 & 0.560 & 0.502 & 0.424& 0.395 & 0.413\\
          % & 336   & \boldres{0.411} & \secondres{0.430} & 0.447 & 0.453 & 0.495 & 0.445 & 0.503 & 0.456 & 0.675 & 0.563 & 0.576 & 0.467 & 0.447 & 0.453 \\
          % & 720   & \boldres{0.427} & 0.455 & 0.457 & 0.462 & 0.611 & 0.510 & 0.511 & 0.481 & 0.683 & 0.585 & 0.835 & 0.583 & 0.457 & 0.462\\
          \rowcolor{tabhighlight}
          & {\textbf{avg}}
          % yelong 6m
          & 0.370 & 0.368
         % yelong 50m
          & 0.367 & 0.366 
          % yelong 110m
          & \boldres{0.347} & \boldres{0.356}
          % yelong 300m
          & \secondres{0.351}&\secondres{0.358}
          % Moirai 
          & 0.407&0.385
          % TimesFM
          & 0.433&0.418
          % Moment
          & 0.670&0.536
        % visionTS
          &0.374 & 0.372
          %Chronos
          & 0.555&0.465
          %TimeMoE
          &0.356 &0.392
          %FEDformer
          & 0.448&0.452
          %TimesNet
          & 0.400&0.406
          %Dlinear
          & 0.362&0.379
          %PatchTST
          &\secondres{0.351}&0.381
          \\
    \midrule
    % \multirow{4}[0]{*}{ETTm2} & 96    & \boldres{0.197} & 0.286 & \secondres{0.198} & 0.288 & 0.211 & 0.274 & 0.202 & \boldres{0.270} & 0.260 & 0.335 & \boldres{0.197} & 0.271 & \secondres{0.198} & 0.288 \\
    %       & 192   & \secondres{0.250} & 0.322 & \boldres{0.235} & \boldres{0.312} & 0.281 & 0.318 & 0.289 & 0.321 & 0.289 & 0.350 & \secondres{0.254} & \secondres{0.314} & \boldres{0.235} & \boldres{0.312} \\
    %       & 336   & 0.337 & 0.375 & \boldres{0.293} & \boldres{0.348} & 0.341 & 0.355 & 0.360 & 0.366 & 0.324 & 0.369 & \secondres{0.313} & 0.353 & \boldres{0.293} & \boldres{0.348} \\
    %       & 720  & 0.480 & 0.461 & 0.427 & 0.428 & \secondres{0.485} & 0.428 & 0.462 & 0.430 & \boldres{0.394} & 0.409 & 0.416 & 0.415 & 0.427 & 0.428 \\
    %       \rowcolor{tabhighlight}
    %       & {\textbf{AVG}} & 0.316 & 0.361 & \boldres{0.288} & 0.344 & 0.329 & 0.343 & 0.328 & 0.346 & 0.316 & 0.365 & \secondres{0.295} & \secondres{0.338} & \boldres{0.288} & 0.344 \\
     \multirow{4}[1]{*}{ETTm2} & 96    
          % yelong 6m
          & 0.163 & 0.240 
          % yelong 50m
          &  0.162& \secondres{0.235} 
          % yelong 110m
          & \secondres{0.161} & \boldres{0.234} 
          % yelong 300m
          & \boldres{0.160}&\boldres{0.234}
          % Moirai 
          & 0.205&0.273
          % TimesFM
          & 0.202&0.270
          % Moment
          & 0.260&0.335
            % visionTS
          &0.228 & 0.282
          %Chronos
          & 0.197&0.271
          %TimeMoE
          & 0.197&0.286
          %FEDformer
          &0.203 &0.287
          %TimesNet
          & 0.187&0.267
          %Dlinear
          & 0.167&0.260
          %PatchTST
          &0.165&0.255
          \\

          & 192  
          % yelong 6m
          & \boldres{0.218} & 0.281 
          % yelong 50m
          & \secondres{0.219} & \secondres{0.277} 
          % yelong 110m
          & \boldres{0.218} & \boldres{0.276} 
          % yelong 300m
          & \boldres{0.218}&\boldres{0.276}
          % Moirai 
          & 0.275&0.316
          % TimesFM
          & 0.289&0.321
          % Moment
          & 0.289&0.350
        % visionTS
          &0.262 & 0305
          %Chronos
          & 0.254&0.314
          %TimeMoE
          & 0.235&0.312
          %FEDformer
          & 0.269&0.328
          %TimesNet
          & 0.249&0.309
          %Dlinear
          & 0.224&0.303
          %PatchTST
          &0.220&0.292
          \\

          & 336    
          % yelong 6m
          & 0.269 &0.317
          % yelong 50m
          & 0.270 &  0.313
          % yelong 110m
          & \boldres{0.266} &\secondres{0.312}
          % yelong 300m
          & \secondres{0.268}&\boldres{0.311}
          % Moirai 
          & 0.329&0.350
          % TimesFM
          & 0.360&0.366
          % Moment
          & 0.324&0.369
        % visionTS
          &0.293 & 0.328
          %Chronos
          & 0.313&0.353
          %TimeMoE
          & 0.293&0.348
          %FEDformer
          & 0.325&0.366
          %TimesNet
          & 0.321&0.351
          %Dlinear
          & 0.281&0.342
          %PatchTST
          &0.274&0.329
          \\
          & 720   
          % yelong 6m
          & 0.350 & 0.371
         % yelong 50m
          & 0.351 & 0.367 
          % yelong 110m
          & \boldres{0.340} & \boldres{0.363}
          % yelong 300m
          & \secondres{0.348}&\secondres{0.366}
          % Moirai 
          & 0.402&0.408
          % TimesFM
          & 0.462&0.430
          % Moment
          & 0.394&0.409
        % visionTS
          &0.343 & 0.370
          %Chronos
          & 0.416&0.415
          %TimeMoE
          & 0.427&0.428
          %FEDformer
          &0.421 &0.415
          %TimesNet
          & 0.497&0.403
          %Dlinear
          & 0.397&0.421
          %PatchTST
          &0.362&0.385
          \\
          % & 192   & \secondres{0.388} & \secondres{0.412} & 0.395 & 0.413 & 0.434 & 0.415 & 0.465 & 0.434 & 0.688 & 0.560 & 0.502 & 0.424& 0.395 & 0.413\\
          % & 336   & \boldres{0.411} & \secondres{0.430} & 0.447 & 0.453 & 0.495 & 0.445 & 0.503 & 0.456 & 0.675 & 0.563 & 0.576 & 0.467 & 0.447 & 0.453 \\
          % & 720   & \boldres{0.427} & 0.455 & 0.457 & 0.462 & 0.611 & 0.510 & 0.511 & 0.481 & 0.683 & 0.585 & 0.835 & 0.583 & 0.457 & 0.462\\
          \rowcolor{tabhighlight}
          & {\textbf{avg}}
          % yelong 6m
          & 0.250 & 0.302
         % yelong 50m
          & 0.250 & \secondres{0.298} 
          % yelong 110m
          & \boldres{0.246} & \boldres{0.296}
          % yelong 300m
          & \secondres{0.249}&\boldres{0.296}
          % Moirai 
          & 0.303&0.337
          % TimesFM
          & 0.328&0.346
          % Moment
          & 0.316&0.365
        % visionTS
          &0.282 & 0.321
          %Chronos
          & 0.295&0.338
          %TimeMoE
          & 0.288&0.344
          %FEDformer
          &0.305 &0.349
          %TimesNet
          & 0.391&0.333
          %Dlinear
          & 0.356&0.331
          %PatchTST
          &0.255&0.315
          \\
    \midrule
    % \multirow{4}[0]{*}{Weather} & 96    & \secondres{0.159} & \secondres{0.213} & \boldres{0.157} & \boldres{0.211} & 0.199 & \boldres{0.211} & - & - & 0.243 & 0.255 & 0.194 & 0.235 &\boldres{0.157} & \boldres{0.211}\\
    %       & 192   & 0.215 & 0.266 & \boldres{0.208} & \secondres{0.256} & 0.246 & \boldres{0.251} & - & - & 0.278 & 0.329 & 0.249 & 0.285 & \boldres{0.208} & \secondres{0.256} \\
    %       & 336   & 0.291 & 0.322 & \boldres{0.255} & \boldres{0.290} & 0.286 & \secondres{0.291} & - & - & 0.306 & 0.346 & 0.302 & 0.327 & \boldres{0.255} & \boldres{0.290} \\
    %       & 720   & 0.415 & 0.400 & 0.405 & 0.397 & 0.373 & \secondres{0.354} & - & - & \secondres{0.350} & 0.374 & 0.372 & 0.378 & 0.405 & 0.397 \\
    %       \rowcolor{tabhighlight}
    %       & {\textbf{AVG}} & 0.270 & 0.300 & \boldres{0.256} & 0.288 & \secondres{0.264} & \boldres{0.273} & - & - & 0.294 & 0.326 & 0.279 & 0.306& \boldres{0.256} & 0.288  \\
     \multirow{4}[1]{*}{Weather} & 96    
          % yelong 6m
          & 0.169 & 0.240 
          % yelong 50m
          & 0.159 & 0.195 
          % yelong 110m
          & 0.153 & \secondres{0.190} 
          % yelong 300m
          & \boldres{0.144}&\boldres{0.180}
          % Moirai 
          & 0.198&0.211
          % TimesFM
          & -&-
          % Moment
          & 0.243&0.255
            % visionTS
          &0.220 & 0.257
          %Chronos
          & 0.194&0.235
          %TimeMoE
          & 0.157&0.211
          %FEDformer
          & 0.217&0.396
          %TimesNet
          & 0.172&0.220
          %Dlinear
          & 0.152&0.237
          %PatchTST
          &\secondres{0.149}&0.198
          \\

          & 192  
          % yelong 6m
          & 0.212 & 0.249 
          % yelong 50m
          & 0.202 & 0.237 
          % yelong 110m
          & 0.196 & \secondres{0.234} 
          % yelong 300m
          & \boldres{0.186}&\boldres{0.223}
          % Moirai 
          & 0.246&0.251
          % TimesFM
          & - &- 
          % Moment
          & 0.278&0.329
        % visionTS
          &0.244 & 0.275
          %Chronos
          & 0.249&0.285
          %TimeMoE
          & 0.208&0.256
          %FEDformer
          & 0.276&0.336
          %TimesNet
          & 0.219&0.261
          %Dlinear
          & 0.220&0.282
          %PatchTST
          &\secondres{0.194}&0.241
          \\

          & 336    
          % yelong 6m
          & 0.258 &0.285
          % yelong 50m
          & 0.250 & 0.275 
          % yelong 110m
          & 0.246 &\secondres{0.273}
          % yelong 300m
          & \boldres{0.235}&\boldres{0.263}
          % Moirai 
          &0.274 &0.291
          % TimesFM
          & -&-
          % Moment
          & 0.306&0.375
        % visionTS
          &0.280 & 0.299
          %Chronos
          & 0.302&0.327
          %TimeMoE
          & 0.255&0.290
          %FEDformer
          & 0.339&0.380
          %TimesNet
          & 0.280&0.306
          %Dlinear
          & 0.265&0.319
          %PatchTST
          &\secondres{0.245}&0.282
          \\
          & 720   
          % yelong 6m
          & 0.316 & 0.328
         % yelong 50m
          & \secondres{0.313} & 0.320 
          % yelong 110m
          & 0.314 & \secondres{0.325}
          % yelong 300m
          & \boldres{0.302}&\boldres{0.313}
          % Moirai 
          & 0.337&0.340
          % TimesFM
          & -&-
          % Moment
          & 0.350&0.374
        % visionTS
          &0.330 & 0.337
          %Chronos
          & 0.372&0.378
          %TimeMoE
          & 0.405&0.397
          %FEDformer
          & 0.403&0.428
          %TimesNet
          & 0.365&0.359
          %Dlinear
          & 0.323&0.362
          %PatchTST
          &0.314&0.334
          \\
          % & 192   & \secondres{0.388} & \secondres{0.412} & 0.395 & 0.413 & 0.434 & 0.415 & 0.465 & 0.434 & 0.688 & 0.560 & 0.502 & 0.424& 0.395 & 0.413\\
          % & 336   & \boldres{0.411} & \secondres{0.430} & 0.447 & 0.453 & 0.495 & 0.445 & 0.503 & 0.456 & 0.675 & 0.563 & 0.576 & 0.467 & 0.447 & 0.453 \\
          % & 720   & \boldres{0.427} & 0.455 & 0.457 & 0.462 & 0.611 & 0.510 & 0.511 & 0.481 & 0.683 & 0.585 & 0.835 & 0.583 & 0.457 & 0.462\\
          \rowcolor{tabhighlight}
          & {\textbf{avg}}
          % yelong 6m
          & 0.239 & 0.268
         % yelong 50m
          & 0.231 & 0.257 
          % yelong 110m
          & 0.227 & \secondres{0.255}
          % yelong 300m
          & \boldres{0.217}&\boldres{0.245}
          % Moirai 
          &0.264 &0.273
          % TimesFM
          & - & -
          % Moment
          & 0.294&0.326
        % visionTS
          &0.269 & 0.292
          %Chronos
          & 0.279&0.306
          %TimeMoE
          & 0.256&0.289
          %FEDformer
          &0.309 & 0.360
          %TimesNet
          & 0.259&0.287
          %Dlinear
          & 0.240&0.300
          %PatchTST
          &\secondres{0.226}&0.264
          \\
    \midrule
    % \rowc\rowcolor{blue!15}
    % \multicolumn{2}{c|}{\scalebox{1.1}
    % {\textbf{Average}}} & \secondres{0.336} & 0.380 & \boldres{0.322} & \secondres{0.372} & 0.359 & 0.373 & 0.396 & 0.413 & 0.461 & 0.454 & 0.416 & 0.405 & \secondres{0.336} & 0.380 \\
    % \midrule
    \rowc
    \multicolumn{2}{c|}{\bf Average Rank}
    & 5.45 & 5.00
    & 4.75 & 2.90
    & \boldres{2.30} & \boldres{1.60}
    & \secondres{2.40} & \secondres{1.65}
    & 9.25 & 6.90 
    & 11.50 & 10.50 
    & 11.80 & 12.35
    & 5.45 &6.25
    & 11.20 & 10.65
    & 5.85 & 7,9
    &11.70 & 12.55
    & 10.15 &9.80
    &7.45 & 9.15
    & 4.20 & 6.25
    \\
    \bottomrule
    \end{tabular}%
    % }
  }
  \label{tab:zero_shot_full}%
  \vskip -0.20in
\end{table*}

% You can have as much text here as you want. The main body must be at most $8$ pages long.
% For the final version, one more page can be added.
% If you want, you can use an appendix like this one.  

% The $\mathtt{\backslash onecolumn}$ command above can be kept in place if you prefer a one-column appendix, or can be removed if you prefer a two-column appendix.  Apart from this possible change, the style (font size, spacing, margins, page numbering, etc.) should be kept the same as the main body.

\begin{table*}[htp]
\centering
\caption{GIFT-Eval datasets}
\resizebox{0.7\columnwidth}{!}{
\begin{tabular}{cccccccc}
\toprule
Dataset&Source &domain &Frequency &\# Series & \# Tasks\\
\midrule
Weather&\citealt{wu2021autoformer}&Nature &10T,H,D&1&7\\
BizITObs&\citealt{palaskar2023automixer}&Web/CloudOps&10S,5T,H&24&12\\
Bitbrains&\citealt{shen2015statistical}&Web/CloudOps&5T,H&1750&8\\
Restaurant&\citealt{recruit-restaurant-visitor-forecasting}&Sales&D&807&1\\
ETT&\citealt{zhou2021informer}&Energy&15T,H,D,W-THU&2&18\\
libcity&\citealt{libcity}&Transport&5T,H,D,15T,&518&14\\
Solar&\citealt{lai2018modeling}&Energy&10T,H,D,W-FRI&137&8\\
Hierarchical Sales&\citealt{mancuso2021machine}&Salses&D,W-WED&118&2\\
M4&\citealt{godahewa2021monash}&Econ/Fin& A-DEC,Q-DEC,M,W-SUN,D,H&100741&6\\
Hospital & \citealt{godahewa2021monash}&Healthcare &M &767&1\\
COVID Death &\citealt{godahewa2021monash}&Healthcare &D &226&1\\
US Births& \citealt{godahewa2021monash}&Healthcare &D,W-TUE,M&1&3\\
Saugeen&\citealt{godahewa2021monash}&Nature&D,W-THU,M&1&3\\
Temperature Rain&\citealt{godahewa2021monash}&Nature&D&32072&1\\
KDD CUP 2018&\citealt{godahewa2021monash}&Nature&H,D&270&4\\
Car Parts&\citealt{godahewa2021monash}&Sales&M&2674&1\\
Electricity&\citealt{godahewa2021monash}&Energy&15T,H,D,W-FRI&370&8\\
\bottomrule
\end{tabular}\label{tab:gift-eval-dataset}
}
\end{table*}
\label{app:GIFT_Eval_details}

\begin{table*}[ht]
\centering
\caption{Results on GIFT-Eval aggregated by domain. The table shows MASE,CRPS and
Rank for each method.}
\vskip -0.1in
\resizebox{1.1\columnwidth}{!}{
\begin{tabular}{l|cccccccccccccccccccccccccccc}
% \rowcolor[HTML]{FFFFFF} 
\toprule
% \hline
Model & \multicolumn{3}{c}{Econ/Fin}  &\multicolumn{3}{c}{Energy} &\multicolumn{3}{c}{Healthcare} &\multicolumn{3}{c}{Nature}&\multicolumn{3}{c}{Sales}&\multicolumn{3}{c}{Transport}&\multicolumn{3}{c}{Web/CloudOps}\\
&MASE&CRPS&Rank&MASE&CRPS&Rank&MASE&CRPS&Rank&MASE&CRPS&Rank&MASE&CRPS&Rank&MASE&CRPS&Rank&MASE&CRPS&Rank\\
\midrule
\textsc{YINGLONG}$_{300m}$
&	0.899	&0.786	&14
&	0.870	&0.539	&5.19
&	0.704	&0.574	&13.20
&	0.746	&0.287	&5.27
&	0.742	&0.368	&13.00
&	0.666	&0.435	&6.40
&	0.501	&0.461	&8.65\\
\textsc{YINGLONG}$_{100m}$
&0.939	&0.790	&15.5
&	0.906	&0.560	&6.25
&	0.710	&0.579	&14
&	0.742	&0.288	&4.8
&	0.737	&0.365	&12.75
&	0.692	&0.452	&8.33
&	0.483	&0.455	&8.45\\
\textsc{YINGLONG}$_{50m}$
&	0.918	&0.793	&15.33
&   0.909	&0.570	&7.16
&	0.706	&0.546	&12.8
&	0.748	&0.296	&5.93
&	0.788	&0.390	&17.25
&	0.698	&0.458	&9.47
&	0.512	&0.465	&8.15\\
\textsc{YINGLONG}$_{6m}$
&	1.080	&0.899	&20
&	0.978	&0.617	&11.06
&	0.727	&0.589	&16.6
&	0.779	&0.318	&11.2
&	0.767	&0.385	&15.5
&	0.746	&0.488	&12.93
&	0.554	&0.499	&11.85\\
TabPFN-TS
& 0.802 & 0.717 & 7.33
& 0.900 & 0.564 & 8.53
& 0.554 &0.411 & 3
& 0.780 & 0.297 & 6.2 
& 0.713 & 0.351 & 9.25 
& 0.682 & 0.447 & 9 
& 0.618 & 0.552 & 13.2\\ 
chronos-bolt$_b$
&0.799 	&0.717 	&6.67 
&0.845 	&0.550 	&7 
&0.691 	&0.532 	&12 
&0.667 	&0.265 	&3.07 
&0.693 	&0.344 	&4 
&0.686 	&0.474 	&9.93 
&0.624 	&0.591 	&15.00 \\
chronos-bolt$_s$
&0.816 	&0.700 	&6.00 
&0.864 	&0.564 	&8.19 
&0.671 	&0.511 	&11.80 
&0.704 	&0.284 	&4.67 
&0.696 	&0.347 	&6.25 
&0.692 	&0.474 	&11.67 
&0.627 	&0.559 	&14.05 \\
TimesFM-V2$_{500m}$
&0.641 	&0.546 	&2.67 
&0.890 	&0.574 	&8.81 
&{0.597} 	&0.455 	&{5.80} 
&{0.624} 	&{0.284} 	&10.67 
&0.699 	&{0.342} 	&{4.75} 
&0.645 	&0.432 	&{5.73} 
&0.515 	&0.518 	&13.55 \\
Moirai$_\mathrm{large}$
& 0.845 & 0.732 & 8.33
& 1.025 & 0.63 & 11.53
& 0.700 & 0.534 & 11.40 
& 0.750 & 0.31 & 8.87 
& 0.711 & 0.364 & 10.75 
& {0.601} & {0.39} & {8} 
& 0.666 & 0.591 & 13.35\\
Moirai$_{\mathrm{base}}$
& 0.906 & 0.786 & 9.83 
& 0.998 & 0.623 & 11.38 
& 0.683 & 0.518 & 9.8 
& 0.771 & 0.314 & 8.87 
& 0.694 & 0.346 & 5
& {0.637} & {0.413} & 9.07 
& 0.745 & 0.617 & 14.5 \\
Moirai$_{\mathrm{small}}$
& 0.986 & 0.785 & 13.17 
& 1.069 & 0.656 & 13.78
& 0.849 & 0.703 & 20.8 
& 0.807 & 0.335 & 12.47
& 0.731 & 0.361 & 11.75
& 0.732 & 0.476 & 14.47 
& 0.673 & 0.614 & 15 \\
Chronos$_{\mathrm{large}}$
& {0.783} & 0.758 & 11.17 
& 0.919 & 0.628 & 13.47
& 0.599 & {0.446} & 6.20 
& 0.813 & 0.364 & 16.53 
& 0.724 & 0.362 & 13.25 
& 0.714 & 0.512 & 15.33 
& 0.675 & 0.647 & 18.5 \\
Chronos$_{\mathrm{base}}$
& {0.783} & 0.751 & 10.17 
& 0.924 & 0.631 & 13.63
& 0.645 & 0.485 & 8.8
& 0.823 & 0.366 & 17.27 
& 0.726 & 0.363 & 13.5 
& 0.712 & 0.512 & 15.13 
& 0.676 & 0.651 & 17.95 \\
Chronos$_{\mathrm{small}}$
& 0.797 & 0.763 & 12
& 0.947 & 0.648 & 15.34
& 0.607 & 0.496 & 9
& 0.851 & 0.383 & 18.4 
& 0.733 & 0.366 & 15 
& 0.737 & 0.530 & 17.93 
& 0.678 & 0.629 & 16.2 \\
TimesFM
& 0.824 & 0.716 & 8.67
& 1.016 & 0.673 & 15.06
& 0.698 & 0.652 & 12.4 
& 0.880 & 0.333 & 13.07
& 0.700 & {0.344} & 5 
& 0.741 & 0.510 & 14.2 
& 1.419 & 0.739 & 21.95 \\
Lag-Llama 
& 2.910 & 1.839 & 30.33
& 1.392 & 0.923 & 23 
& 1.621 & 1.599 & 29.4 
& 0.932 & 0.385 & 18.2
& 0.841 & 0.442 & 20.5 
& 0.842 & 0.572 & 19.73 
& 0.753 & 0.734 & 22.55\\
TTMs
& 1.297 & 1.142 & 26.17 
& 1.059 & 0.847 & 23.03 
& 1.150 & 1.226 & 28.20 
& 0.845 & 0.404 & 20.53 
& 0.878 & 0.540 & 26.50 
& 0.891 & 0.730 & 25.47 
& 0.887 & 0.851 & 25.05 \\
Timer
& 1.809 & 1.475 & 28.83 
& 1.287 & 1.019 & 26.38 
& 1.390 & 1.700 & 29.6 
& 0.881 & 0.444 & 23.6 
& 0.775 & 0.468 & 23 
& 0.894 & 0.743 & 26.07 
& 0.711 & 0.771 & 24.40 \\
VisionTS
& 0.931 & 1.046 & 24.83 
& 0.993 & 0.782 & 21.72 
& 0.749 & 0.681 & 21 
& 0.860 & 0.406 & 20.8 
& 0.817 & 0.492 & 25.50 
& 0.739 & 0.601 & 22.20 
& {0.472} & 0.603 & 17.15\\ 
\midrule
PatchTST
& 0.908 & 0.803 & 11.5
& 0.983 & 0.612 & 11.03 
& 0.686 & 0.576 & 15 
& 0.916 & 0.347 & 14.87
& {0.690} & 0.348 & 6.75 
& 0.709 & 0.461 & 11.53 
& {0.462} & {0.437} & {5.45} \\
iTransformer
& 0.989 & 0.848 & 15.17 
& 1.110 & 0.695 & 13.44 
& 0.774 & 0.628 & 15.80 
& 0.851 & 0.342 & 13.73
& 0.699 & 0.351 & 9.25
& 0.707 & 0.460 & 11.53 
& 0.488 & {0.454} & {6.9} \\
TFT
& 1.034 & 0.841 & 14.67 
& 1.009 & 0.630 & 13.28
& 0.660 & 0.512 & 13.2 
& 0.871 & 0.348 & 14.87 
& 0.716 & 0.352 & 13.25 
& 0.679 & 0.443 & 9.47
& 0.662 & 0.503 & 8.5\\ 
TIDE
& 1.507 & 1.084 & 25.5
& 1.167 & 0.751 & 17.91 
& 0.803 & 0.912 & 23.00 
& 1.372 & 0.561 & 23.07 
& 0.981 & 0.484 & 24.25 
& 0.790 & 0.531 & 17.80 
& 0.623 & 0.568 & 16.65 \\
DeepAR
& 1.541 & 1.221 & 22 
& 1.782 & 1.072 & 23.56
& 0.765 & 0.723 & 17.8 
& 1.644 & 0.535 & 21.47 
& 0.707 & 0.352 & 11 
& 0.745 & 0.484 & 12.4 
& 0.850 & 0.633 & 17.9 \\
Crossformer 
& 29.323 & 109.221 & 32.00 
& 2.193 & 1.224 & 22.72
& 8.529 & 5.177 & 24.40 
& 2.979 & 0.690 & 21 
& 1.593 & 5.768 & 31.75 
& 1.769 & 0.811 & 15.93 
& 0.92 & 0.615 & 18.05\\
N-BEATS
& 0.861 & 0.967 & 20.67 
& 1.184 & 0.935 & 24.59 
& 0.691 & 0.713 & 12.4 
& 0.933 & 0.532 & 25.13 
& 0.704 & 0.414 & 18.25 
& 0.731 & 0.593 & 21.67 
& 0.543 & 0.570 & 16.4 \\
DLinear
& 1.133 & 1.124 & 26.00 
& 1.151 & 0.879 & 23.97
& 0.792 & 0.806 & 24.20 
& 1.117 & 0.496 & 25.20 
& 0.808 & 0.481 & 24.25 
& 0.808 & 0.654 & 24.13
& 0.724 & 0.660 & 20.30 \\
\midrule
Auto Arima 
& 0.866 & 0.821 & 13.50 
& 1.011 & 0.833 & 21 
& 0.784 & 0.570 & 13.00 
& 1.018 & 0.658 & 24.47 
& 0.813 & 0.458 & 23.75 
& 0.974 & 0.763 & 25.67 
& 0.957 & 0.904 & 23.15 \\
Auto Theta 
& 0.983 & 0.841 & 14.17 
& 1.358 & 1.702 & 26.72
& 0.951 & 0.803 & 20.4 
& 1.060 & 0.910 & 27.07 
& 0.873 & 0.48 & 24.50 
& 1.082 & 1.326 & 29.13
& 0.521 & 0.608 & 18.9\\ 

Auto ETS
& 0.899 & 0.940 & 16 
& 1.479 & 10.102 & 25.41 
& 0.797 & 0.586 & 11.8 
& 1.121 & 7.021 & 27.13
& 0.887 & 2.195 & 28.25 
& 1.205 & 42.407 & 28.73 
& 0.718 & 2.637 & 26.3 \\
Seasonal Naive
& 1.000 & 1.000 & 22.67 
& 1.000 & 1.000 & 25.44 
& 1.000 & 1.000 & 24.60 
& 1.000 & 1.000 & 29.67 
& 1.000 & 1.000 & 30.25 
& 1.000 & 1.000 & 27.80
& 1.000 & 1.000 & 25.10 \\
Naive
& 1.433 & 1.170 & 23.17 
& 1.555 & 1.533 & 28.44 
& 1.157 & 1.194 & 26.80 
& 0.962 & 1.326 & 30.13 
& 1.002 & 0.896 & 30
& 1.260 & 2.069 & 31.13
& 1.133 & 1.066 & 25.9 \\
\bottomrule
\end{tabular}
}
\label{Tab:GIFT-Eval domain}
\vskip -0.15in
\end{table*}

\begin{table*}[ht]
    \centering
    \resizebox{0.5\columnwidth}{!}{
        \setlength\tabcolsep{1.2pt}
    \begin{tabular}{lccccccccccccccccccccccccccccc}
        \toprule
        % \textbf{Metric} & \textbf{YINGLong\_300m} & \textbf{YINGLong\_50m} & \textbf{Auto\_Arima} & \textbf{Auto\_ETS} & \textbf{Auto\_Theta} & \textbf{Chronos\_base} & \textbf{Chronos\_large} & \textbf{Chronos\_small} & \textbf{Crossformer} & \textbf{DLinear} & \textbf{DeepAR} & \textbf{Lag-Llama} & \textbf{Moirai\_base} & \textbf{Moirai\_large} & \textbf{Moirai\_small} & \textbf{N-BEATS} & \textbf{Naive} & \textbf{PatchTST} & \textbf{Seasonal\_Naive} & \textbf{TFT} & \textbf{TIDE} & \textbf{TTMs} & \textbf{TabPFN-TS} & \textbf{Timer} & \textbf{TimesFM} & \textbf{VisionTS} & \textbf{iTransformer} & \textbf{YINGLong\_100m} & \textbf{YINGLong\_5m} \\
        Model & \multicolumn{3}{c}{Long}  &\multicolumn{3}{c}{Medium} &\multicolumn{3}{c}{Short} \\
&MASE&CRPS&Rank
&MASE&CRPS&Rank
&MASE&CRPS&Rank\\
\midrule
\textsc{YINGLong}$_{\mathrm{300m}}$ 
&0.529	&0.368	&5.62
&0.837	&0.450	&6.14
&0.758	&0.511	&8.53\\
\textsc{YINGLong}$_{\mathrm{100m}}$
& 0.540	&0.379	&6.71
&0.853	&0.465	&7.71
&0.765	&0.514	&8.67\\
\textsc{YINGLong}$_{\mathrm{50m}}$ 
&0.550	&0.384	&7.1
&0.872	&0.472	&7.71
&0.776	&0.524	&9.76\\
\textsc{YINGLong}$_{\mathrm{6m}}$ 
&0.599	&0.418	&11.43
&0.948	&0.515	&12.19
&0.819	&0.557	&13.13\\
TabPFN-TS
&0.587	&0.405	&10.48
&0.936	&0.498	&10.38
&0.754	&0.506	&7.69\\
Chronos-bolt$_{b}$
&0.568	&0.424	&12.62
&0.894	&0.515	&11.10
&0.735	&0.500	&6.13\\
Chronos-bolt$_{s}$
&0.587	&0.431	&13.38
&0.916	&0.531	&13.19
&0.741	&0.494	&6.36\\
TimesFM-V2$_{500m}$
&0.525	&0.412	&11.38
&0.807	&0.486	&10.81
&0.704	&0.479	&7.22\\
Moirai$_{\textrm{large}}$
&0.601	&0.418	&11.10
&0.957	&0.510	&10.90
&0.806	&0.543	&10.49\\
Moirai$_{\textrm{base}}$
&0.628	&0.426	&11.57
&1.013	&0.532	&11.86
&0.817	&0.547	&10.16\\
Moirai$_{\textrm{small}}$
&0.631	&0.437	&12.38
&1.018	&0.535	&12.67
&0.888	&0.606	&15.44\\
Chronos$_{\textrm{large}}$
&0.632	&0.502	&18.71
&1.030	&0.622	&18.14
&0.761	&0.538	&11.93\\
Chronos$_{\textrm{base}}$
&0.634	&0.504	&18.33
&1.037	&0.630	&19.00
&0.768	&0.542	&11.93\\
Chronos$_{\textrm{small}}$
&0.658	&0.522	&19.76
&1.044	&0.625	&18.76
&0.779	&0.552	&13.24\\
TimesFM
&0.990	&0.518	&18.90
&1.441	&0.630	&18.00
&0.823	&0.577	&12.53\\
Lag-Llama
&0.724	&0.536	&20.10
&1.175	&0.672	&21.19
&1.262	&0.876	&23.64\\
TTMs
&0.731	&0.596	&22.48
&1.202	&0.756	&23.76
&0.993	&0.822	&24.75\\
Timer
&0.753	&0.650	&24.71
&1.223	&0.830	&25.67
&1.067	&0.891	&26.04\\
VisionTS
&0.522	&0.456	&16.71
&0.847	&0.583	&18.38
&0.871	&0.751	&23.67\\
\midrule
PatchTST
&0.537	&0.368	&7.52
&0.856	&0.461	&7.33
&0.832	&0.571	&13.04\\
iTransformer
&0.566	&0.391	&9.29
&0.867	&0.470	&8.62
&0.889	&0.611	&14.15\\
TFT
&0.589	&0.379	&8.38
&0.949	&0.468	&8.24
&0.883	&0.592	&14.87\\
TIDE
&0.655	&0.448	&15.29
&0.986	&0.563	&16.05
&1.140	&0.795	&21.93\\
DeepAR
&1.105	&0.628	&20.48
&1.334	&0.640	&17.14
&1.200	&0.795	&19.91\\
Crossformer
&0.921	&0.255	&13.62
&1.849	&0.461	&13.95
&3.574	&4.008	&27.35\\
N-BEATS
&0.644	&0.565	&20.29
&1.032	&0.678	&21.24
&0.862	&0.748	&22.80\\
DLinear
&0.700	&0.566	&22.05
&1.094	&0.684	&22.95
&1.015	&0.794	&24.40\\
\midrule
Auto Arima
&0.985	&0.805	&24.57
&1.022	&0.833	&23.90
&0.935	&0.735	&20.13\\
Auto Theta
&0.869	&1.397	&27.05
&1.175	&1.531	&27.10
&0.955	&0.816	&22.25\\
Auto ETS
&0.956	&369.207	&29.24
&1.610	&6.230	&28.71
&0.984	&1.347	&22.33\\
Seasional Naive
&1.000	&1.000	&26.48
&1.000	&1.000	&25.62
&1.000	&1.000	&26.60\\
Naive
&1.403	&1.892	&30.29
&1.462	&1.873	&29.57
&1.143	&1.093	&26.96\\
\bottomrule
        \bottomrule
    \end{tabular}
    }
    \caption{Results on GIFT-Eval aggregated by prediction length. The table shows MASE,CRPS and Rank for each method.}
    \label{tab:gift:length}
\end{table*}

\begin{table*}[htbp]
    \centering
    \caption{Results on GIFT-Eval aggregated by frequency. The table shows MASE, CRPS and Rank for each method.}
    \resizebox{1.0\columnwidth}{!}{
    \setlength\tabcolsep{1.2pt}
    \begin{tabular}{lccccccccccccccccccccccccccccc}
        \toprule
        \textbf{Daily} 
        & \textsc{YINGLong}$_{\mathrm{300m}}$ 
        &  \textsc{YINGLong}$_{\mathrm{100m}}$
        & \textsc{YINGLong}$_{\mathrm{50m}}$ 
        & \textsc{YINGLong}$_{\mathrm{6m}}$ 
        &TabPFN-TS
        & Chronos-bolt$_{b}$
        
        & Chronos-bolt$_{s}$
        
        & TimesFM-V2$_{500m}$
        & Chronos$_{\mathrm{large}}$ & Chronos$_{\mathrm{base}}$ & Chronos$_{\mathrm{small}}$
        & Moirai$_{\mathrm{large}}$ & Moirai$_{\mathrm{base}}$ & Moirai${_\mathrm{small}}$
        & TTMs	&Timer\\
        \midrule
        % 300m-100k-100000	50m-unet-100000	small-100k-100000-4096	tiny-100k-100000	Chronos_base	Chronos_large	Chronos_small	Moirai_base	Moirai_large	Moirai_small	TTMs	TabPFN-TS	Timer	TimesFM	VisionTS	Lag-Llama
MASE &0.712	&0.724	&0.729	&0.758	&0.706	&0.685	&0.689	&0.700	
&0.716	&0.714	&0.737	&0.727	&0.738	&0.752	&0.943	&0.982	\\
CRPS &0.367	&0.372	&0.370	&0.386	&0.347	&0.346	&0.350	&0.357	
&0.377	&0.378	&0.397	&0.375	&0.379	&0.384	&0.586	&0.637	\\
Rank &7.67  &8.60   &9.53   &12.73  &7.53   &5.00   &6.13   &7.33
&12.33  &12.73  &15.00  &11.60  &10.53  &12.20  &24.93  &26.00\\
      \midrule

     &TimesFM	&VisionTS	&Lag-Llama & Crossformer	&DLinear	&DeepAR	&N-BEATS	&PatchTST&	TFT&	TIDE	&iTransforme&	Auto Arima&	Auto ETS	&Auto Theta&	Naive	&Seasonal Naive\\
      \midrule
MASE    &0.746	&0.822	&1.188	&4.832	&0.887	&0.906	&0.775	&0.749	
&0.725	&1.146	&0.831	&0.882	&0.901	&0.936	&1.000	&1.000	\\
CRPS    &0.413	&0.504	&0.644	&3.596	&0.543	&0.491	&0.524	&0.392	
&0.370	&0.610	&0.438	&0.469	&0.907	&0.543	&0.794	&1.000	\\
Rank    &10.73  &23.33  &24.80  &28.73  &24.20  &19.07  &23.33  &13.20
&12.60  &22.00  &15.33  &17.93  &22.60  &22.93  &27.80  &29.53\\
      \midrule
       \textbf{Hourly} 
       & \textsc{YINGLong}$_{\mathrm{300m}}$ 
        &  \textsc{YINGLong}$_{\mathrm{100m}}$
        & \textsc{YINGLong}$_{\mathrm{50m}}$ 
        & \textsc{YINGLong}$_{\mathrm{6m}}$ 
        &TabPFN-TS
        & Chronos-bolt$_{b}$
        
        & Chronos-bolt$_{s}$
        
        & TimesFM-V2$_{500m}$
        & Chronos$_{\mathrm{large}}$ & Chronos$_{\mathrm{base}}$ & Chronos$_{\mathrm{small}}$
        & Moirai$_{\mathrm{large}}$ & Moirai$_{\mathrm{base}}$ & Moirai${_\mathrm{small}}$
        & TTMs	&Timer\\
        \midrule
MASE   &0.725	&0.738	&0.737	&0.796	&0.770	&0.662	&0.690	&0.734	
&0.763	&0.763	&0.773	&0.763	&0.775	&0.867	&0.853	&0.941	\\
CRPS   &0.390	&0.396	&0.397	&0.426	&0.414	&0.374	&0.382	&0.409	
&0.464	&0.462	&0.468	&0.409	&0.413	&0.454	&0.569	&0.633	\\
Rank   &7.03    &8.16   &7.77   &12.03  &10.10  &6.65   &7.55   &9.71
&14.87  &14.55  &15.45  &8.39   &8.61   &14.58  &22.74  &24.55\\
      \midrule
      &TimesFM	&VisionTS	&Lag-Llama & Crossformer	&DLinear	&DeepAR	&N-BEATS	&PatchTST&	TFT&	TIDE	&iTransforme&	Auto Arima&	Auto ETS	&Auto Theta&	Naive	&Seasonal Naive\\
      \midrule
MASE    &0.824	&0.770	&0.895	&1.733	&0.943	&1.309	&0.872	&0.774	
&0.825	&0.959	&0.805	&1.022	&1.269	&1.276	&1.460	&1.000	\\
CRPS    &0.469	&0.525	&0.489	&0.534	&0.606	&0.623	&0.600	&0.407	
&0.428	&0.511	&0.424	&0.743	&50.063	&1.573	&1.667	&1.000	\\
Rank    &15.29  &20.68  &18.55  &17.94  &24.06  &18.39  &23.23  &10.00
&12.52  &18.55  &11.81  &25.84  &29.68  &29.77  &30.58  &28.39\\
	\midrule
     \textbf{Minutely} 
          & \textsc{YINGLong}$_{\mathrm{300m}}$ 
        &  \textsc{YINGLong}$_{\mathrm{100m}}$
        & \textsc{YINGLong}$_{\mathrm{50m}}$ 
        & \textsc{YINGLong}$_{\mathrm{6m}}$ 
        &TabPFN-TS
        & Chronos-bolt$_{b}$
        
        & Chronos-bolt$_{s}$
        
        & TimesFM-V2$_{500m}$
        & Chronos$_{\mathrm{large}}$ & Chronos$_{\mathrm{base}}$ & Chronos$_{\mathrm{small}}$
        & Moirai$_{\mathrm{large}}$ & Moirai$_{\mathrm{base}}$ & Moirai${_\mathrm{small}}$
        & TTMs	&Timer\\
        \midrule
MASE    &0.786	&0.801	&0.821	&0.872	&0.825	&0.796	&0.816	&0.727	
&0.914	&0.914	&0.950	&0.868	&0.867	&0.924	&1.081	&1.155	\\
CRPS    &0.497	&0.512	&0.527	&0.572	&0.534	&0.553	&0.574	&0.534	
&0.659	&0.659	&0.686	&0.560	&0.568	&0.597	&0.784	&0.955	\\
Rank    &4.70   &5.43   &6.67   &10.63  &9.13   &9.50   &11.53  &10.57  
&17.00  &16.87  &18.30  &10.93  &11.50  &12.43  &22.33  &26.67\\
      \midrule
     &TimesFM	&VisionTS	&Lag-Llama & Crossformer	&DLinear	&DeepAR	&N-BEATS	&PatchTST&	TFT&	TIDE	&iTransforme&	Auto Arima&	Auto ETS	&Auto Theta&	Naive	&Seasonal Naive\\
      \midrule
MASE    &1.458	&0.871	&1.199	&2.044	&1.139	&1.564	&0.996	&0.892	
&0.907	&1.042	&0.872	&0.991	&1.306	&1.106	&1.211	&1.000	\\
CRPS    &0.657	&0.693	&0.840	&0.868	&0.793	&0.862	&0.792	&0.551	
&0.554	&0.677	&0.552	&0.981	&7.131	&1.388	&1.701	&1.000	\\
Rank    &16.93  &19.30  &22.67  &19.50  &22.73  &21.40  &22.07  &9.07
&9.53   &17.53  &9.67   &24.70  &26.50  &27.43  &28.77  &26.00  \\
      \midrule
      \textbf{Monthly} 
        & \textsc{YINGLong}$_{\mathrm{300m}}$ 
        &  \textsc{YINGLong}$_{\mathrm{100m}}$
        & \textsc{YINGLong}$_{\mathrm{50m}}$ 
        & \textsc{YINGLong}$_{\mathrm{6m}}$ 
        &TabPFN-TS
        & Chronos-bolt$_{b}$
        
        & Chronos-bolt$_{s}$
        
        & TimesFM-V2$_{500m}$
        & Chronos$_{\mathrm{large}}$ & Chronos$_{\mathrm{base}}$ & Chronos$_{\mathrm{small}}$
        & Moirai$_{\mathrm{large}}$ & Moirai$_{\mathrm{base}}$ & Moirai${_\mathrm{small}}$
        & TTMs	&Timer\\
        \midrule
MASE    &0.867	&0.846	&0.868	&0.895	&0.744	&0.842	&0.810	&0.700	
&0.812	&0.857	&0.827	&0.825	&0.815	&1.016	&1.182	&1.210	\\
CRPS    &0.806	&0.780	&0.809	&0.846	&0.680	&0.787	&0.758	&0.660	
&0.807	&0.849	&0.818	&0.781	&0.753	&0.979	&1.409	&1.450	\\
Rank    &14.60  &12.80  &14.20  &18.00  &2.80   &9.40   &9.40   &6.60   
&14.20  &15.80  &14.40  &10.80  &8.00   &21.60  &27.40  &26.60  \\
      \midrule
     &TimesFM	&VisionTS	&Lag-Llama & Crossformer	&DLinear	&DeepAR	&N-BEATS	&PatchTST&	TFT&	TIDE	&iTransforme&	Auto Arima&	Auto ETS	&Auto Theta&	Naive	&Seasonal Naive\\
      \midrule
MASE    &0.800	&0.915	&1.430	&2.950	&0.997	&1.215	&0.851	&0.859	
&0.901	&1.099	&0.907	&0.759	&0.821	&0.932	&1.204	&1.000	\\
CRPS    &0.733	&1.030	&1.352	&5.640	&1.141	&1.030	&0.962	&0.832	
&0.840	&1.157	&0.803	&0.759	&0.770	&0.873	&1.524	&1.000	\\
Rank    &8.40   &25.20  &22.40  &19.60  &26.20  &19.20  &20.40  &14.80
&15.20  &25.40  &12.00  &12.60  &10.80  &16.80  &28.80  &23.60\\	
\midrule
     \textbf{Quarterly} 
        & \textsc{YINGLong}$_{\mathrm{300m}}$ & \textsc{YINGLong}$_{\mathrm{100m}}$
        & \textsc{YINGLong}$_{\mathrm{50m}}$ & \textsc{YINGLong}$_{\mathrm{6m}}$ 
        &TabPFN-TS  & Chronos-bolt$_{b}$    & Chronos-bolt$_{s}$
        & TimesFM-V2$_{500m}$   & Chronos$_{\mathrm{large}}$ & Chronos$_{\mathrm{base}}$ & Chronos$_{\mathrm{small}}$
        & Moirai$_{\mathrm{large}}$ & Moirai$_{\mathrm{base}}$ & Moirai${_\mathrm{small}}$  & TTMs	&Timer\\
        \midrule
MASE    &0.869	&0.921	&0.876	&0.918	&0.732	&0.765	&0.780	&0.603	
&0.769	&0.769	&0.775	&0.713	&0.713	&0.775	&1.240	&1.838	\\
CRPS    &0.865	&0.889	&0.882	&0.914	&0.757	&0.777	&0.787	&0.623	
&0.840	&0.840	&0.846	&0.740	&0.740	&0.793	&1.287	&1.787	\\
Rank    &19.00  &21.00  &20.00  &22.00  &4.00   &5.00   &6.00   &1.00
&15.00  &14.00  &17.00  &3.00   &2.00   &7.00   &29.00  &30.00  \\
      \midrule
      &TimesFM	&VisionTS	&Lag-Llama & Crossformer	&DLinear	&DeepAR	&N-BEATS	&PatchTST&	TFT&	TIDE	&iTransforme&	Auto Arima&	Auto ETS	&Auto Theta&	Naive	&Seasonal Naive\\
      \midrule
MASE    &0.875	&0.850	&3.730	&15.875	   &0.913	&0.900	&0.756	
&0.825	&0.813	&1.050	&0.769	&0.800	&0.725	&0.744	&0.925	&1.000	\\
CRPS    &0.853	&1.048	&2.517	&119.960   &1.109	&0.841	&0.972	
&0.835	&0.837	&1.018	&0.797	&0.823	&0.798	&0.797	&0.951	&1.000	\\
Rank    &18.00  &27.00  &31.00  &32.00     &28.00   &16.00  &24.00
&12.00  &13.00  &26.00  &9.00   &11.00  &10.00  &8.00  &23.00 &25.00\\
\midrule
     \textbf{Secondly} 
       & \textsc{YINGLong}$_{\mathrm{300m}}$ & \textsc{YINGLong}$_{\mathrm{100m}}$
        & \textsc{YINGLong}$_{\mathrm{50m}}$ & \textsc{YINGLong}$_{\mathrm{6m}}$ 
        &TabPFN-TS  & Chronos-bolt$_{b}$    & Chronos-bolt$_{s}$
        & TimesFM-V2$_{500m}$   & Chronos$_{\mathrm{large}}$ & Chronos$_{\mathrm{base}}$ & Chronos$_{\mathrm{small}}$
        & Moirai$_{\mathrm{large}}$ & Moirai$_{\mathrm{base}}$ & Moirai${_\mathrm{small}}$  & TTMs	&Timer\\
        \midrule
MASE    &0.264	&0.241	&0.276	&0.321	&0.467	&0.682	&0.632	&0.227	
&0.506	&0.523	&0.523	&0.524	&0.786	&0.510	&0.728	&0.517	\\
CRPS    &0.497	&0.482	&0.513	&0.573	&0.785	&1.138	&0.874	&0.429	
&0.818	&0.859	&0.793	&0.873	&1.025	&0.977	&1.529	&0.957	\\
Rank    &7.83   &6.50   &8.83   &12.17  &16.50  &27.00  &20.33  &3.83   
&18.33  &19.50  &17.17  &21.17  &24.33  &22.17  &28.83  &23.17  \\
      \midrule
      &TimesFM	&VisionTS	&Lag-Llama & Crossformer	&DLinear	&DeepAR	&N-BEATS	&PatchTST&	TFT&	TIDE	&iTransforme&	Auto Arima&	Auto ETS	&Auto Theta&	Naive	&Seasonal Naive\\
      \midrule
MASE    &0.787	&0.216	&0.598	&0.615	&0.368	&0.376	&0.271	&0.224	
&0.537	&0.323	&0.235	&1.000	&0.551	&0.159	&1.983	&1.000	\\
CRPS    &1.298	&0.691	&1.116	&0.729	&0.782	&0.754	&0.598	&0.536	
&0.672	&0.705	&0.510	&1.000	&1.934	&0.315	&1.441	&1.000	\\
Rank    &29.17  &14.83  &24.33  &16.67  &17.67  &16.83  &9.67   &7.00   
&10.67  &15.00  &4.67   &13.83  &29.17  &1.00   &25.00  &14.83  \\
\midrule
     \textbf{Weekly} 
       & \textsc{YINGLong}$_{\mathrm{300m}}$ & \textsc{YINGLong}$_{\mathrm{100m}}$
        & \textsc{YINGLong}$_{\mathrm{50m}}$ & \textsc{YINGLong}$_{\mathrm{6m}}$ 
        &TabPFN-TS  & Chronos-bolt$_{b}$    & Chronos-bolt$_{s}$
        & TimesFM-V2$_{500m}$   & Chronos$_{\mathrm{large}}$ & Chronos$_{\mathrm{base}}$ & Chronos$_{\mathrm{small}}$
        & Moirai$_{\mathrm{large}}$ & Moirai$_{\mathrm{base}}$ & Moirai${_\mathrm{small}}$  & TTMs	&Timer\\
        \midrule
MASE    &0.851	&0.902	&0.899	&0.963	&0.739	&0.747	&0.763	&0.863	
&0.737	&0.762	&0.745	&0.931	&0.911	&0.967	&1.133	&1.136	\\
CRPS    &0.595	&0.621	&0.642	&0.692	&0.520	&0.523	&0.532	&0.589	
&0.529	&0.542	&0.536	&0.634	&0.640	&0.688	&1.058	&0.979	\\
Rank    &10.00  &10.75  12.25   &15.25  &4.50   &5.25   &5.88   &9.63
&7.50   &8.13   &8.88   &11.38  &11.25  &14.38  &26.88  &25.75\\
              \midrule
      &TimesFM	&VisionTS	&Lag-Llama & Crossformer	&DLinear	&DeepAR	&N-BEATS	&PatchTST&	TFT&	TIDE	&iTransforme&	Auto Arima&	Auto ETS	&Auto Theta&	Naive	&Seasonal Naive\\
      \midrule
MASE    &0.847	&1.038	&1.629	&4.364	&1.137  &1.460	&1.082	&0.929	
&0.921	&1.294	&1.248	&0.946	&0.932	&1.028	&1.000	&1.000	\\
CRPS    &0.602	&0.943	&1.203	&11.775	&0.948	&0.994	&0.971	&0.666	
&0.726	&0.956	&0.956	&0.731	&0.774	&0.787	&0.874	&1.000	\\
Rank    &9.38   &25.13  &27.63  &31.13  &25.50  &20.88  &24.25  &13.88
&18.25  &20.75  &19.88  &17.75  &18.25  &20.50  &22.00  &25.2\\
\midrule
     \textbf{Yearly} 
        & \textsc{YINGLong}$_{\mathrm{300m}}$ & \textsc{YINGLong}$_{\mathrm{100m}}$
        & \textsc{YINGLong}$_{\mathrm{50m}}$ & \textsc{YINGLong}$_{\mathrm{6m}}$ 
        &TabPFN-TS  & Chronos-bolt$_{b}$    & Chronos-bolt$_{s}$
        & TimesFM-V2$_{500m}$   & Chronos$_{\mathrm{large}}$ & Chronos$_{\mathrm{base}}$ & Chronos$_{\mathrm{small}}$
        & Moirai$_{\mathrm{large}}$ & Moirai$_{\mathrm{base}}$ & Moirai${_\mathrm{small}}$  & TTMs	&Timer\\
        \midrule
MASE    &1.086	&1.188	&1.097	&1.328	&0.796	&0.883	&0.929	&0.639	
&0.917	&0.917	&0.942	&0.748	&0.758	&0.751	&1.288	&2.899	\\
CRPS    &1.099	&1.151	&1.090	&1.286	&0.822	&0.880	&0.926	&0.657	
&0.978	&0.978	&1.007	&0.754	&0.761	&0.761	&1.390	&2.923	\\
Rank    &23.00  &25.00  &22.00  &28.00  &8.00   &13.00  &14.00  &1.00   &18.00  &17.00  &21.00  &2.00   &3.00   &4.00   &29.00  &31.00  \\        
\midrule
       &TimesFM	&VisionTS	&Lag-Llama & Crossformer	&DLinear	&DeepAR	&N-BEATS	&PatchTST&	TFT&	TIDE	&iTransforme&	Auto Arima&	Auto ETS	&Auto Theta&	Naive	&Seasonal Naive\\
      \midrule
MASE    &0.844	&0.965	&2.409	&7.909	&1.048	&0.856	&0.793	&0.829	
&0.778	&1.264	&0.849	&0.935	&0.776	&0.783	&1.000	&1.000	\\
CRPS    &0.848	&1.152	&2.039	&102.899&1.217	&0.819	&0.971	&0.848	
&0.797	&1.123	&0.848	&0.942	&0.804	&0.833	&0.993	&1.000	\\
Rank    &11.00  &26.00  &30.00  &32.00  &27.00  &7.00   &16.00  &10.00
&5.00   &24.00  &12.00  &15.00   &6.00  &9.00   &19.00  &20.00\\
        \bottomrule
    \end{tabular}\label{tab:gift:freq}
    }
\end{table*}
\begin{table*}[h]
    \centering
    \resizebox{1.0\columnwidth}{!}{
    \setlength\tabcolsep{1.2pt}
    % \begin{tabular}{|l|*{30}{c|}}
        % \hline
    \begin{tabular}{lccccccccccccccccccccccccccccc}
        \toprule
        % \textbf{Metric} & \textbf{YINGLong\_300m} & \textbf{YINGLong\_50m} & \textbf{Auto\_Arima} & \textbf{Auto\_ETS} & \textbf{Auto\_Theta} & \textbf{Chronos\_base} & \textbf{Chronos\_large} & \textbf{Chronos\_small} & \textbf{Crossformer} & \textbf{DLinear} & \textbf{DeepAR} & \textbf{Lag-Llama} & \textbf{Moirai\_base} & \textbf{Moirai\_large} & \textbf{Moirai\_small} & \textbf{N-BEATS} & \textbf{Naive} & \textbf{PatchTST} & \textbf{Seasonal\_Naive} & \textbf{TFT} & \textbf{TIDE} & \textbf{TTMs} & \textbf{TabPFN-TS} & \textbf{Timer} & \textbf{TimesFM} & \textbf{VisionTS} & \textbf{iTransformer} & \textbf{YINGLong\_100m} & \textbf{YINGLong\_5m} \\
                \textbf{Multivariate} 
        & \textsc{YINGLong}$_{\mathrm{300m}}$ 
        &  \textsc{YINGLong}$_{\mathrm{100m}}$
        & \textsc{YINGLong}$_{\mathrm{50m}}$ 
        & \textsc{YINGLong}$_{\mathrm{6m}}$ 
        &TabPFN-TS
        & Chronos-bolt$_{b}$
        
        & Chronos-bolt$_{s}$
        
        & TimesFM-V2$_{500m}$
        & Chronos$_{\mathrm{large}}$ & Chronos$_{\mathrm{base}}$ & Chronos$_{\mathrm{small}}$
        & Moirai$_{\mathrm{large}}$ & Moirai$_{\mathrm{base}}$ & Moirai${_\mathrm{small}}$
        & TTMs	&Timer	 \\
        \midrule
MASE	&0.658	&0.65	&0.668	&0.709
&0.756	
&0.725	&0.736
&0.629
&0.788  & 0.794 & 0.804 
& 0.802 & 0.826  & 0.800	
&0.930 &0.895\\
CRPS	&0.421	&0.422	&0.43	&0.459	
&0.482
&0.490 &0.487 
&0.448 
& 0.552&0.555 &0.555 & 0.515&0.516& 0.519 & 0.694 & 0.716
\\
Rank	&5.95	&5.63	&6.19	&9.95
&10.60
&10.70 &11.21
&10.72
&16.67 & 16.47 & 16.40
&11.98 & 12.09& 12.72
&23.30 & 24.81\\
\midrule
    &TimesFM	&VisionTS	&Lag-Llama & Crossformer	&DLinear	&DeepAR	&N-BEATS	&PatchTST&	TFT&	TIDE	&iTransforme&	Auto Arima&	Auto ETS	&Auto Theta&	Naive	&Seasonal Naive\\
      \midrule
MASE	&1.173&0.695&0.957&1.160&0.963&1.495&0.782&0.711&0.840&1.007&0.737&1.033&1.055&0.801&1.147&1.000\\
CRPS	&0.582&0.585&0.647&0.669&0.663&0.802&0.641	&0.451&0.495&0.659&0.478&0.837&4.053&0.926&1.259&1.000\\
Rank	&17.67 &19.35&21.60&21.28&22.37&22.30	&20.77&9.33&11.95&18.98&9.98&23.02
&26.74
&23.63&27.63&26.00\\
\midrule 
\textbf{Unvariate} & \textsc{YINGLong}$_{\mathrm{300m}}$ 
        &  \textsc{YINGLong}$_{\mathrm{100m}}$
        & \textsc{YINGLong}$_{\mathrm{50m}}$ 
        & \textsc{YINGLong}$_{\mathrm{6m}}$ 
        &TabPFN-TS
        & Chronos-bolt$_{b}$
        
        & Chronos-bolt$_{s}$
        
        & TimesFM-V2$_{500m}$
        & Chronos$_{\mathrm{large}}$ & Chronos$_{\mathrm{base}}$ & Chronos$_{\mathrm{small}}$
        & Moirai$_{\mathrm{large}}$ & Moirai$_{\mathrm{base}}$ & Moirai${_\mathrm{small}}$
        & TTMs	&Timer	 \\
        \midrule
MASE	&0.766	&0.793	&0.799	&0.861	&0.742	&0.725	&0.739	
&0.724	&0.775	&0.780	&0.797	&0.773	&0.795	&0.890	&1.001	&1.131	\\
CRPS	&0.500	&0.514	&0.522	&0.564	&0.479	&0.482	&0.488		
&0.479	&0.543	&0.547	&0.564	&0.499	&0.514	&0.575	&0.803	&0.913	\\
Rank	&8.52   &9.96   &10.78  &14.63	&7.50   &6.94   &7.89 
&7.44   &13.20  &13.56  &15.41  &9.70   &9.83   &15.33  &24.63 &26.35\\
\midrule
    &TimesFM	&VisionTS	&Lag-Llama & Crossformer	&DLinear	&DeepAR	&N-BEATS	&PatchTST&	TFT&	TIDE	&iTransforme&	Auto Arima&	Auto ETS	&Auto Theta&	Naive	&Seasonal Naive\\
      \midrule
MASE    &0.829	&0.845	&1.233	&4.000    &0.944	&1.016	&0.892		&0.805	&0.808	&0.959	&0.857   &0.912	&1.115 &1.147 &1.358 &1.000					\\
CRPS	&0.569	&0.683	&0.831	&2.464	  &0.758	&0.662	&0.730	
&0.535	&0.524	&0.646	&0.564	   &0.721	&9.018	&1.162	&1.491 &1.000		\\
Rank	&13.04	&22.35  &22.93  &21.63    &24.54    &17.15  &22.83
&11.63&12.09  &19.41  &13.43  &21.02 &23.98   &24.91 & 28.74 & 26.65\\
\bottomrule

    \end{tabular}
    }
    \caption{Results on GIFT-Eval aggregated by number of variates. The table shows MASE, CRPS and Rank for each method.}
    \label{tab:gift:nvariate}
\end{table*}

\clearpage
\section{Error Reduction Pattern Analysis}
\begin{figure*}[h]
    \centering
    \subfigure[Relative MSE Reduction ETTm1]{
        \includegraphics[width=0.45\columnwidth]{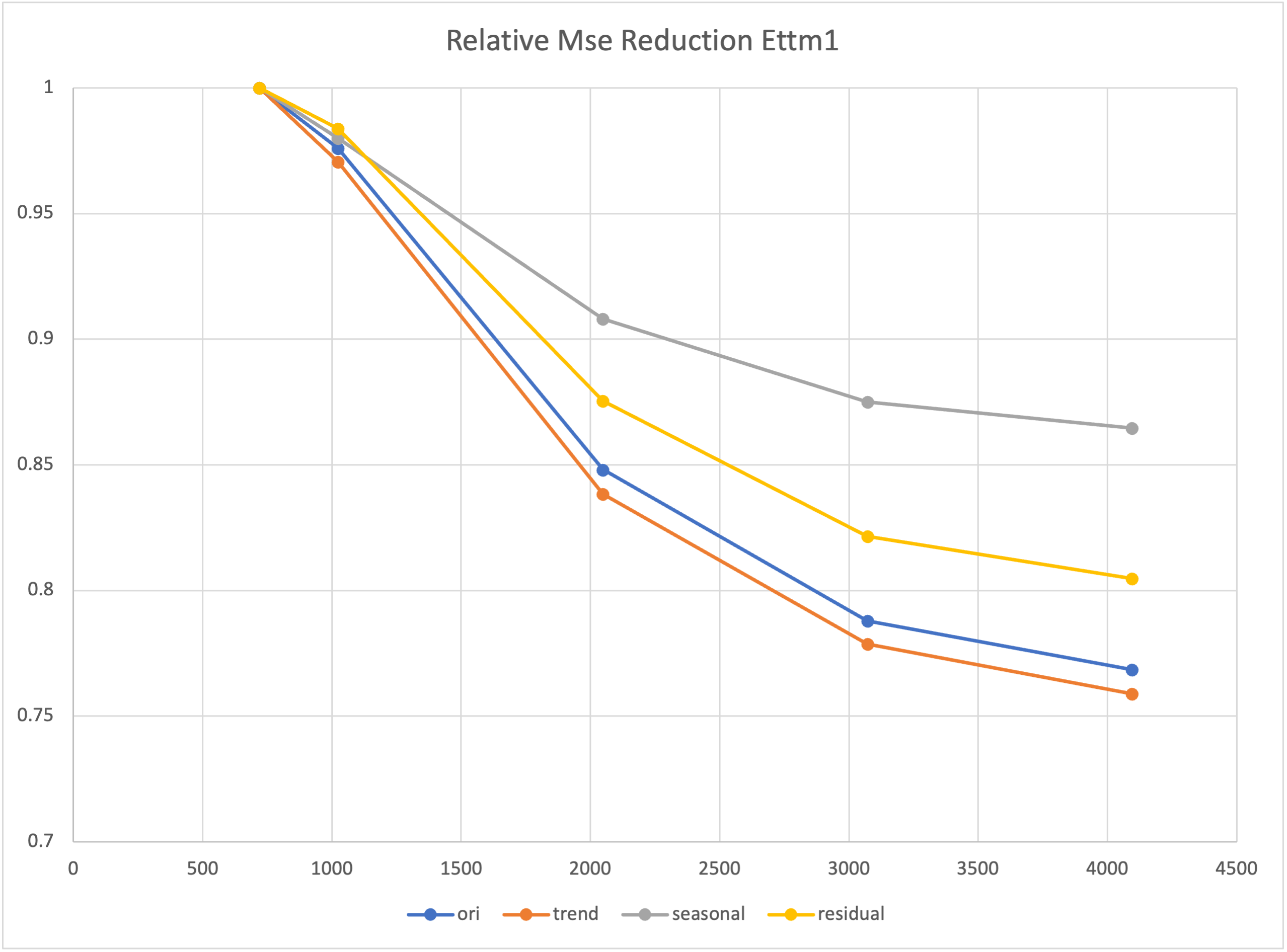}
        \label{fig:mse_reduction_1}
    }
    \hfill
    \subfigure[Relative MSE Reduction ETTh2]{
        \includegraphics[width=0.45\columnwidth]{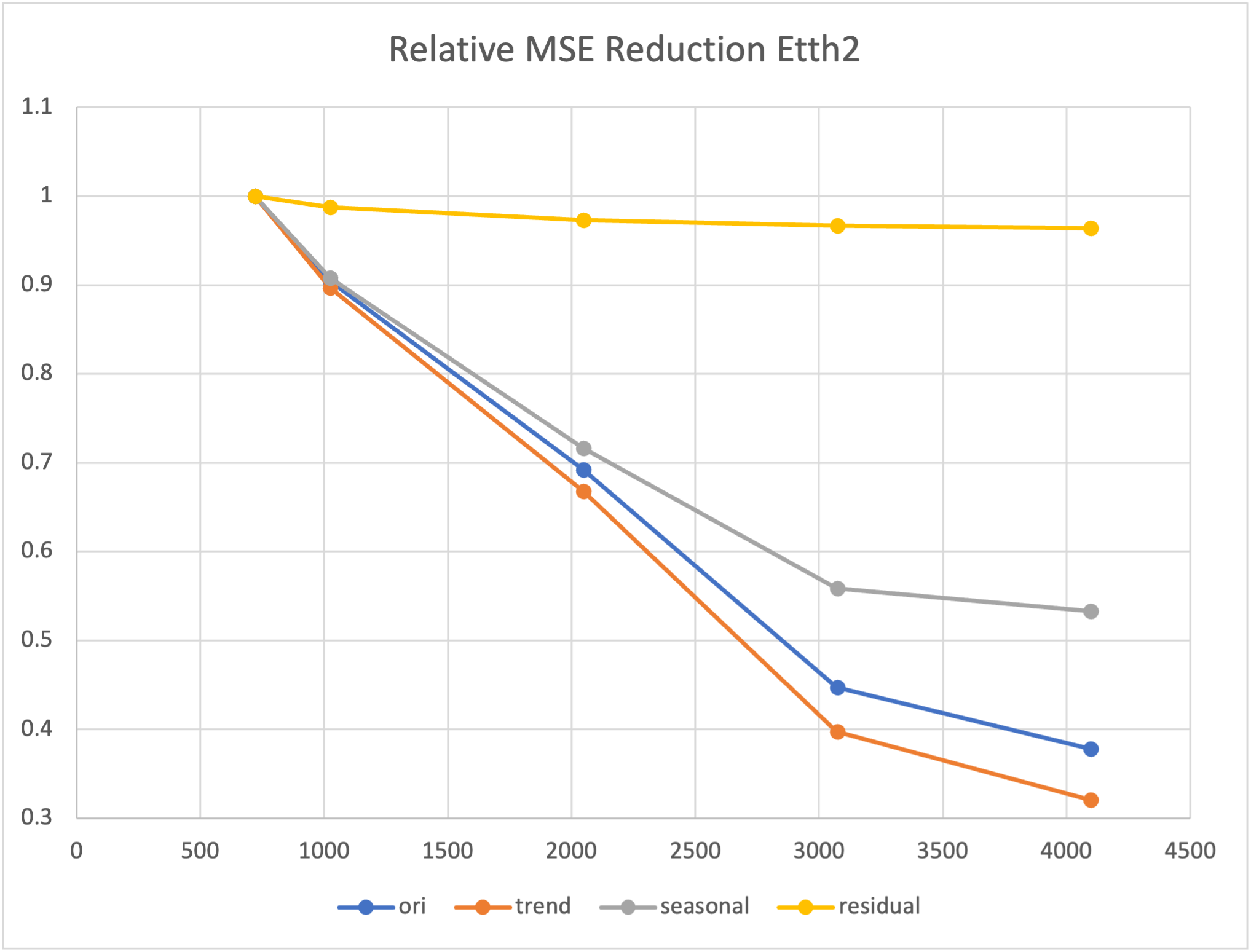}
        \label{fig:mse_reduction_2}
    }
    \caption{Comparison of relative MSE reduction in ETTm1 models.}
    \label{fig:mse_reduction_side_by_side}
\end{figure*}

We performed STL decomposition on the target and predicted sequences generated by our model with various DCoT lengths to determine the primary source of error reduction. It is hypothesized that error reduction primarily arises from DCoT tokens, which likely encapsulate information related to general signal patterns or low-frequency trends. Consequently, the majority of error reduction may be attributed to improvements in trend prediction. Our initial experiments corroborate this hypothesis. Furthermore, we analyzed relative improvements with extending DCoT lengths, as absolute MSE reductions may yield dataset-dependent results, given that some datasets are more amenable to trend-based enhancements rather than seasonal improvements. As illustrated in Figure \ref{fig:mse_reduction_side_by_side}, relative improvements highlight that error reductions in the trend are more pronounced compared to seasonal reductions, while the residual component remains stable due to its generally unlearnable nature.
\begin{table}[h]
    \centering
    \sisetup{table-number-alignment=center, round-mode=places, round-precision=5}
    \begin{tabular}{@{}cccccc@{}}
        \toprule
        \textbf{DCOT(Output)} & \textbf{MSE\_all} & \textbf{MSE\_trend} & \textbf{MSE\_seasonal} & \textbf{MSE\_residual} \\ 
        \midrule
        720  & \num{0.550927696} & \num{0.374313515} & \num{0.017017007} & \num{0.062446221} \\
        1024 & \num{0.537688743} & \num{0.363281141} & \num{0.01667509}  & \num{0.061433353} \\
        2048 & \num{0.467228681} & \num{0.313806443} & \num{0.015451411} & \num{0.054663218} \\
        3072 & \num{0.434103499} & \num{0.291480331} & \num{0.014888517} & \num{0.051302693} \\
        4096 & \num{0.423354131} & \num{0.284063046} & \num{0.014712034} & \num{0.050248258} \\
        \bottomrule
    \end{tabular}
    \caption{Performance metrics across different DCOT outputs for analysis in ETTm1.}
    \label{tab:dcot_performance}
\end{table}

\section{Influence of Input Ensemble}
We compare the average performance of forecasts generated using multiple input lengths against the single-best component, as detailed in Appendix Table~\ref{tab:ablation:input_length}. Our input-ensemble strategy consistently enhances accuracy by 1\% to 4\% without necessitating additional model training. This presents a scalable 'free lunch' approach, yielding post-training improvements through simple ensemble averaging. When compared to the single-worst component in the setup, the accuracy improvement is even more pronounced. Since comparative accuracy assessments cannot be made prior to determining the target value, this ensemble technique proves to be both practical and robust for real-world applications.
\begin{table}[th]
\centering
\caption{Influence of Multi-Input Ensemble}
\label{tab:ablation:input_length}
\scalebox{0.78}{
\setlength\tabcolsep{1.8pt}
\begin{tabular}{@{}lcccccccccccccccccc@{}}
\toprule
\multirow{2}{*}{Model} & \multicolumn{8}{c}{ETTh1} & \multicolumn{8}{c}{Weather} \\
\cmidrule(lr){2-9}\cmidrule(lr){10-17}
& \multicolumn{2}{c}{1024} & \multicolumn{2}{c}{2048} & \multicolumn{2}{c}{4096} & \multicolumn{2}{c}{Ensemble}
& \multicolumn{2}{c}{1024} & \multicolumn{2}{c}{2048} & \multicolumn{2}{c}{4096} & \multicolumn{2}{c}{Ensemble} \\
& MSE & MAE & MSE & MAE & MSE & MAE & MSE & MAE 
& MSE & MAE & MSE & MAE & MSE & MAE & MSE & MAE \\
\midrule
YINGLONG$_{6m}$  & 0.411 & 0.413 & 0.414 & 0.423 & 0.437 & 0.45  & 0.408 & 0.412 
   & 0.248 & 0.27  & 0.247 & 0.272 & 0.243 & 0.274 & 0.239 & 0.268 \\
YINGLONG$_{50m}$ & 0.409 & 0.412 & 0.396 & 0.409 & 0.388 & 0.41  & 0.405 & 0.408 
   & 0.246 & 0.266 & 0.235 & 0.261 & 0.226 & 0.257 & 0.231 & 0.257 \\
YINGLONG$_{100m}$&0.406 &0.410 &0.412 &0.414 &0.389 &0.407 &0.399 &0.408 
   &0.242 &0.262 &0.233 &0.256 &0.23 &0.256 &0.227 &0.255 \\
YINGLONG$_{300m}$&0.425 &0.417 &0.412 &0.414 &0.393 &0.412 &0.398 &0.407 
   &0.233 &0.255 &0.22 &0.245 &0.215 &0.245 &0.217 &0.245 \\
\bottomrule
\end{tabular}
}
\vskip -0.1in
\end{table}
\section{Structure ablations}
We performed a structural ablation study on our U-transformer and token merge designs. As illustrated in Figure~\ref{fig:structure_ablation}, both designs contribute to 1\% to 5\% improvements in MSE and MAE, depending on the dataset tested. Generally, the U-transformer architecture employed in $YINGLONG$ yields the best results, which is why it serves as the backbone architecture. However, it is important to note that the architectural design alone is not the primary driver of the state-of-the-art performance seen in $YINGLONG$. The key contributors are the joint forecasting paradigm and the DCoT design.
\begin{figure}[h]
    \centering
    \subfigure[MSE Comparison]{
        \adjustbox{max width=0.45\columnwidth}{
        \begin{tabular}{lccc}
            \toprule
            \textbf{Dataset} & \textbf{Transformer} & \textbf{Token-Merge+} & \textbf{uTransformer+} \\
            \midrule
            etth1   & 0.407 & 0.408 & 0.408 \\
            etth2   & 0.361 & 0.348 & 0.344 \\
            ettm1   & 0.381 & 0.375 & 0.370 \\
            ettm2   & 0.255 & 0.252 & 0.250 \\
            weather & 0.242 & 0.241 & 0.239 \\
            \bottomrule
        \end{tabular}
        }
        \label{tab:mse_comparison}
    }
    \hfill
    \subfigure[MAE Comparison]{
        \adjustbox{max width=0.45\columnwidth}{
        \begin{tabular}{lccc}
            \toprule
            \textbf{Dataset} & \textbf{Transformer} & \textbf{Token-Merge+} & \textbf{uTransformer+} \\
            \midrule
            etth1   & 0.423 & 0.420 & 0.412 \\
            etth2   & 0.402 & 0.390 & 0.382 \\
            ettm1   & 0.374 & 0.367 & 0.368 \\
            ettm2   & 0.317 & 0.305 & 0.302 \\
            weather & 0.277 & 0.273 & 0.268 \\
            \bottomrule
        \end{tabular}
        }
        \label{tab:mae_comparison}
    }
    \caption{Structure ablation:6M size Transformer, add Token-Merge, and uTransformer($Yinglong$) model variates across different datasets for MSE and MAE metrics.}
    \label{fig:structure_ablation}
\end{figure}

\section{Output Scaling for vanilla transformer model}
We investigated the output scaling effect across a range of vanilla transformer models with sizes from 6 million to 300 million parameters within our joint forecasting paradigm. For all models of varying sizes, the output scaling effect was significant, and the scaling effect with model size persisted. Specifically, for the largest 300M parameter model, our joint forecasting approach resulted in substantial improvements in Mean Absolute Scaled Error (MASE) and Continuous Ranked Probability Score (CRPS) by 24.9\% and 30.0\%, respectively, when comparing the DCoT setting with a token length of 4096 to a non-DCoT configuration. This effect clearly demonstrates robust output scaling within the extensive GIFT-Eval benchmark, which encompasses 23 datasets. These findings indicate that the observed scaling effect is neither specific to a particular architectural model design nor limited to a single dataset.
\vspace{-0.3cm}
\begin{figure*}[ht]
    \centering
    \includegraphics[width=0.45\textwidth]{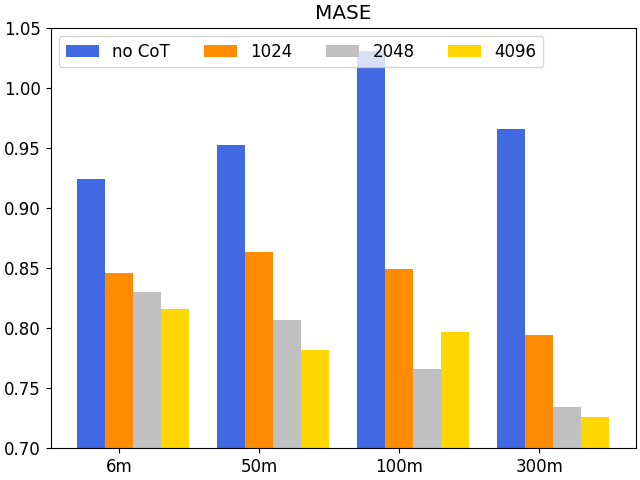}
    \includegraphics[width=0.45\textwidth]{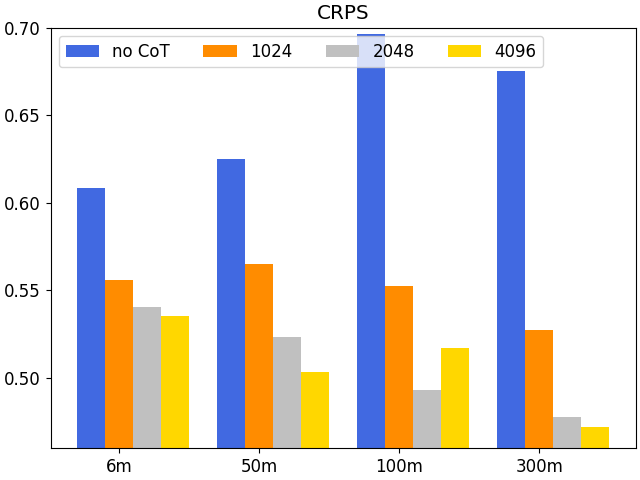}
    \caption{Output Scaling for 6M to 300M vanilla transformer model following joint forecasting paradigm under different DCOT setting}
    \label{fig:ablation:DCOT_transformer}
    \vspace{-0.3cm}
\end{figure*}

\newpage

% \begin{figure*}[ht]
%     \centering
%     \begin{minipage}{0.45\textwidth}
%         \centering
%         \includegraphics[width=\textwidth]{figures/ETTh1-scaling.png}
%         \caption{Output-Scaling:Detail}
%         \label{fig:output-image1}
%     \end{minipage}\hfill
%     \begin{minipage}{0.45\textwidth}
%         \centering
%         \includegraphics[width=\textwidth]{figures/ETTh1-scaling-avg.png}
%         \caption{Output-Scaling:Avg}
%         \label{fig:output-image2}
%     \end{minipage}
% \end{figure*}

% \begin{figure*}[ht]
%     \centering
%     \begin{minipage}{0.45\textwidth}
%         \centering
%         \includegraphics[width=\textwidth]{figures/ETTm2-scaling.png}
%         \caption{Input-Scaling:Detail}
%         \label{fig:input-image1}
%     \end{minipage}\hfill
%     \begin{minipage}{0.45\textwidth}
%         \centering
%         \includegraphics[width=\textwidth]{figures/ETTm2-scaling-avg.png}
%         \caption{Input-Scaling:Avg}
%         \label{fig:input-image2}
%     \end{minipage}
% \end{figure*}

%%%%%%%%%%%%%%%%%%%%%%%%%%%%%%%%%%%%%%%%%%%%%%%%%%%%%%%%%%%%%%%%%%%%%%%%%%%%%%%
%%%%%%%%%%%%%%%%%%%%%%%%%%%%%%%%%%%%%%%%%%%%%%%%%%%%%%%%%%%%%%%%%%%%%%%%%%%%%%%

%%%%%%%%%%%%%%%%%%%%%%%%%%%%%%%%%%%%%%%%%%%%%%%%%%%%%%%%%%%%

\end{document}